%% file: acl_latex.tex
\PassOptionsToPackage{table}{xcolor}

\documentclass[11pt]{article}

\usepackage[final]{acl}
\usepackage{times}
\usepackage{latexsym}

\usepackage[T1]{fontenc}

\usepackage[utf8]{inputenc}

\usepackage{microtype}

\usepackage{inconsolata}

\usepackage{graphicx}

\usepackage{subfigure}
\usepackage{amsmath}
\usepackage{listings}
\usepackage{tcolorbox}
\usepackage[table]{xcolor}
\usepackage{booktabs}
\usepackage{amssymb}
\usepackage{amsfonts}
\usepackage{arydshln}

\newcommand{\method}{\textsc{Uncode}}

\definecolor{mylightpurple}{RGB}{217,217,255}

\definecolor{PromptTitleBlue}{RGB}{20, 40, 120}
\definecolor{PromptContentBlue}{RGB}{235, 240, 255}

\title{Empirical Analysis of Decoding Biases in Masked Diffusion Models}

\author{Pengcheng Huang$^{1}$, Tianming Liu$^{2}$, Zhenghao Liu$^{1}$\thanks{ \ \ indicates corresponding author.}, \\ \textbf{Yukun Yan$^{3}$, Shuo Wang$^{3}$, Tong Xiao$^{1}$, Zulong Chen$^{4}$, Maosong Sun$^{3}$} \\ 
$^1$School of Computer Science and Engineering, Northeastern University, Shenyang, China \\
$^2$Software College, Northeastern University, Shenyang, China\\
$^3$Department of Computer Science and Technology, Tsinghua University, Beijing, China\\
$^4$Alibaba Group, Hangzhou, China \\
}

\begin{document}
\maketitle

\input{sections/0_abs}

\input{sections/1_intro}

\input{sections/2_related_work}

\input{sections/3_pre_ex}

\input{sections/4_method}

\input{sections/5_ex_setup}

\input{sections/6_res}

\input{sections/7_analysis}

\input{sections/8_conclusion}

\input{sections/9_limitations}


\bibliography{bib_new}
\appendix

\input{sections/9_append}

\end{document}

%% file: sections/0_abs.tex
\begin{abstract}

Masked Diffusion Models (MDMs) have recently emerged as a promising non-autoregressive paradigm for sequence generation. However, their performance is highly sensitive to the choice of decoding strategy. In this work, we reveal that prevalent uncertainty-based decoding strategies induce two decoding biases in MDMs: rigid boundary bias and trivial token bias. These biases limit the model's reasoning ability and ultimately degrade generation quality.
To address these challenges, we propose \textbf{UN}masking \textbf{C}alibration for Dec\textbf{O}ding \textbf{DE}biasing (\method{}), a decoding calibration framework that regularizes uncertainty-based decoding by incorporating two complementary priors to shape global decoding trajectories and promote content informativeness. Extensive experiments on three advanced MDMs across seven reasoning- and planning-intensive benchmarks demonstrate that \method{} consistently outperforms existing decoding strategies by more than 7\%, while achieving performance comparable to autoregressive models of similar parameter scales. The code is publicly available at \url{https://github.com/NEUIR/Uncode}.  

\end{abstract}

%% file: sections/1_intro.tex
\section{Introduction}
\label{sec:introduction}

Diffusion Large Language Models (dLLMs) have recently emerged as a competitive alternative to conventional autoregressive models (ARMs)~\cite{li2025survey,zhu2025llada}, offering non-autoregressive decoding that updates multiple tokens per step and excels at global reasoning tasks~\cite{li2023diffusion}. Among dLLMs, Masked Diffusion Models (MDMs) are particularly appealing due to their simple masked-denoising formulation and strong empirical performance~\cite{khanna2025mercury,nie2025large,dream2025}. Unlike autoregressive models that generate text strictly left-to-right, MDMs operate through iterative unmasking~\cite{wu2025fast,lou2023discrete}. This design removes sequential constraints, permitting tokens to be generated in a flexible order~\cite{kim2025train}. In practice, however, most existing MDMs employ uncertainty-based decoding strategies~\cite{chang2022maskgit,nie2025large}, which greedily unmask tokens at positions with the lowest predictive uncertainty during each step.
 
\begin{figure}[t!]
    \centering
    \includegraphics[width=\linewidth]{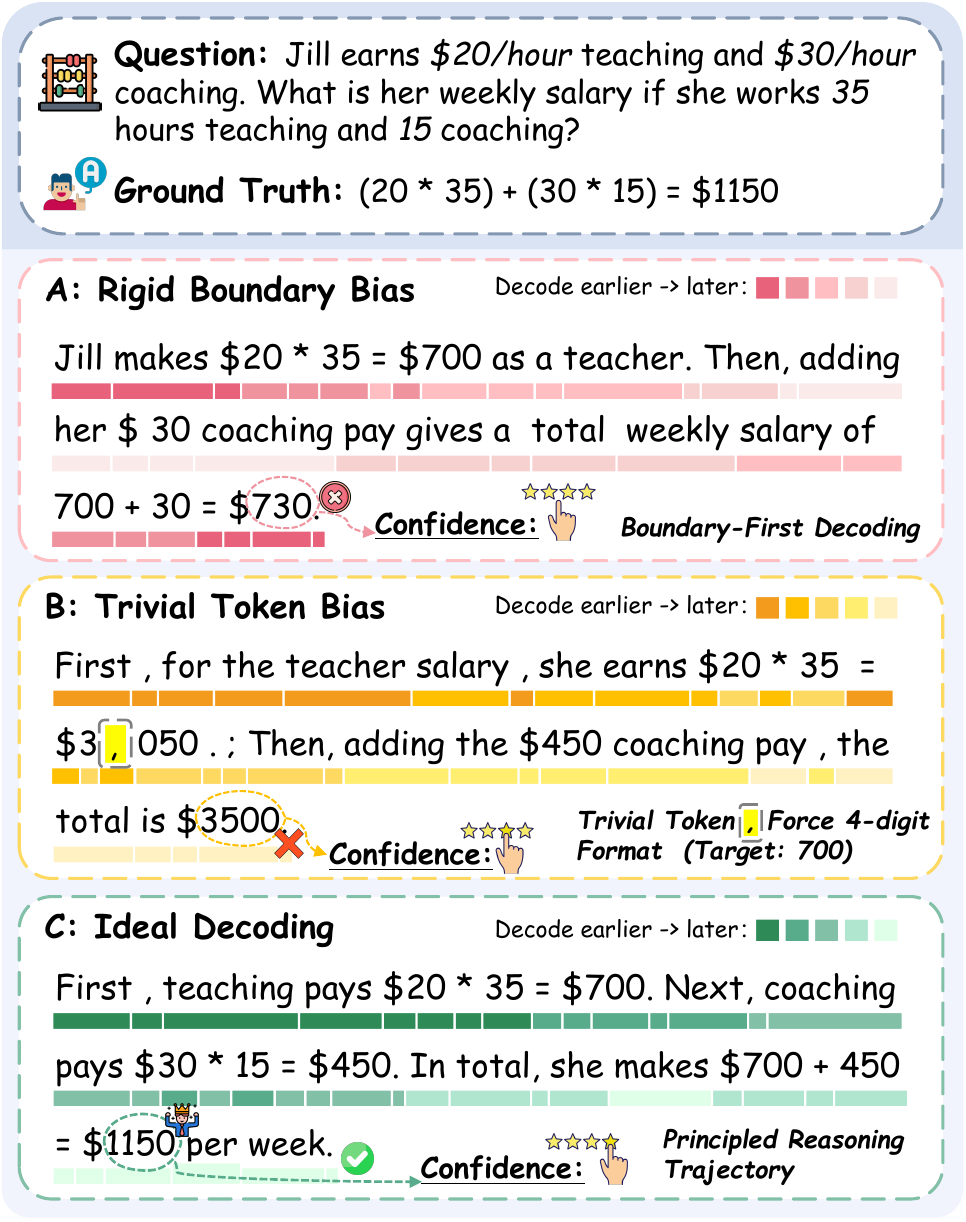}
    \caption{Illustration of decoding biases during reasoning. Uncertainty-based decoding prioritizes (A) rigid boundaries and (B) trivial tokens, causing systematic deviations from optimal problem-solving paths. In contrast, panel (C) presents the ideal unmasking pattern for producing coherent logical chains to achieve reliable inference. }
    \vspace{-1em}
    \label{fig:intro}
\end{figure}

Recent studies indicate that modern language models exhibit intrinsic ``reasoning trajectories'', where the generation follows a logical reason-then-answer sequence and model confidence in the correct answer progressively increases as intermediate steps are generated~\cite{wang2024chain}. In contrast, as illustrated in Figure~\ref{fig:intro}, we identify two structural biases in uncertainty-based decoding that hinder MDMs from following such trajectories and thereby prevent the formation of effective reasoning chains that are essential for complex problem-solving. 
Specifically, we observe: (1) \textit{Rigid Boundary Bias}, where the decoding trajectory follows a rigid pattern: boundary tokens are consistently decoded first, with updates progressively moving inward toward the center. This behavior prevents the model from adapting its decoding order to task-specific structural requirements and can cause answers to be resolved prematurely without establishing sufficient supporting rationale. (2) \textit{Trivial Token Bias}, where high-frequency but semantically trivial tokens (e.g., punctuation or structural markers) are disproportionately prioritized, hindering the formation of critical reasoning steps and ultimately leading to degraded performance.

To mitigate these biases, we propose \textbf{UN}masking \textbf{C}alibration for Dec\textbf{O}ding \textbf{DE}biasing (\method{}), a decoding calibration framework that calibrates uncertainty-based decoding for MDMs. Specifically, \method{} adjusts the raw uncertainty scores using two complementary priors. 
The first prior, the Positional Trajectory Prior, introduces a position-dependent reweighting to discourage a rigid boundary-first decoding pattern and allow the model to flexibly tailor its trajectory to diverse task structures. The second prior, the Semantic Informativeness Prior, modulates uncertainty scores based on corpus-level frequency statistics, penalizing high-frequency, semantically uninformative tokens to redirect the focus of refinement toward tokens that provide more informative content.

To validate the effectiveness of \method{}, we conduct extensive experiments on three representative MDMs across seven diverse benchmarks covering a broad range of reasoning and planning tasks. The results demonstrate that \method{} consistently improves overall generation performance, achieving an average gain of over 7\% compared to state-of-the-art decoding baselines, while achieving performance comparable to autoregressive models at a similar scale. By mitigating decoding biases, \method{} overcomes the limitations imposed by rigid boundary-first decoding in existing MDMs, enabling more effective construction of reasoning trajectories and reducing uncertainty in answer prediction, thereby leading to pronounced improvements. Furthermore, our analysis shows that the advantages of \method{} can be kept when integrated with different efficient sampling frameworks, achieving substantial speedups while simultaneously improving generation quality.

%% file: sections/2_related_work.tex
\section{Related Work}
\label{sec:related_work}
Masked Diffusion Models (MDMs) have emerged as a promising non-autoregressive alternative for text generation~\cite{song2025seed,nie2025large}. Unlike autoregressive models that generate sequences in a left-to-right order, MDMs utilize an iterative unmasking process, allowing tokens to be revealed in a non-sequential and arbitrary manner. This flexibility makes the decoding strategy pivotal to MDM performance~\cite{garg2025masked,kim2025train,lee2025lookahead}, as it directly controls the generation order and the effective context available at each step. As a result, numerous studies have explored how to leverage this flexibility to improve MDM performance~\cite{li2025survey}.


Early implementations of MDMs adopted a simple uniform sampling approach~\cite{NEURIPS2021_958c5305}, where tokens are randomly selected for unmasking at each step. While this strategy aligns with the order-agnostic training distribution~\cite{garg2025masked}, its unguided nature often leads to suboptimal generation quality~\cite{peng2025path}. To address this limitation, research has shifted towards uncertainty-driven heuristics that optimize the unmasking process by utilizing the model's intrinsic prediction signals~\cite{kim2025train,dream2025,ben2025accelerated}. Central to these strategies are three key metrics: confidence~\cite{chang2022maskgit}, which prioritizes tokens with the maximum predicted probability; margin~\cite{kim2025train}, which measures the probability gap between the top two candidates; and entropy~\cite{dream2025}, which quantifies overall predictive uncertainty. These approaches have been shown to improve stability and performance across a range of tasks~\cite{ben2025accelerated}.

Building upon these foundations, a variety of heuristics~\cite{wu2025fast,li2025beyond,yu2025dimple,wei2025accelerating,hong2025wide,horvitz2025no,israel2025accelerating} have further refined these frameworks to enable adaptive unmasking, achieving notable efficiency gains with minimal performance degradation. In parallel, semi-autoregressive approaches~\cite{arriola2025block,nie2025large} incorporate local causal dependencies into the unmasking process and often improve performance on logical reasoning tasks. Despite these advances, both lines of work still rely on uncertainty-based scoring to prioritize which positions to unmask. While prior research mainly optimizes \emph{how} uncertainty is used during decoding, we study systematic biases induced by this uncertainty-driven prioritization in MDMs, and identify two decoding biases inherent to uncertainty-based samplers.

%% file: sections/3_pre_ex.tex
\section{Motivating Analysis: Decoding Biases in Masked Diffusion Models}
\label{sec:preliminaries}

In this section, we examine two decoding biases that arise when uncertainty-based samplers are applied to MDMs: \textit{Rigid Boundary Bias} and \textit{Trivial Token Bias}. Our empirical analyses reveal that both biases systematically shape the decoding behavior of MDMs, suppressing the emergence of critical content and ultimately undermining the performance of MDMs across different tasks.

\textbf{Rigid Boundary Bias.}
Recent studies indicate that achieving strong performance with MDMs requires the model to use differentiated decoding orders tailored to each task~\cite{kim2025train}. Nevertheless, we observe that uncertainty-based samplers consistently induce a boundary-first decoding pattern, which we refer to as \textit{Rigid Boundary Bias}.

To illustrate this phenomenon, we conduct a series of controlled experiments, as shown in Figure~\ref{fig:boundary_bias}. We first visualize LLaDA-8B-Instruct's unmasking dynamics on the GSM8K reasoning benchmark using the confidence-based sampler in Figure~\ref{fig:sub_confidence}. Specifically, for each token position and decoding step, we track whether the token is unmasked (1 if unmasked, 0 otherwise) and report the average unmasking probability across all instances. The results reveal a pronounced pattern: tokens near the sequence boundaries are systematically prioritized early, yielding a characteristic ``U-shaped'' decoding trajectory. Notably, we observe this pattern consistently across diverse datasets and uncertainty-based decoding strategies (Appendix~\ref{appendix:u_shape}). As further analyzed in Appendix~\ref{appendix:boundary_bias_analysis}, this bias stems from the premature unmasking of boundary tokens (BOS and EOS), where the attention mechanism's local positional bias leads to elevated confidence for tokens near the sequence boundaries. This effect subsequently causes the unmasking process to move from both ends toward the center, resulting in a rigid boundary-first decoding pattern.

\input{figs/decode_order}

To better understand the impact of rigid boundary bias, we evaluate several decoding strategies on two representative datasets: GSM8K, which requires strict step-by-step sequential reasoning, and 4×4 Sudoku, a logic puzzle that benefits from globally coordinated planning. As shown in Figure~\ref{fig:sub_performance}, we compare three distinct strategies: the uniform sampler, the greedy confidence-based sampler, and the semi-autoregressive (Semi-AR) sampler.
The results reveal markedly different performance across both tasks. On GSM8K, the uncertainty-based sampler struggles, achieving only 59.1\% accuracy, substantially lower than the 77.9\% attained by Semi-AR decoding. Conversely, on Sudoku, the uncertainty-based sampler obtains the highest accuracy, outperforming both Semi-AR and uniform decoding. 
This task-dependent divergence suggests that while Semi-AR decoding facilitates sequential reasoning by introducing coarse-grained stepwise structure, uncertainty-based decoding may be advantageous for globally coordinated planning. These findings highlight the rigidity inherent in uncertainty-based decoding as a key challenge, emphasizing the need for decoding flexibility to fully unleash the reasoning capability of MDMs.

\textbf{Trivial Token Bias.}
We further observe that uncertainty-based samplers disproportionately prioritize semantically trivial, high-frequency tokens (e.g., \texttt{\textbackslash n}, \texttt{<space>}, \texttt{the}, \texttt{.}, \texttt{!}) throughout the decoding process. We refer to this phenomenon as \textit{Trivial Token Bias} (Figure~\ref{fig:trivial_tokens}).

We first quantify this bias by tracking the proportion of trivial tokens (definition provided in Appendix~\ref{appendix:trivial_tokens}) unmasked by the standard uncertainty-based (Confidence) sampler as decoding proceeds. As illustrated in Figure~\ref{fig:trivial_ratio}, trivial tokens consistently constitute nearly 40\% of the selections in MDMs. This significantly exceeds the $\sim$20\% baseline observed in ARMs. This striking discrepancy raises a key question: \textit{Is the over-selection of trivial tokens in MDMs necessary for effective problem-solving, or is it instead detrimental?}

\input{figs/trvial_tokens}

To investigate the impact of this bias, we conduct an intervention experiment on GSM8K by probabilistically suppressing the unmasking of trivial tokens. Specifically, at each decoding step, if the sampler selects a trivial token to unmask, we reject this choice with probability $p$ and instead re-sample from the subset of candidate positions whose predictions are non-trivial. As shown in Figure~\ref{fig:trivial_intervention}, the performance of MDMs improves monotonically as trivial-token selections are suppressed more frequently, indicating that over-selecting trivial tokens is not always necessary and can sometimes hinder reasoning. 
We hypothesize that trivial tokens are prioritized because greedy uncertainty-based decoding selects, at each step, the positions with the lowest uncertainty globally. Since trivial tokens (e.g., whitespace and punctuation) are typically easier to predict than content-bearing tokens, they are assigned lower uncertainty scores and are consequently unmasked earlier. Although such tokens carry little semantic information, revealing them early can impose surface-level structural commitments (e.g., punctuation and line breaks), which may reshape the uncertainty landscape and subsequent position-selection dynamics. This reduces flexibility in forming informative intermediate states and can lead to suboptimal reasoning trajectories. We provide a case study in Figure~\ref{tab:case_study_gsm8k_2} (Appendix~\ref{appendix:case_study}) to illustrate this phenomenon.


%% file: figs/decode_order.tex

\begin{figure}[t]
    \centering
    


    \subfigure[Heatmap of unmasking probabilities.]{
        \raisebox{-0.45em}{
            \includegraphics[width=0.469\linewidth]{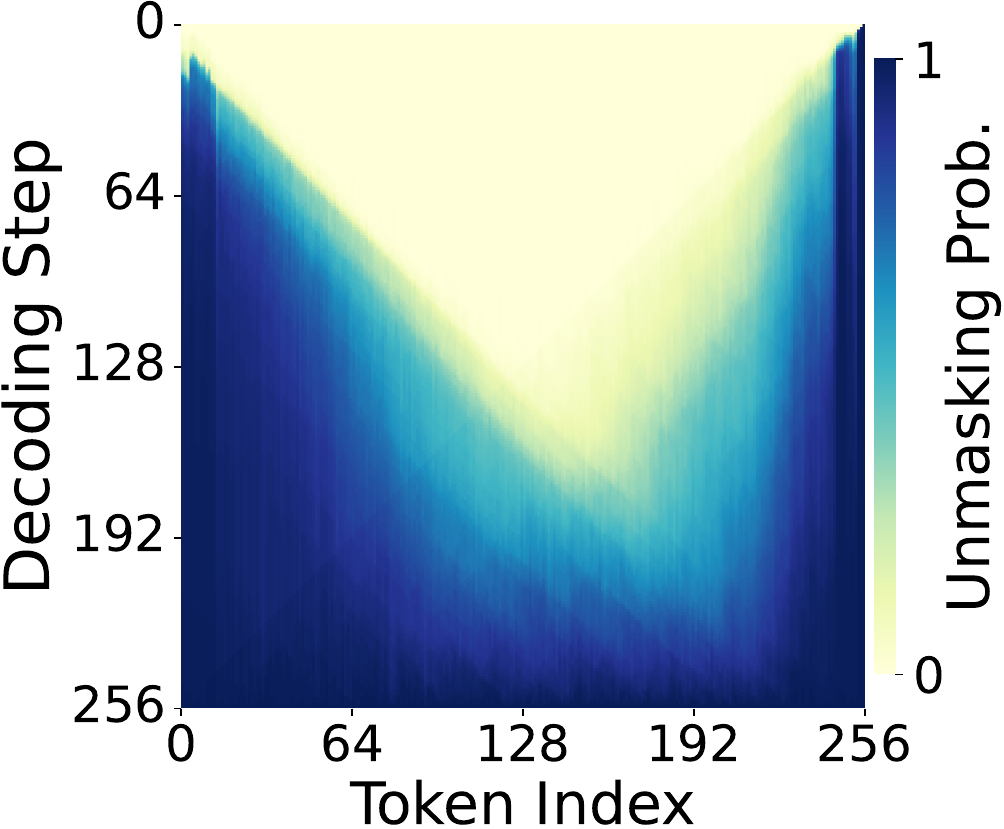}
        }
        \label{fig:sub_confidence}
    }%
    \hfill 
    \subfigure[Accuracy on reasoning vs. planning.]{
        \includegraphics[width=0.453\linewidth]{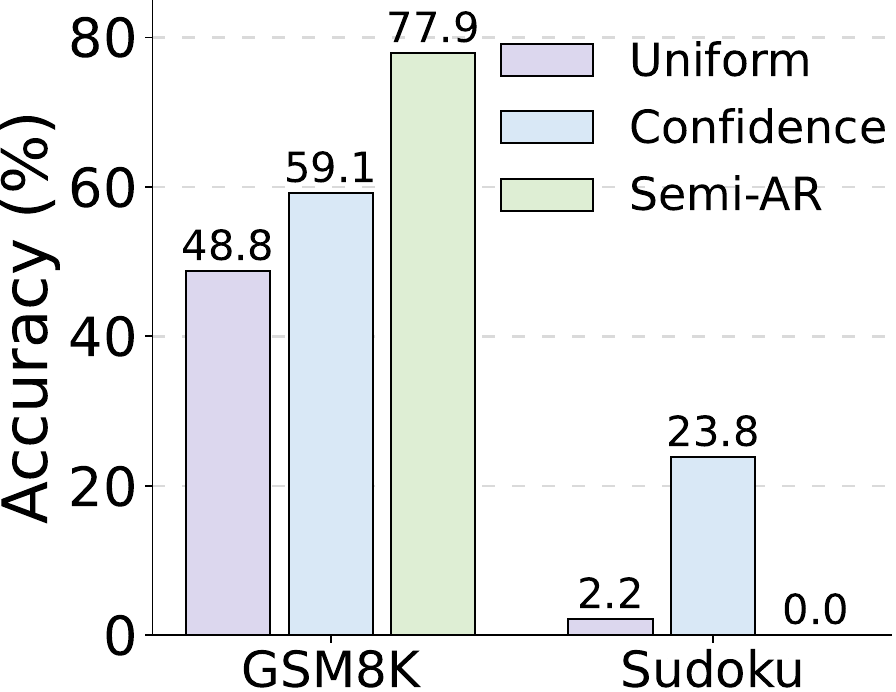}
        \label{fig:sub_performance}
    }
        
    \caption{Visualization of the Rigid Boundary Bias and its impact on downstream performance. (a) Unmasking probability for each token position across decoding steps on GSM8K, with both sequence length and decoding steps set to 256, where darker blue intensities denote higher unmasking probabilities. (b) Accuracy comparison of different decoding strategies on GSM8K (reasoning) and Sudoku (planning).}
    \label{fig:boundary_bias}
\end{figure}

%% file: figs/trvial_tokens.tex
\begin{figure}[t]
    \centering

    \subfigure[Over selection of trivial tokens.]{
    \raisebox{0.1em}{
            \includegraphics[width=0.46\linewidth]{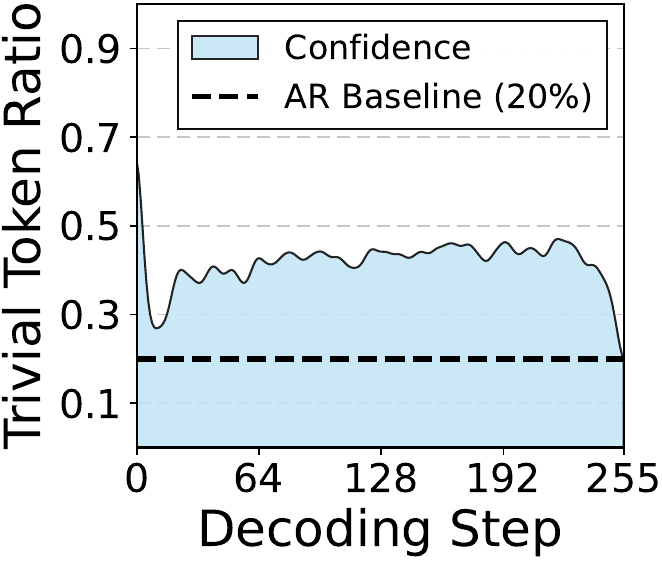}
        }
        \label{fig:trivial_ratio}
    }%
    \hfill
    \subfigure[Impact of trivial token suppression.]{
        \includegraphics[width=0.45\linewidth]{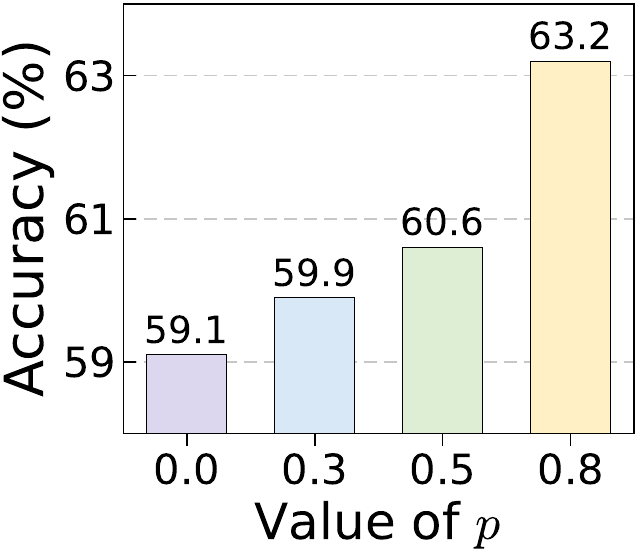}
        \label{fig:trivial_intervention}
    }
    \vspace{-0.5em}
    \caption{Verification of trivial token bias and trivial-token suppression efficacy. (a) The proportion of trivial tokens unmasked at each decoding step under the uncertainty-based sampler consistently exceeds the AR baseline (Qwen-2.5-7B-Instruct, dashed line). (b) GSM8K accuracy improves monotonically with suppression probability $p$. }
    \label{fig:trivial_tokens}
    \vspace{-1em}
\end{figure}



%% file: sections/4_method.tex
\section{Mitigating Decoding Bias with Unmasking Calibration}
\label{sec:methodology}
In this section, we briefly review the essential background of Masked Diffusion Models (MDMs) (\S\ref{subsec:background}), and then introduce \method{}, which addresses both the rigid boundary bias and the trivial-token bias in MDMs by calibrating the unmasking schedule during decoding (\S\ref{subsec:our_method}).

\subsection{Background of Masked Diffusion Models}
\label{subsec:background}

This section establishes the formal training objective of MDMs and describes the decoding paradigm that our method aims to optimize.

Unlike autoregressive models, which are trained via next-token prediction, MDMs learn to reconstruct original tokens from partially masked sequences. During training, a sequence $x_0$ is corrupted by independently masking tokens with a masking ratio $\tau \sim \mathcal{U}(0,1]$, yielding a partially masked sequence $x_\tau$. The model $p_\theta$ is trained to reconstruct the masked tokens by minimizing cross-entropy loss on the masked positions:
\begin{equation}
\small
    \mathcal{L}_{\text{MDM}} = - \mathbb{E}_{\tau, x_0, x_\tau} \left[ \frac{1}{\tau} \sum_{i=1}^{L} \mathbf{1}[x_\tau^i = \mathbf{M}] \log p_\theta(x_0^i \mid x_\tau)
    \right],
\end{equation}
where $x^i$ denotes the token at the $i$-th position, $\mathbf{M}$ denotes the mask token, and $\mathbf{1}[\cdot]$ is the indicator function. By sampling $\tau$ over a wide range, the model learns to handle various partial-context conditions, enabling it to predict any masked token from an arbitrary set of revealed tokens. This order-agnostic training naturally supports iterative, any-order decoding strategies at inference time.


During inference, the decoding process is formulated as iterative unmasking that gradually recovers a sequence from a fully masked state $x_T$ to a complete sequence $x_0$ using a predefined unmasking sampler. Specifically, let $\mathcal{M}_t$ denote the set of positions that remain masked at step $t$. The sampler first assigns an unmasking priority $s_t^i$ to all masked tokens. A subset $\mathcal{I}_t \subseteq \mathcal{M}_t$ is then selected for unmasking by identifying the $K$ positions within $\mathcal{M}_t$ that possess the highest priority scores:
\begin{equation}
\label{eq:selection}
\small
\mathcal{I}_t = \left\{ i \in \mathcal{M}_t \mid \text{rank}(s_t^i) \leq K \right\}.
\end{equation}
For each selected position $i \in \mathcal{I}_t$, the token $x^{i}_{t-1}$ is sampled according to the corresponding conditional probability distribution:
\begin{equation}
\label{eq:sampling}
\small
x^{i}_{t-1} \sim p_\theta(\cdot \mid x_t, i),
\end{equation}
and used to update the sequence, yielding a less-masked version, while the remaining positions stay masked. This procedure iterates until every position has been unmasked.
\subsection{Calibrating Decoding Biases in MDMs for Effective Inference}
\label{subsec:our_method}
Given the current set of remaining masked positions $\mathcal{M}_t$, existing MDMs typically use an uncertainty-based sampler by assigning a priority score to each masked position and unmasking the positions with the highest priority scores. Formally, at each step $t$, the sampler assigns a priority score $s_t^i$ to every masked position $i \in \mathcal{M}_t$ using a scoring function $\mathcal{F}(\cdot)$ (e.g., confidence or negative entropy) derived from the model's predictive distribution~\cite{chang2022maskgit,dream2025}:
\begin{equation}
\small
s_t^i = \mathcal{F}\!\left(p_\theta(\cdot \mid x_t, i)\right), \quad \forall i \in \mathcal{M}_t.
\end{equation}
However, this uncertainty-driven heuristic induces two systematic biases in MDM decoding: rigid boundary bias and trivial token bias (Section~\ref{sec:preliminaries}). To mitigate these biases, we propose \method{}, a lightweight calibration framework that adjusts the unmasking priorities by incorporating both a positional trajectory prior and a semantic informativeness prior. The calibrated unmasking priority is defined as:
\begin{equation}
\label{eq:score_function}
\small
\tilde{s}^{i}_t = \mathcal{P}^i \cdot \mathcal{S}^{i}_t \cdot \underbrace{\mathcal{F}\big(p_{\theta}(\cdot \mid x_t, i)\big)}_{\text{Raw Uncertainty Score}},
\end{equation}
where $\mathcal{P}^i$ and $\mathcal{S}^{i}_t$ are positive scalars that rescale the raw uncertainty score. Specifically, $\mathcal{P}^i$ is a \textit{positional trajectory prior} designed to mitigate rigid boundary-first decoding, while $\mathcal{S}^{i}_t$ is a \textit{semantic informativeness prior} aiming to reduce the dominance of trivial tokens during inference.

\textbf{Positional Trajectory Prior ($\mathcal{P}^i$).}
To mitigate the rigid boundary-first bias observed in uncertainty-based decoding, we introduce a positional trajectory prior $\mathcal{P}^i$ to modulate the spatial distribution of unmasking priorities, enabling consistent improvements across diverse tasks. This prior adjusts the unmasking priority of each position based on its absolute position, effectively reshaping the unmasking order. We instantiate $\mathcal{P}^i$ as a simple position-dependent decay function:
\begin{equation}
\small
\mathcal{P}^{i} = e^{-\lambda \cdot i},
\end{equation}
where $i$ denotes the token index and $\lambda$ controls the strength of positional regularization. When $\lambda \to 0$, the prior becomes uniform, recovering uncertainty-driven unmasking and permitting arbitrary generation trajectories. Conversely, larger values of $\lambda$ impose a stronger sequential bias, encouraging more left-to-right decoding. In this way, \method{} provides explicit control over the trade-off between non-autoregressive flexibility and causal dependency across diverse tasks. 

\textbf{Semantic Informativeness Prior ($\mathcal{S}^{i}_t$).}
The uncertainty-based decoding strategy usually prioritises high-frequency and semantically trivial tokens, thereby squandering valuable reasoning steps. To alleviate this effect, we introduce a semantic informativeness prior that reweights unmasking priorities based on token-level self-information~\cite{cover1999elements}.
Specifically, for each masked position $i$ at decoding step $t$, we consider the model's greedy predicted $\hat{x}_t^i = \arg\max_{v \in \mathcal{V}} p_\theta\!\left(v \mid x_t, i\right)$, where $v$ represents a candidate token from the vocabulary $\mathcal{V}$. We then assign an informativeness weight based on the token's prior frequency, estimated from a large-scale general corpus $\mathcal{D}'$:
\begin{equation}
\small
\mathcal{S}^{i}_t =  -\log p_{\mathcal{D}'}(\hat{x}^i_{t}). 
\end{equation}
This prior downweights frequent, low-information tokens and correspondingly increases the relative priority of semantically informative content, encouraging such tokens to be resolved earlier during decoding. To prevent excessively large weights for rare tokens, we follow~\citet{zhang2024pretraining} and clip the informativeness scores at a threshold $\alpha$, i.e.,
$\small \mathcal{S}_t^i \leftarrow \min(\mathcal{S}_t^i, \alpha)$.

%% file: sections/5_ex_setup.tex
\section{Experimental Methodology}
\label{sec:exp_methodology}
This section describes datasets, baselines, and implementation details used in our experiments.

\textbf{Datasets.}
We evaluate \method{} on seven benchmarks across four domains, selected to represent diverse structural dependencies: (1) \texttt{Mathematical reasoning.} To assess step-by-step logical deduction, we use GSM8K~\cite{cobbe2021training} and MATH500~\cite{lightman2023let}. (2) \texttt{Code generation.} Focusing on strict syntactic and causal dependencies, we evaluate on HumanEval~\cite{chen2021evaluating} and MBPP~\cite{austin2021program}. (3) \texttt{Scientific reasoning.} To test knowledge-intensive reasoning that combines parametric knowledge with logical deduction, we use GPQA~\cite{rein2024gpqa}. (4) \texttt{Planning.} For tasks prioritizing global constraint satisfaction, we include 4$\times$4 Sudoku~\cite{ye2024beyond} and Countdown~\cite{nolte2024transformers}. We report pass@1 for code generation and accuracy for all other tasks. Detailed dataset descriptions and statistics are provided in Appendix~\ref{appendix:dataset_descriptions}.

\textbf{Baselines.}
We compare \method{} against strong baselines from two families: (i) autoregressive models of comparable parameter scale, and (ii) representative decoding strategies for MDMs. To ensure a fair comparison, all MDM baselines are evaluated on the same backbone model under a matched decoding budget. Additional details about the baselines are provided in Appendix~\ref{appendix:baselines_details}.


\textit{Autoregressive baselines.}
We include three instruction-tuned ARMs of similar size: LLaMA-3.1-8B-Instruct~\cite{dubey2024llama}, Mistral-7B-Instruct~\cite{jiang2023mistral7b}, and Qwen-2.5-7B-Instruct~\cite{team2024qwen2}.

\textit{MDM decoding baselines.}
We further compare with several representative MDM decoding strategies:
(1) \texttt{Uniform}~\cite{NEURIPS2021_958c5305}, which selects masked positions uniformly at random.
(2) \texttt{Uncertainty-based} methods, which greedily rank masked positions by predicted uncertainty. We evaluate three representative proxies: maximum confidence~\cite{chang2022maskgit}, negative entropy~\cite{dream2025}, and margin~\cite{kim2025train}.
(3) \texttt{Semi-autoregressive} (Semi-AR) decoding~\cite{nie2025large}, which generates the sequence block by block. It enforces a global autoregressive order across blocks, while determining the unmasking positions within each block according to the MDM's prediction confidence.
(4) \texttt{Adaptive multi-token} samplers, which unmask multiple tokens per step using uncertainty thresholds to improve decoding efficiency without significantly compromising generation quality. We evaluate two representative methods: Entropy-Bounded (EB) Sampler~\cite{ben2025accelerated} and Fast-dLLM~\cite{wu2025fast}.

\input{tables/main_res}

\textbf{Implementation Details.}
We conduct experiments on three state-of-the-art masked diffusion language models: LLaDA-8B-Instruct~\cite{nie2025large}, LLaDA-1.5-8B~\cite{zhu2025llada}, and Dream-7B~\cite{dream2025}. During inference, we adopt the same generation length settings as in~\citet{nie2025large} and set the number of denoising steps to be equal to the sequence length. We also verify the generalization to longer sequences in Appendix~\ref{appendix:post_long_sequence}. We use prediction confidence as the base uncertainty proxy for calibration. For the semantic-informativeness prior, we estimate token-frequency statistics by fitting $p_{\mathcal{D}'}(\cdot)$ on a 16GB subset of the C4 corpus~\cite{raffel2020exploring}, following the protocol of~\citet{zhang2024pretraining} (see Appendix~\ref{appendix:effect_of_calibration_corpus} for an analysis of the calibration-corpus choice). We fix the clipping threshold to $\alpha = 10$ across all experiments. Regarding the positional-regularization coefficient $\lambda$, we set $\lambda=0$ for Sudoku to favor globally coordinated planning, $\lambda=0.25$ for the remaining tasks to encourage more progressive decoding, and $\lambda=0.5$ for Countdown. More implementation details are provided in Appendix~\ref{appendix:implementation_details}.

%% file: tables/main_res.tex
\begin{table*}[t]
\centering
\small 
\begin{tabular}{l cccccccc}
\toprule
\textbf{Methods \& LLMs} & \textbf{HumanEval} & \textbf{MBPP} & \textbf{GSM8K} & \textbf{MATH500} & \textbf{GPQA} & \textbf{Countdown} & \textbf{Sudoku} & \textbf{Avg.}\\
\midrule
\rowcolor{gray!15}
\multicolumn{9}{c}{\textbf{\textit{Autoregressive LLMs}}} \\
LLaMA-3.1-8B  & \underline{53.1} & \underline{56.7} & \textbf{83.9} & \underline{23.8} & \underline{31.0}  & \textbf{27.0} & 0.0 & \underline{39.4}\\
Mistral-7B  & 43.9 & 37.0 & 49.4 & 7.2 & 28.1  & \underline{22.7} & 0.0 & 26.9\\
Qwen-2.5-7B & \textbf{78.1} & \textbf{62.8} & \underline{71.9} & \textbf{64.2} & \textbf{32.8}  & 7.7 & 0.0 & \textbf{45.3}\\
\midrule
\rowcolor{gray!15}
\multicolumn{9}{c}{\textbf{\textit{LLaDA-8B-Instruct}}} \\
Uniform (\citeyear{NEURIPS2021_958c5305}) & 15.2 & 24.6 & 48.8 & 15.0 & 29.0  & 14.4 & 2.2 & 21.3\\
Confidence (\citeyear{chang2022maskgit}) & 27.4 & 42.4 & 59.1 & 20.8 & 27.9  & 34.0 & 23.8 & 33.6\\
Entropy (\citeyear{dream2025}) & 28.1 & 42.2 & 60.9 & 11.2 & 28.4  & 33.8 & 1.6 & 29.4\\
Margin (\citeyear{kim2025train}) & 32.3 & 42.4 & 58.3 & 19.8 & 28.4  & 33.9 & \underline{26.6} & 34.5\\
EB-Sampler (\citeyear{ben2025accelerated}) & 26.8 & 43.3 & 61.2 & 11.6 & \textbf{29.5}  & \underline{34.1} & 24.2 & 33.0\\
Semi-AR$^\dagger$ (\citeyear{nie2025large}) & \underline{39.0} & \underline{45.2} & 77.9 & 27.6 & 27.7  & 32.6 & 0.0 & \underline{35.7}\\
Fast-dLLM$^\dagger$ (\citeyear{wu2025fast}) & 35.4 & 44.7 & \underline{78.2} & \underline{28.4} & 28.6  & 23.6 & 0.0 & 34.1 \\
\rowcolor{blue!15}
\textbf{\method{}} & \textbf{42.1} & \textbf{47.8} & \textbf{79.2} & \textbf{34.8} & \underline{29.2} & \textbf{36.3} & \textbf{29.8} & \textbf{42.7} \\
\midrule
\rowcolor{gray!15}
\multicolumn{9}{c}{\textbf{\textit{LLaDA-1.5-8B}}} \\
Uniform (\citeyear{NEURIPS2021_958c5305}) & 17.7 & 23.0 & 52.7 & 20.0 & 28.1  & 15.8 & 3.4 & 23.0\\
Confidence (\citeyear{chang2022maskgit}) & 28.1 & 43.3 & 60.7 & 22.8 & \underline{28.7}  & 33.8 & 24.8 & 34.6\\
Entropy (\citeyear{dream2025}) & 32.9 & 44.0 & 60.3 & 11.2 & 26.6  & \underline{34.7} & 0.2 & 30.0\\
Margin (\citeyear{kim2025train}) & 25.0 & 43.3 & 57.5 & 23.2 & 28.4  & 31.8 & \textbf{33.6} & 34.7\\
EB-Sampler (\citeyear{ben2025accelerated}) & 32.9 & 43.6 & 61.1 & 13.4 & 26.6  & 34.6 & 0.2 & 30.3\\
Semi-AR$^\dagger$ (\citeyear{nie2025large}) & \underline{39.6} & \underline{46.8} & 80.7 & \underline{34.2} & 26.1  & 32.4 & 0.0 & \underline{37.1}\\
Fast-dLLM$^\dagger$ (\citeyear{wu2025fast}) & 37.2 & 46.1 & \underline{80.8} & 31.2 & 27.9  & 23.6 & 0.0 & 36.7\\
\rowcolor{blue!15}
\textbf{\method{}} & \textbf{46.3} & \textbf{49.9} & \textbf{82.2} & \textbf{37.4} & \textbf{28.8} & \textbf{35.0} & \underline{33.4} & \textbf{44.7}\\
\bottomrule
\end{tabular}
\caption{
Experimental results on seven different benchmarks. The best score is highlighted in \textbf{bold}, and the second-best is \underline{underlined}. Following prior practices~\cite{nie2025large}, we use 4-shot settings for GSM8K and MATH500, 5-shot for GPQA and Sudoku, 0-shot for HumanEval and MBPP, and 3-shot for Countdown. Methods marked with $^\dagger$ denote samplers using the Semi-AR strategy. Results on Dream-7B are provided in Table~\ref{tab:main_results_revised_with_dream} (Appendix~\ref{appendix:dream_dream}). }
\vspace{-0.9em}
\label{tab:main_results_revised}
\end{table*}

%% file: sections/6_res.tex
\section{Experiment Results}
\label{sec:results}
In this section, we first present the overall performance of \method{}, followed by ablation results. Subsequently, we analyze the decoding dynamics of \method{} and assess its compatibility with efficient decoding strategies.

\subsection{Overall Performance}
We present the main results of \method{} across all benchmarks in Table~\ref{tab:main_results_revised}.

Overall, \method{} consistently outperforms all baseline decoding strategies, achieving over a 7\% average performance improvement. Notably, this advantage generalizes well across different backbone models and task categories, with \method{} achieving near-best performance among all baselines across the seven benchmarks. In contrast, existing decoding methods exhibit pronounced task specialization. For example, uncertainty-based samplers perform well on planning tasks such as Sudoku, outperforming the vanilla uniform decoding strategy by over 20\%, but tend to underperform on reasoning-intensive tasks, including code generation and mathematical reasoning. Semi-AR methods, which introduce a degree of left-to-right constraint on top of confidence-based sampling, substantially improve reasoning performance; for instance, achieving over 10\% average gains on HumanEval and GSM8K compared to the confidence sampler. However, these gains do not transfer to planning tasks, where the imposed sequential bias becomes restrictive. 

By explicitly controlling the generation order and suppressing trivial token selections, \method{} enables MDMs to achieve strong and stable performance across both reasoning and planning tasks. On LLaDA-1.5-8B, it even achieves comparable overall performance to Qwen-2.5-7B-Instruct. These results suggest that decoding-side improvements alone can better unlock the potential of MDMs, enabling performance that is comparable to that of autoregressive models.

%% file: sections/7_analysis.tex
\input{figs/ablation}

\subsection{Ablation Studies}
\label{ablation}
We conduct ablation studies to verify the effectiveness of \method{}; the results are shown in Figure~\ref{fig:ablation_results}.

To understand the significance of the positional trajectory prior (Pos. Prior), we compare \method{} with models that are solely calibrated with the Semantic Informativeness Prior. As shown in Figure~\ref{fig:ablation_results}, omitting the positional trajectory prior leads to a substantial decrease in performance (42.7\% $\rightarrow$ 34.9\% for LLaDA and 44.7\% $\rightarrow$ 35.5\% for LLaDA-1.5, respectively). This demonstrates the critical role of flexibly regulating the decoding order in breaking rigid boundary-first trajectory and fully harnessing the potential of MDMs in various practical scenarios. To further verify that \method{} genuinely leverages this non-autoregressive flexibility rather than degenerating into autoregressive-like generation, we provide additional comparisons against strict fixed-order baselines in Appendix~\ref{appendix:post_baseline_disentanglement}. Complementing the positional adjustment, we evaluate the importance of the Semantic Informativeness Prior (Sem. Prior) by comparing \method{} with a model calibrated exclusively by the positional prior. As shown in Figure~\ref{fig:ablation_results}, introducing the Sem. Prior further improves average performance by 2.3\% and 2.7\% in LLaDA and LLaDA-1.5, respectively. This demonstrates that the semantic informativeness prior, by mitigating the over-selection of trivial tokens, more effectively enhances MDM performance.


\subsection{Further Analysis: Decoding Dynamics and Efficiency of \method{}}
\label{subsec:further_analysis}
In this section, we first delve into the internal mechanisms driving the superior performance of \method{}, and subsequently demonstrate its compatibility with efficient decoding strategies.

\input{figs/compare}
\input{tables/efficient_ex}

\textbf{Decoding Dynamics of \method{}.}
As illustrated in Figure~\ref{fig:analysis_entropy}, we conduct experiments to investigate the effectiveness of \method{} in regulating the generation dynamics of MDMs.

To assess whether the model adheres to a logical ``reasoning-then-answer'' generation order, we examine answer token dynamics by tracking their unmasking steps and predictive entropies. As shown in Figure~\ref{fig:analysis_answer_entropy}, the confidence sampler exhibits a distinct pattern of premature unmasking, where answer tokens appear at early steps despite having high entropy. This suggests the model is attempting to ``guess'' the answer without sufficient context. This phenomenon is driven by a rigid boundary bias, which compels the model to follow a ``U-shaped'' trajectory that forces answer prediction before the reasoning chain is fully constructed. Conversely, \method{} leverages the positional trajectory prior to break this constraint, effectively deferring answer revelation until the global context is established, yielding high-confidence predictions grounded in sufficient reasoning.

We further evaluate the efficiency of global context construction by tracking the average predictive entropy of all remaining masked tokens. As shown in Figure~\ref{fig:analysis_predictive_entropy}, the confidence sampler maintains a consistently higher level of entropy throughout the process. This suggests that its tendency to over-select trivial tokens causes the model to remain uncertain about the global context throughout the decoding process. In contrast, \method{} consistently achieves lower predictive entropy, validating the effectiveness of our Semantic Informativeness Prior. By redirecting focus toward semantically rich tokens, \method{} allows the model to rapidly establish a robust context, guiding effective reasoning for the subsequent generation.

\textbf{Integrating \method{} with Efficient Decoding Strategies.}
Furthermore, we evaluate the potential of \method{} to serve as a plug-and-play unmasking policy within existing efficient decoding frameworks, as inference efficiency is a critical prerequisite for the practical deployment of MDMs~\cite{israel2025accelerating}. Our evaluation encompasses three representative paradigms: the deterministic $\tau$-leaping~\cite{chen2023accelerating}, the adaptive EB-Sampler~\cite{ben2025accelerated}, and Fast-dLLM~\cite{wu2025fast} (with KV cache optimization). Detailed implementations are provided in Appendix~\ref{appendix:integrating_w_efficient}. 
As demonstrated in Table~\ref{tab:efficient_ex}, \method{} seamlessly integrates with these efficient decoding frameworks, consistently outperforming their native policies across most benchmarks, yielding an average performance improvement of over 3\%. Crucially, the integrated variants preserve the efficiency benefits, achieving inference speeds exceeding $2\times$ that of vanilla decoding.

%% file: figs/ablation.tex
\begin{figure}[t]
    \centering
    \includegraphics[width=0.91\columnwidth]{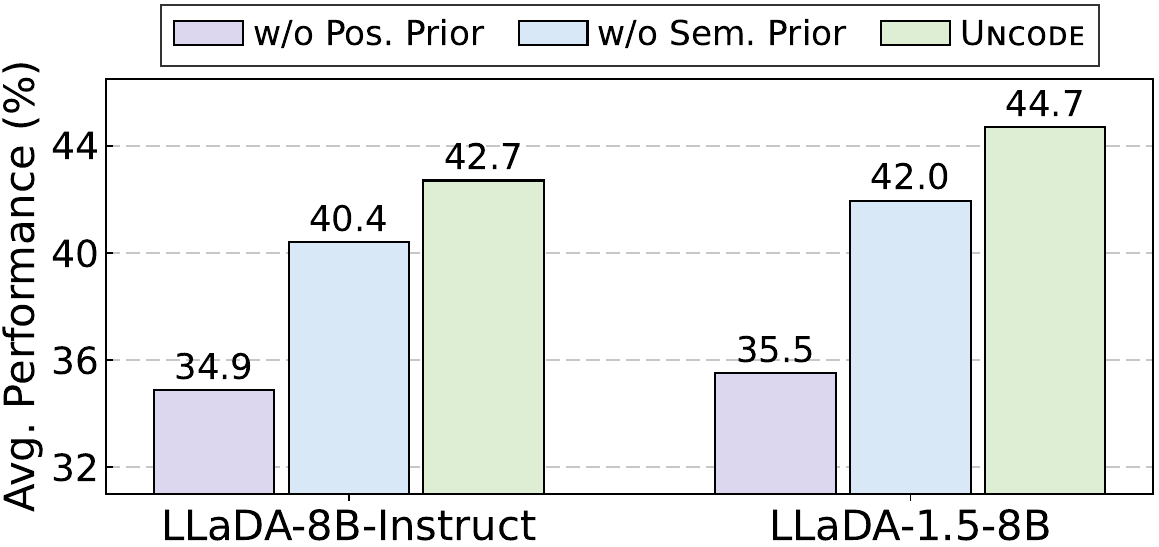}
    \vspace{-0.5em}
    \caption{Ablation results of individual modules on LLaDA-8B-Instruct and LLaDA-1.5-8B, reporting the average performance across all evaluation benchmarks. }
    \label{fig:ablation_results}
\end{figure}


%% file: figs/compare.tex
\begin{figure}[t]
    \centering
    \subfigure[Distribution of answer token entropy.]{
        \includegraphics[width=0.463\columnwidth]{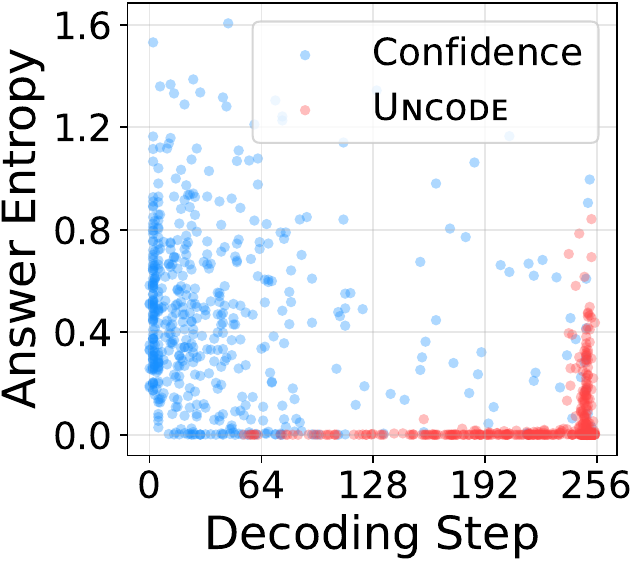}
        \label{fig:analysis_answer_entropy}
    }
    \hfill
    \subfigure[Evolution of predictive uncertainty.]{
        \includegraphics[width=0.463\columnwidth]{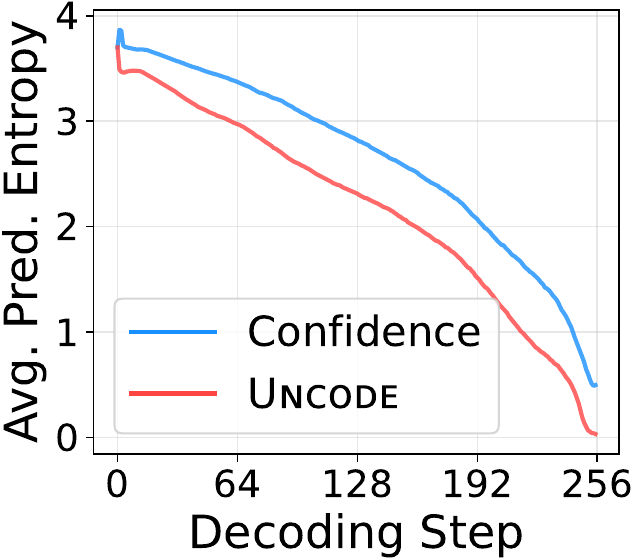}
        \label{fig:analysis_predictive_entropy}
    }
    \vspace{-0.5em}
    \caption{Analysis of uncertainty dynamics during decoding. (a) Answer Entropy vs. Decoding Step. While the baseline (Confidence, blue) often prematurely unmasks answer tokens with high uncertainty, \method{} (red) delays these answer tokens to later steps (bottom-right cluster), ensuring they are generated with high confidence. (b) \method{} achieves faster global uncertainty reduction than the baseline.}
    \vspace{-1em}
    \label{fig:analysis_entropy}
\end{figure}


%% file: tables/efficient_ex.tex
\setlength{\tabcolsep}{5.3pt} 

\begin{table*}[t]
\centering
\small 
\begin{tabular}{l cccccccc}
\toprule
\textbf{Methods} & \textbf{HumanEval} & \textbf{MBPP} & \textbf{GSM8K} & \textbf{MATH500} & \textbf{GPQA} & \textbf{Countdown} & \textbf{Sudoku} & \textbf{Avg.}\\
\midrule
$\tau$-leaping (\citeyear{chen2023accelerating}) & 17.7 & 38.9 & 55.2 & 16.2 & 28.9  & 32.1 & 3.2 & 27.5 {\scriptsize (2.14x)} \\
\rowcolor{blue!15}
\; + \method{} & 22.6 & 40.5 & 75.4 & 30.8 & 29.5  & 28.4 & 25.4 & 36.1 {\scriptsize (1.99x)}\\
\midrule
EB-sampler (\citeyear{ben2025accelerated}) & 26.8 & 43.3 & 61.2 & 11.6 & 29.5  & 34.1 & 24.2 & 33.0 {\scriptsize (2.32x)}\\
\rowcolor{blue!15}
\; + \method{} & 41.5 & 46.6 & 79.3 & 35.2 & 28.4  & 36.2 & 25.6 & 41.8 {\scriptsize (2.28x)}\\
\midrule
Fast-dLLM (\citeyear{wu2025fast}) & 35.0 & 45.9 & 77.4 & 25.2 & 27.8  & 24.4 & 0.0 & 33.9 {\scriptsize (4.21x)}\\
\rowcolor{blue!15}
\; + \method{} & 36.0 & 48.2 & 77.8 & 29.2 & 28.4  & 36.3 & 0.6 & 36.7 {\scriptsize (3.99x)}\\
\bottomrule
\end{tabular}
\caption{Performance and efficiency analysis of \method{} with integrated efficient decoding strategies. The last column reports the average performance, with the speedup factor relative to vanilla decoding shown in parentheses.}
\vspace{-1em}
\label{tab:efficient_ex}
\end{table*}


%% file: sections/8_conclusion.tex
\section{Conclusion}

In this work, we identify two decoding biases that arise when uncertainty-based samplers are applied to Masked Diffusion Models (MDMs): rigid boundary bias and trivial token bias, both of which limit the performance potential of MDMs. To address these challenges, we propose \method{}, a decoding framework that incorporates a Positional Trajectory Prior and a Semantic Informativeness Prior to mitigate these biases. Extensive experiments on three advanced MDMs across diverse benchmarks demonstrate that \method{} consistently outperforms existing methods, achieving significant improvements in generation quality and comparable performance to autoregressive models of similar size. Additionally, we demonstrate that \method{} is compatible with current efficient decoding frameworks, enabling high-quality, accelerated inference without the need for additional training.



%% file: sections/9_limitations.tex
\section*{Limitations}
While \method{} achieves promising results, it introduces a weight $\lambda$ for the Positional Trajectory Prior. Although a lightweight search over a small set of candidates (e.g., four values) proves effective, it requires ground-truth labels for validation, which limits its applicability in fully unsupervised settings. To address this, we introduce an automated, adaptive strategy in Appendix~\ref{appendix:adaptive_lambd}, eliminating the dependency on annotated data. Future work may further replace this heuristic with a learned controller.
Next, although \method{} integrates seamlessly with efficient decoding strategies, computing the Positional Trajectory and Semantic Informativeness priors introduces a small per-step overhead. However, as detailed in Appendix~\ref{appendix:post_efficiency}, our empirical results indicate that this cost is negligible relative to the model's forward pass, allowing the overall inference efficiency to remain comparable to standard decoding baselines.

\section*{Acknowledgments}
This work is partly supported by the National Natural Science Foundation of China (No. 62576082) and the Alibaba Innovative Research Program. This work is also supported by the AI9Stars community.

%% file: sections/9_append.tex
\clearpage
\appendix
\section{Appendix}

\subsection{License}
The licenses for the datasets used in this study are as follows: HumanEval, MBPP, GSM8K, and MATH500 are released under the MIT License; GPQA is released under the CC BY 4.0 License; and Countdown and Sudoku are released under the Apache License 2.0.

\subsection{In-Depth Examination of the Rigid Boundary Bias Phenomenon}
\label{appendix:boundary_bias_analysis}
In Section~\ref{sec:preliminaries}, we observe that uncertainty-based samplers induce a rigid boundary bias in the decoding behavior of MDMs. Building on this empirical observation, we now examine why uncertainty-based decoding, when applied to MDMs, so consistently gives rise to a boundary-first trajectory. Our analysis identifies two interacting factors that jointly contribute to this phenomenon: the high predictability of boundary tokens and inherent local positional biases in the model.

The first factor arises from the use of fixed training templates in MDMs. Training sequences are typically augmented with explicit boundary tokens such as \texttt{<bos>} and \texttt{<eos>}, which appear at highly regular positions and with high frequency throughout the training corpus. This strong positional regularity makes boundary tokens structurally predictable, leading the model to assign them high confidence during inference. Under uncertainty-based decoding, these high-confidence boundary positions are therefore selected for unmasking at the earliest decoding steps, causing both ends of the sequence to be fixed first and resulting in a boundary-first decoding pattern.

Second, once boundary tokens are fixed, inherent local positional biases in the model’s attention mechanism~\cite{wu2025emergence,jiang2025d} make it easier for the model to assign higher confidence to positions near already revealed tokens than on those farther away. As a result, the decoding process tends to propagate inward from the confirmed boundaries toward the center of the sequence, producing a characteristic ``U-shaped'' decoding trajectory. To test this hypothesis, we conduct a controlled intervention experiment on 100 GSM8K examples using confidence-based decoding. Before decoding begins, we manually unmask a single token at a specified relative position (25\% or 75\% of the sequence length), initializing it as a trivial token (e.g., a space). We then run the standard decoding procedure and visualize the resulting unmasking dynamics. As shown in Figure~\ref{fig:interface_ex}, the artificially inserted token consistently acts as a new anchor in the decoding trajectory. During the earliest decoding steps, unmasking probabilities peak at anchor positions, including the sequence boundaries and the inserted token, and subsequently spread to neighboring positions. The same qualitative behavior is observed when other frequent tokens, such as ``the'' or ``.'', are used as anchors.

\input{appendix_figs/interface_ex}

\subsection{Unmasking Dynamics of Uncertainty-Based Samplers Across Datasets}
\label{appendix:u_shape}
Following our observation of the rigid boundary bias on GSM8K using the confidence-based sampler in Section~\ref{sec:preliminaries}, we now systematically assess its pervasiveness across uncertainty-based decoding in MDMs. Specifically, we extend our visualization of unmasking dynamics to all benchmarks employed in our main experiments, examining three common uncertainty-based sampling criteria: confidence-based, entropy-based, and margin-based samplers. The resulting heatmaps for GSM8K (Figure~\ref{fig:gsm8k_method_comparison}), MBPP (Figure~\ref{fig:mbpp_method_comparison}), HumanEval (Figure~\ref{fig:humaneval_method_comparison}), GPQA (Figure~\ref{fig:gpqa_method_comparison}), Countdown (Figure~\ref{fig:countdown_method_comparison}), and Sudoku (Figure~\ref{fig:sudoku_method_comparison}) consistently exhibit a boundary-first decoding pattern, regardless of the specific uncertainty metric or task domain.

\input{appendix_figs/entropy_and_margin_GSM8K}
\input{appendix_figs/entropy_and_margin_MBPP}
\input{appendix_figs/entropy_and_margin_humaneval}
\input{appendix_figs/entropy_and_margin_GPQA}
\input{appendix_figs/entropy_and_margin_countdown}
\input{appendix_figs/entropy_and_margin_sudoku}

Furthermore, this uniformity suggests that the bias is not an artifact of a specific calculation method (e.g., confidence vs. entropy) but rather stems from the underlying probability landscape of the model itself. Since MDMs are typically trained to predict tokens in any order, they tend to exhibit high confidence (low entropy/ large margin) for syntactically deterministic tokens, such as punctuation or structural boundaries, while remaining uncertain about the semantically dense tokens required for reasoning. Consequently, all uncertainty-based metrics inevitably prioritize these ``easy'' tokens, leading to the observed premature unmasking of boundary and other easy tokens.

Beyond the decoding dynamics themselves, this bias also has clear performance implications.
Across reasoning-intensive tasks such as code generation and mathematical problem solving, uncertainty-based samplers underperform semi-autoregressive methods by over 5 points on average. In contrast, on globally constrained planning tasks such as Countdown and Sudoku, uncertainty-based samplers outperform semi-autoregressive decoding, where strict left-to-right constraints prove limiting. These contrasting behaviors further highlight the central role of decoding order in determining MDM performance. At the same time, they reveal a key limitation of existing approaches: both uncertainty-based samplers and semi-autoregressive methods impose a single, fixed decoding trajectory, which restricts the ability of MDMs to adapt their generation order to the structural demands of different tasks.

%

\subsection{Definition of Trivial Tokens}
\label{appendix:trivial_tokens}
For our analysis, we define trivial tokens as semantically uninformative, high-frequency tokens that appear frequently in the corpus but contribute little meaningful content, following the protocol established by ~\citet{martinez2024mitigating}. These tokens primarily consist of common structural symbols, punctuation marks, and a small set of function or filler words. Notably, although \texttt{<|endoftext|>} appears frequently in model outputs, we exclude it from this definition, as MDMs typically utilize it as a functional padding token during instruction tuning~\cite{nie2025large}. We intentionally adopt a conservative definition, restricting our scope to tokens that are unambiguously structural or uninformative, to avoid conflating triviality with task-relevant function words. The complete list of trivial tokens used in our study is provided below. Note that this definition of ``trivial tokens'' is solely used for the \textit{empirical analysis} in Section~\ref{sec:preliminaries} to demonstrate the existence of the trivial-token bias in MDMs.

\definecolor{MyBoxTitleBlue}{RGB}{20, 40, 120}
\definecolor{MyBoxContentBlue}{RGB}{235, 240, 255}
\definecolor{TokenBlue}{RGB}{205, 220, 245} 

\begin{tcolorbox}[
    colback=MyBoxContentBlue,
    colframe=MyBoxTitleBlue,
    fonttitle=\bfseries,
    colbacktitle=MyBoxTitleBlue,
    coltitle=white,
    arc=3mm, 
    title=List of Trivial Tokens
]
{%
\setlength{\fboxsep}{3pt} 
\colorbox{TokenBlue}{\texttt{<SPACE>}} \colorbox{TokenBlue}{\texttt{\textbackslash n}} \colorbox{TokenBlue}{\texttt{\textbackslash t}} \colorbox{TokenBlue}{\texttt{.}} \colorbox{TokenBlue}{\texttt{,}} \colorbox{TokenBlue}{\texttt{?}} \colorbox{TokenBlue}{\texttt{!}} \colorbox{TokenBlue}{\texttt{:}} \colorbox{TokenBlue}{\texttt{;}} \colorbox{TokenBlue}{\texttt{\textbackslash
}} \colorbox{TokenBlue}{\texttt{"}} \colorbox{TokenBlue}{\texttt{'}} \colorbox{TokenBlue}{\texttt{\$}} \colorbox{TokenBlue}{\texttt{is}} \colorbox{TokenBlue}{\texttt{the}} \colorbox{TokenBlue}{\texttt{of}} \colorbox{TokenBlue}{\texttt{a}} \colorbox{TokenBlue}{\texttt{an}} 
}
\end{tcolorbox}

To verify that the observed trivial-token bias is a robust phenomenon and is insensitive to the specific definition of trivial tokens, we analyze the generation behavior under two extreme definitions: (1) expanding the list to include all common NLTK stopwords, and (2) restricting the list to only whitespace tokens. As shown in Table~\ref{tab:post_trivial_definition}, while the absolute percentages vary depending on the definition, the core conclusion remains robust: MDMs consistently generate a significantly higher proportion of low-information tokens compared to AR models across all definitions.

\begin{table}[h]
\centering
\small
\begin{tabular}{lcc}
\toprule
\textbf{Definition} & \textbf{MDM} & \textbf{AR} \\
\midrule
Conservative definition (Ours) & 41.7\% & 20.9\% \\
Expanded w/ NLTK stopwords & 49.0\% & 33.7\% \\
Whitespace-only & 8.9\% & 7.2\% \\
\bottomrule
\end{tabular}
\caption{Robustness of the trivial-token bias observation across different definitions. Values denote the average ratio of trivial tokens generated by MDM and AR.}
\label{tab:post_trivial_definition}
\end{table}



\subsection{Dataset Descriptions and Statistics}
\label{appendix:dataset_descriptions}
We evaluate \method{} on a diverse set of benchmarks covering code generation, mathematical reasoning, scientific question answering, and planning tasks. Dataset statistics are summarized in Table~\ref{tab:dataset_stats}.

\textbf{Code Generation.}
HumanEval~\cite{chen2021evaluating} is a widely used benchmark for Python code generation, where models are required to complete function implementations from natural language descriptions and are evaluated using pass@1 based on unit tests. MBPP (Mostly Basic Programming Problems)~\cite{austin2021program} evaluates code generation on a diverse collection of entry-level programming tasks, also assessed using pass@1.

\textbf{Mathematical Reasoning.}
GSM8K~\cite{cobbe2021training} consists of grade-school-level math word problems that require multi-step numerical reasoning, and performance is measured by exact-match accuracy on the final answer. MATH500~\cite{lightman2023let} is a curated subset of the MATH dataset containing challenging competition-level problems, evaluating advanced mathematical reasoning via accuracy.

\textbf{Scientific Question Answering.}
GPQA~\cite{rein2024gpqa} is a scientific QA benchmark covering biology, physics, and chemistry, designed to test domain knowledge and multi-step reasoning, with accuracy as the evaluation metric.

\textbf{Planning Tasks.}
Countdown~\cite{nolte2024transformers} is a planning-oriented task where models must construct a sequence of arithmetic operations to reach a target number from a given set of numbers. Sudoku~\cite{ye2024beyond} evaluates global constraint satisfaction by requiring models to complete partially filled Sudoku grids. Performance on both planning tasks is measured by accuracy.

\subsection{Details of Implementation}
\label{appendix:implementation_details}
In this subsection, we provide a detailed description of our experimental setup (\S\ref{sec:exp_methodology}) and the implementation specifics of our proposed \method{}.

All experiments were conducted on four NVIDIA A100 GPUs with 80GB of memory. To ensure reproducibility, we use a fixed random seed of 42. During inference, the temperature is set to 0.0 to ensure deterministic generation. Following~\citet{nie2025large}, we set the maximum response length to 256 tokens for HumanEval, GSM8K, and GPQA; 128 tokens for MBPP, Countdown, and Sudoku; and 1024 tokens for MATH500. For the positional regularization coefficient $\lambda$, we consider a small set of fixed values $\{0, 0.25, 0.5, 1.0\}$ to control the degree of sequential bias in decoding. When $\lambda = 0$, no positional trajectory prior is applied, resulting in fully order-agnostic decoding. Empirically, setting $\lambda = 1.0$ yields behavior that closely resembles strictly left-to-right decoding. Based on these observations, we set $\lambda = 0$ for Sudoku to favor globally coordinated planning, $\lambda = 0.25$ for the remaining tasks to encourage more progressive decoding, and $\lambda = 0.5$ for Countdown.
All $\lambda$ values are fixed across backbone models.

\input{tables/data_statistics}

\textbf{Evaluation Framework.}
To ensure a fair and rigorous comparison, all baseline methods were evaluated using the same evaluation procedures. For GSM8K and GPQA, we employed the lm-evaluation-harness framework~\cite{eval-harness}. For other tasks requiring specialized metrics or formats, we adopted publicly available or officially recommended evaluation scripts, following~\cite{nie2025large,zhao2025d1,wang2025mro}.

\textbf{Prompting Strategies.}
For GSM8K, GPQA, and MATH500, we follow~\citet{nie2025large} and use the standard prompts provided in the lm-evaluation-harness framework. For Sudoku, we follow the evaluation protocol of~\citet{zhao2025d1} and use their corresponding prompt.
For HumanEval and MBPP, since~\citet{nie2025large} does not release the corresponding prompt templates, we adopt the prompt format from~\citet{wang2025mro}. The prompt is shown below:
\begin{tcolorbox}[
    colback=PromptContentBlue,
    colframe=PromptTitleBlue,
    fonttitle=\bfseries,
    colbacktitle=PromptTitleBlue,
    coltitle=white,
    arc=2mm,
    title={Prompt for HumanEval and MBPP}
]
\textbf{Role}: You are a professional Python coding assistant \\
\textbf{Task}: Complete the following function implementation strictly and clearly without any additional comments or explanations.
\end{tcolorbox}

For the Countdown task, we follow~\citet{wang2025mro} and adopt a prompt that explicitly instructs the model to construct a sequence of arithmetic operations leading to the target number, as shown below.
\begin{tcolorbox}[
    colback=PromptContentBlue,
    colframe=PromptTitleBlue,
    fonttitle=\bfseries,
    colbacktitle=PromptTitleBlue,
    coltitle=white,
    arc=2mm,
    title={Prompt for Countdown Task}
]
For the given numbers, find a sequence of arithmetic operations that results in the target number. Show your reasoning and conclude with "The answer is: "
\end{tcolorbox}

\subsection{Implementation Details of Baselines}
\label{appendix:baselines_details}
This subsection describes the implementation details of the baseline decoding strategies evaluated in our experiments.

\textbf{Uniform Sampler.}
This strategy serves as the most basic decoding baseline for MDMs. At each decoding step, a fixed number of masked positions are selected uniformly at random and unmasked.

\textbf{Confidence-Based Sampler.}
Confidence-based decoding is one of the most widely used sampling strategies for MDMs and is adopted by models such as LLaDA~\cite{nie2025large}. At each decoding step $t$, it assigns an unmasking priority to every masked position based on the model's predictive confidence, and then selects the positions with the highest confidence to unmask. Formally, let $x_t$ denote the partially revealed sequence at step $t$. For a masked position $i$, the confidence score is defined as the maximum probability of the most likely token under the model's predictive distribution:
\begin{equation}
\small
s_{\text{conf},t}^i = \max_{v \in \mathcal{V}} p_\theta( v \mid x_t,i),
\end{equation}
where $\mathcal{V}$ denotes the vocabulary. The sampler then unmasks the position with the largest score. Unless otherwise specified, we unmask one position per decoding step for this sampler.

\textbf{Entropy-Based Sampler.}
This strategy evaluates the uncertainty of the entire predictive distribution by using entropy as a proxy~\cite{ben2025accelerated} at each position. A lower entropy indicates a more peaked and confident distribution. The score is given by the negative entropy of the predictive distribution after applying the softmax function:
\begin{equation}
\small
    s_{\text{ent},t}^i = \sum_{v \in \mathcal{V}} p_\theta( v | x_{t},i) \log p_\theta(v | x_{t},i).
\end{equation}

\textbf{Margin-Based Sampler.}
This alternative measures the model's uncertainty by computing the probability margin between the two most confident candidate tokens at each position~\cite{kim2025train}. A larger margin indicates a more decisive prediction. The score is defined as:
\begin{equation}
\small
    s_{\text{margin},t}^i = p_\theta( v_1 | x_{t},i) - p_\theta( v_2 | x_{t},i),
\end{equation}
where $v_1$ and $v_2$ denote the two most likely tokens for position $i$ according to the model’s predictive distribution $p_\theta(\cdot \mid x_{t},i)$.

\textbf{EB-Sampler.}
The Entropy-Bounded Sampler (EB-Sampler)~\cite{ben2025accelerated} accelerates generation by unmasking a variable number of tokens at each step. The number of tokens unmasked is controlled by an error-tolerance hyperparameter $\gamma$, which constrains the total uncertainty to preserve generation quality. At each iteration, masked positions in $\mathcal{M}_t$ are ranked by their entropy, and the largest subset is selected such that the cumulative entropy, used as a proxy for joint uncertainty, remains bounded as follows:
\begin{equation}
\label{eq:eb_sampler}
\small
    \sum_{i \in \mathcal{M}_t} H(p_\theta(\cdot|x_{t},i)) - \max H(p_\theta(\cdot|x_{t})) \le \gamma,
\end{equation}
where $H(\cdot)$ denotes the token entropy. This strategy enables more aggressive parallel decoding when predictions are confident. Following \citet{ben2025accelerated}, we set $\gamma = 0.01$ for all datasets in our experiments.

\textbf{Semi-Autoregressive Sampler.}
The semi-autoregressive sampler extends the confidence-based sampler by introducing a partial left-to-right constraint. Specifically, the response is divided into multiple contiguous blocks of equal size. Within each block, unmasking follows the confidence-based sampling strategy, allowing tokens to be revealed in an order-agnostic manner. Across blocks, decoding proceeds sequentially from left to right, so that each block is generated conditioned on all previously completed blocks. In our experiments, we set the number of blocks to 8 for all tasks.

\textbf{Fast-dLLM.} Fast-dLLM~\cite{wu2025fast} builds upon the semi-autoregressive decoding paradigm and further introduces a confidence-aware parallel decoding scheme together with a KV-cache design tailored for MDMs to accelerate inference. Unlike samplers that unmask a fixed number of tokens at each step, Fast-dLLM dynamically unmasks all positions whose predictive confidence exceeds a predefined threshold $\epsilon$. As a result, the number of tokens revealed at each decoding step is adaptive and varies with the model’s confidence distribution. Formally, at decoding step $t$, all masked positions $i$ satisfying the following condition are selected for unmasking:
\begin{equation}
\small
\max_{v \in \mathcal{V}} p_\theta(  v \mid x_{t},i) > \epsilon,
\end{equation}
are unmasked in parallel. Following~\citet{wu2025fast}, we set the confidence threshold to $\epsilon = 0.9$ in all experiments.


\subsection{Details on Integrating with Efficient Decoding Strategies}
\label{appendix:integrating_w_efficient}
In this subsection, we provide implementation details for integrating \method{} with three efficient decoding strategies, as discussed in Section~\ref{subsec:further_analysis}. For $\tau$-leaping~\cite{chen2023accelerating}, we follow the original protocol and unmask a fixed number of tokens ($\tau$) at each decoding step. In our experiments, we set $\tau = 2$. The specific positions to unmask are selected according to the calibrated score used by \method{}. For the EB-Sampler~\cite{ben2025accelerated}, we replace the entropy-based priority with our calibrated score for ranking masked positions (Eq.~\ref{eq:score_function}), while retaining the original cumulative budget constraint defined by the error tolerance. We set the error tolerance parameter to $\gamma = 0.01$ in all experiments. For Fast-dLLM~\cite{wu2025fast}, we apply the unmasking threshold to the calibrated score used in our main experiments, rather than to the raw predictive confidence, enabling positions with high calibrated scores to be unmasked in parallel. We set the threshold to $\epsilon = 0.6$. To further accelerate decoding, we adopt the prefix KV cache mechanism proposed in Fast-dLLM.

\input{figs/hyperparameters}

\subsection{Decoding Behavior under Different Positional and Clipping Settings}
\label{appendix:hyperparameter_analysis}
In this subsection, we analyze the effect of two key hyperparameters in \method{} on decoding behavior, namely the positional coefficient $\lambda$ and the clipping threshold $\alpha$. Rather than identifying a single optimal setting, this analysis aims to characterize the decoding behaviors induced by different parameter choices. The results are summarized in Figure~\ref{fig:ablation_hyperparams}.

Figure~\ref{fig:sub_lambda} illustrates that the positional coefficient $\lambda$ governs the degree of sequential inductive bias in decoding. For most tasks except Sudoku, moderate values (e.g., $\lambda=0.25$) strike a balance between order-agnostic generation and progressive decoding, leading to strong overall performance. In contrast, larger values impose a stronger left-to-right constraint, which reduces decoding flexibility and can degrade performance on tasks that benefit from parallel or non-sequential reasoning. For globally constrained planning tasks such as Sudoku, smaller values of $\lambda$ are preferable, as they avoid prematurely enforcing a sequential decoding order. These results suggest that different tasks naturally align with different decoding trajectories, underscoring the importance of decoding order in MDMs, consistent with observations in~\citet{kim2025train}.

Figure~\ref{fig:sub_alpha} examines the effect of the clipping threshold $\alpha$. Performance remains stable across a broad range of values, except for extreme settings (e.g., $\alpha=100$), indicating that \method{} is robust to the choice of $\alpha$. In our experiments, we fix $\alpha$ to 10 for simplicity and consistency across tasks.



\begin{table*}[t]
    \centering
    \small
    \begin{tabular}{l cccccccc}
    \toprule
    \textbf{Method} & \textbf{HumanEval} & \textbf{MBPP} & \textbf{GSM8K} & \textbf{Math500} & \textbf{GPQA} & \textbf{Countdown} & \textbf{Sudoku} & \textbf{Avg.} \\
    \midrule
    Confidence & 27.4 & 42.4 & 59.1 & 20.8 & 27.9 & 34.0 & 23.8 & 33.6 \\
    L-2-R + Sem. Prior & 40.2 & 45.2 & 78.2 & 32.8 & 27.9 & 35.7 & 0.0 & 37.1 \\
    Semi-AR + Sem. Prior & 40.2 & 44.7 & 78.8 & 33.2 & 27.5 & 32.8 & 0.0 & 36.7 \\
    \rowcolor{blue!15}
    \method{} & 42.1 & 47.8 & 79.2 & 34.8 & 29.2 & 36.3 & 29.8 & 42.7 \\
    \bottomrule
    \end{tabular}
    \caption{Comparison of \method{} against fixed-order baselines augmented with the semantic prior.}
    \label{tab:post_disentangle}
    \end{table*}
    
\subsection{Validation of Non-Autoregressive Flexibility in \method{}}
\label{appendix:post_baseline_disentanglement}
To further isolate the performance contribution of the Positional Trajectory Prior and verify whether \method{} leverages the flexibility of MDMs, we compare it against strict fixed-order baselines augmented with the semantic informativeness prior. Specifically, we evaluate a pure left-to-right strategy (L-2-R + Sem. Prior) and a block-wise left-to-right strategy (Semi-AR + Sem. Prior), reporting the results across all seven benchmarks on LLaDA-8B-Instruct.

As detailed in Table~\ref{tab:post_disentangle}, \method{} significantly outperforms both fixed-order baselines (L-2-R + Sem. and Semi-AR + Sem.) by 5.6\% and 6.0\%, respectively. This confirms that the dynamic decoding trajectory of the Positional Trajectory Prior enables \method{} to leverage non-autoregressive flexibility of MDMs, which is essential for superior performance.

\begin{table*}[t]
\centering
\small
\begin{tabular}{l cccccccc}
\toprule
\textbf{Method} & \textbf{HumanEval} & \textbf{MBPP} & \textbf{GSM8K} & \textbf{Math500} & \textbf{GPQA} & \textbf{Countdown} & \textbf{Sudoku} & \textbf{Avg.} \\
\midrule
Confidence & 18.9 & 40.5 & 58.2 & 24.0 & 27.2 & 22.8 & 0.0 & 27.4 \\
Semi-AR & 37.8 & 39.1 & 75.4 & 33.2 & 26.8 & 27.3 & 0.0 & 34.2 \\
Fast-dLLM & 38.4 & 40.3 & 75.7 & 34.2 & 26.8 & 33.8 & 0.0 & 35.6 \\
\rowcolor{blue!15}
\method{} & 43.3 & 48.2 & 79.9 & 34.4 & 30.4 & 35.7 & 26.0 & 42.6 \\
\bottomrule
\end{tabular}
\caption{Performance under an extended maximum generation length of 512.}
\label{tab:post_long_512}
\end{table*}

\subsection{Long-Sequence Generalization}
\label{appendix:post_long_sequence}
To evaluate whether the proposed method generalizes to longer sequences, we extend the maximum generation length to 512 while keeping all other settings unchanged. Table~\ref{tab:post_long_512} summarizes the results on LLaDA-8B-Instruct.

As shown in Table~\ref{tab:post_long_512}, \method{} exhibits robust performance when the generation length is extended, achieving an average score of 42.6\%. This indicates that the proposed priors effectively guide the MDM to leverage the expanded generation capacity, preserving a substantial performance advantage over the baselines.

\subsection{Inference Overhead Analysis}
\label{appendix:post_efficiency}
To assess the computational efficiency of \method{}, we provide quantitative latency measurements comparing our approach against the standard confidence-based sampler. As shown in Table~\ref{tab:post_latency}, introducing both the Positional Trajectory Prior and the Semantic Informativeness Prior yields only a negligible per-step overhead.

\begin{table}[h]
\centering
\small
\begin{tabular}{lcc}
\toprule
\textbf{Method} & \textbf{MBPP} & \textbf{Math500} \\
\midrule
Confidence & 1.03 & 48.88 \\
\method{} & 1.04 & 49.08 \\
\bottomrule
\end{tabular}
\caption{Latency comparison (seconds per iteration) on LLaDA-8B-Instruct with 4$\times$A100 GPUs.}
\label{tab:post_latency}
\end{table}

\subsection{Adaptive $\lambda$ Selection via Mean Predictive Entropy Minimization}
\label{appendix:adaptive_lambd}
In this subsection, we propose an adaptive strategy for selecting the positional coefficient $\lambda$. We first define the mean predictive entropy and show that it correlates well with generation quality across different $\lambda$ settings. We then leverage entropy minimization to select $\lambda$ automatically, without requiring manual tuning or ground-truth labels.

\textbf{Definition of Mean Predictive Entropy.}  
For an individual data instance, we define the \textit{sequence-level predictive entropy} as the average entropy of the predictive distributions over all masked tokens at each decoding step. Specifically, for each decoding step \( t \), we compute the entropy \( \mathcal{H}_t(i) \) for each masked token \( i \in \mathcal{M}_t \), where \( \mathcal{M}_t \) is the set of masked positions at step \( t \), and \( \mathcal{H}_t(i) \) denotes the entropy of the predictive distribution for token \( i \) at step \( t \). The \textit{sequence-level predictive entropy} for a given instance is then computed by averaging the entropies across all decoding steps \( t \):
\begin{equation}
\small
\mathcal{H}_{\text{seq}} = \frac{1}{T} \sum_{t=1}^{T} \frac{1}{|\mathcal{M}_t|} \sum_{i \in \mathcal{M}_t} \mathcal{H}_t(i),
\end{equation}
where \( T \) is the total number of decoding steps, and \( |\mathcal{M}_t| \) is the number of masked positions at step \( t \).

\input{figs/entropy_performance}

\input{tables/appendix_different_corpus}

Next, to compute the \textit{mean predictive entropy} at the dataset level, we extend this definition to the entire dataset. Let \( \mathcal{D}_{\text{cal}} = \{\mathbf{x}_1, \dots, \mathbf{x}_N\} \) be the calibration subset of \( N \) instances. The \textit{mean predictive entropy} is then defined as the average of the sequence-level predictive entropies across all instances in the calibration set:
\begin{equation}
\small
\bar{\mathcal{H}} = \frac{1}{N} \sum_{j=1}^{N} \mathcal{H}_{\text{seq}}^{(j)},
\end{equation}
where \( \mathcal{H}_{\text{seq}}^{(j)} \) is the sequence-level predictive entropy for the \(j\)-th instance. The resulting \( \bar{\mathcal{H}} \) serves as a proxy for the overall generation quality, which is used for selecting the guidance weight \( \lambda \) in the subsequent strategy.

\textbf{Empirical Validation of the Minimum Entropy Principle.}
To assess whether mean predictive entropy is a reliable proxy for output quality, we perform a correlation analysis on a calibration subset of 100 instances randomly sampled from GSM8K and HumanEval, as shown in Figure~\ref{fig:entropy_per}. For each candidate $\lambda \in \Lambda = \{\lambda_1, \dots, \lambda_K\}$, we decode the entire subset and compute: (i) the mean predictive entropy $\bar{\mathcal{H}}(\lambda)$, and (ii) the corresponding task metric (Accuracy for GSM8K, pass@1 for HumanEval). We then compute the Pearson correlation coefficient ($r$) between the mean predictive entropy and task performance across the $K$ candidate $\lambda$ values.  
As shown in Figure~\ref{fig:entropy_per_gsm} for GSM8K and Figure~\ref{fig:entropy_per_humaneval} for HumanEval, we observe a strong negative correlation, indicating that lower mean entropy corresponds to better task performance. This finding motivates entropy minimization as a label-free criterion for selecting $\lambda$.

\textbf{Adaptive Selection via Entropy Minimization.}
Based on the observation above, we select the guidance weight $\lambda$ by minimizing the mean predictive entropy on a small calibration set. Let $\mathcal{D}_{\mathrm{cal}} = \{\mathbf{x}_1, \dots, \mathbf{x}_N\}$ denote a small subset of validation inputs and $\Lambda = \{\lambda_1, \dots, \lambda_K\}$ be the candidate set. For each $\lambda \in \Lambda$, we decode $\mathcal{D}_{\mathrm{cal}}$ to obtain outputs $\{\mathbf{y}_{\lambda}^{(j)}\}_{j=1}^{N}$. We then compute the mean predictive entropy:
\begin{equation}
\small
\bar{\mathcal{H}}(\lambda) = \frac{1}{N} \sum_{j=1}^{N} \mathcal{H}_{\text{seq}}(\mathbf{y}_{\lambda}^{(j)}),
\end{equation}
where $\mathcal{H}_{\text{seq}}(\mathbf{y})$ denotes the \emph{average predictive entropy} of a decoded sequence. Finally, we select the entropy-minimizing weight:
\begin{equation}
\small
\lambda^* = \arg\min_{\lambda \in \Lambda} \bar{\mathcal{H}}(\lambda).
\end{equation}
By selecting $\lambda$ via entropy minimization on a small calibration subset, this strategy avoids manual tuning and does not require ground-truth labels during selection.

\begin{table*}[t]
    \centering
    \small
    \setlength{\tabcolsep}{5pt}
    \begin{tabular}{l cccccccc}
    \toprule
    \textbf{$\lambda$} & \textbf{HumanEval} & \textbf{MBPP} & \textbf{GSM8K} & \textbf{Math500} & \textbf{GPQA} & \textbf{Countdown} & \textbf{Sudoku} & \textbf{Avg.} \\
    \midrule
    0.00 & 7.93 {\scriptsize (2.397)} & 33.5 {\scriptsize (1.538)} & 59.7 {\scriptsize (2.565)} & 4.0 {\scriptsize (2.482)} & 29.0 {\scriptsize (3.852)} & 34.0 {\scriptsize (2.532)} & \textbf{29.8} {\scriptsize (1.288)} & 28.28 \\
    0.25 & \textbf{42.1} {\scriptsize (1.465)} & \textbf{47.8} {\scriptsize (0.495)} & \textbf{79.2} {\scriptsize (2.014)} & \textbf{34.8} {\scriptsize (1.187)} & \textbf{29.2} {\scriptsize (3.821)} & 34.7 {\scriptsize (1.641)} & 0.2 {\scriptsize (2.366)} & \textbf{38.29} \\
    0.50 & 41.5 {\scriptsize (1.556)} & \textbf{47.8} {\scriptsize (0.497)} & 78.9 {\scriptsize (2.017)} & 34.0 {\scriptsize (1.318)} & 28.8 {\scriptsize (3.871)} & \textbf{36.3} {\scriptsize (1.529)} & 0.0 {\scriptsize (2.791)} & 38.19 \\
    1.00 & 40.2 {\scriptsize (1.575)} & 45.4 {\scriptsize (0.543)} & 77.7 {\scriptsize (2.041)} & 34.4 {\scriptsize (1.266)} & 28.4 {\scriptsize (3.895)} & 35.7 {\scriptsize (1.506)} & 0.0 {\scriptsize (2.474)} & 37.93 \\
    \bottomrule
    \end{tabular}
    \caption{Adaptive $\lambda$ selection with only 10 unlabeled calibration instances. Values in parentheses are mean predictive entropy measured on the 10-instance calibration subset.}
    \label{tab:post_adaptive_n10}
    \end{table*}
    
\textbf{Robustness Under Limited Calibration Data.}
\label{appendix:post_adaptive_robustness}
We further evaluate the robustness of the adaptive strategy under limited calibration data. Specifically, we randomly sample only 10 instances to form a calibration set and compute the mean predictive entropy for each candidate $\lambda$, alongside the model's corresponding performance. As shown in Table~\ref{tab:post_adaptive_n10}, even when relying on merely 10 instances to estimate the mean predictive entropy, our adaptive strategy consistently selects the optimal $\lambda$. This demonstrates that entropy minimization remains effective even with highly limited unlabeled calibration data.

\input{tables/main_res_dream}

\subsection{Impact of Calibration Corpus}
\label{appendix:effect_of_calibration_corpus}
In this subsection, we investigate the impact of the calibration corpus $\mathcal{D}'$ on \method{} through a series of controlled ablation studies. Specifically, we examine two key dimensions: (1) the source of the calibration corpus, and (2) the corpus size. The results are summarized in Table~\ref{tab:diff_corpus}.

We first assess the robustness of \method{} across varied calibration corpora. Specifically, we compare the default C4 (16GB) against two alternatives: (i) a domain-specific mixture of BookCorpusOpen and OpenR1-Math-220K (denoted as Book \& Code), and (ii) SlimPajama~\cite{cerebras2023slimpajama}. Across these settings, we observe comparable average performance. These results indicate that \method{} is generally robust to the selection of calibration sources.

Next, we investigate the impact of corpus size on performance stability. We sample three larger subsets from C4, sized 50GB, 200GB, and 400GB. We observe that while increasing the corpus size initially improves accuracy, the performance gains saturate beyond the 200GB scale. We hypothesize that the estimated token distribution converges once the corpus reaches a sufficient size; consequently, further enlarging the dataset provides marginal benefit for the calibration signal. Overall, we observe that \method{} benefits from improved calibration, but does not exhibit strong sensitivity to the specific calibration corpus used.

\input{tables/case_gsm8k}
\input{tables/case_gsm8k_2}

\subsection{Results on Dream}
\label{appendix:dream_dream}
To further evaluate the generalization capability of our approach, we apply \method{} to Dream-v0-Instruct-7B~\cite{dream2025}, a state-of-the-art MDM developed independently of the LLaDA series~\cite{nie2024scaling}. As shown in Table~\ref{tab:main_results_revised_with_dream}, \method{} consistently achieves the best performance across all evaluated tasks, with an average score of 40.1\%. This represents a substantial improvement over the strongest uncertainty-based baseline, margin-based sampling, which reaches 27.8\%. In particular, \method{} achieves 57.9\% on HumanEval and 76.4\% on GSM8K, demonstrating strong performance in both code generation and mathematical reasoning tasks.

These results indicate that \method{} is not limited to a specific model family, but instead addresses a broader limitation inherent to masked diffusion models. Across different architectures, we consistently observe a tendency for uncertainty-based decoding to prioritize boundary tokens early in the decoding process and to favor semantically trivial tokens. This behavior suggests a systematic bias in how unmasking priorities are assigned, rather than an artifact of a particular model or implementation. By incorporating both positional awareness and semantic awareness into the sampling process, \method{} mitigates these biases and enables the model to allocate its generation capacity more effectively. As a result, the decoding process is better aligned with the underlying reasoning or planning structure of the task, leading to more coherent and logically ordered outputs. 

\subsection{Case Study}
\label{appendix:case_study}
In this subsection, we present three case studies, as shown in Figure~\ref{tab:case_study_gsm8k}, Figure~\ref{tab:case_study_gsm8k_2}, and Figure~\ref{tab:case_study_humaneval}, drawn from the GSM8K (mathematical reasoning) and HumanEval (code generation) benchmarks. The first two cases are from GSM8K, and the third is from HumanEval. These case studies compare the baseline confidence-based sampling strategy with our proposed \method{} approach.

As shown in Table~\ref{tab:case_study_gsm8k}, due to rigid boundary bias, the confidence sampler prematurely commits to a final answer of \emph{2975} at an early stage, before computing the correct total number of classes and students. Once this premature decision is made, subsequent tokens are forced to justify the incorrect conclusion rather than revising it, resulting in a fluent yet numerically incorrect solution. In contrast, \method{} incorporates the Positional Trajectory Prior, which delays emitting the final answer and first completes the intermediate computations (e.g., 25 weekday classes, 8 Saturday classes, 33 total classes, 495 students, and the resulting revenue). The decoding trajectory allocates more steps to these intermediate reasoning tokens, allowing for a coherent, step-by-step derivation that leads to the correct answer of \emph{7425}.

As illustrated in Table~\ref{tab:case_study_gsm8k_2}, due to the trivial token bias, the confidence sampler decodes trivial tokens at critical steps (e.g., the comma is decoded at step 121, at position 76, instead of the 7 at position 75), thereby locking the reasoning space. This leads to errors, such as the comma being used as a thousand separator, causing invalid representations like 0,700. Consequently, the model fails to predict 700, ultimately resulting in a calculation error. In contrast, \method{} effectively suppresses the selection of trivial tokens, enabling the model to arrive at the correct solution.

For the HumanEval benchmark, as shown in Table~\ref{tab:case_study_humaneval}, the baseline’s rigid boundary bias often leads to premature sequence termination and incomplete code, resulting in a significant waste of the generation budget. In contrast, \method{} mitigates this issue by incorporating position-aware weighting and content-aware confidence calibration, which introduce an inductive bias toward sequential reasoning. As a result, \method{} ensures that complete and correct code is generated before sequence termination.

These case studies demonstrate that explicit trajectory control and trivial token suppression are critical for complex problem solving. By mitigating both the rigid boundary bias and the trivial token bias, \method{} enables the model to produce outputs that are more coherent, accurate, and complete.

\input{tables/case_code}

%% file: appendix_figs/interface_ex.tex
\begin{figure}[!t]
    \centering
    \subfigure[``\texttt{<space>}'' inserted at 25\% of the sequence length.]{
        \includegraphics[width=0.44\linewidth]{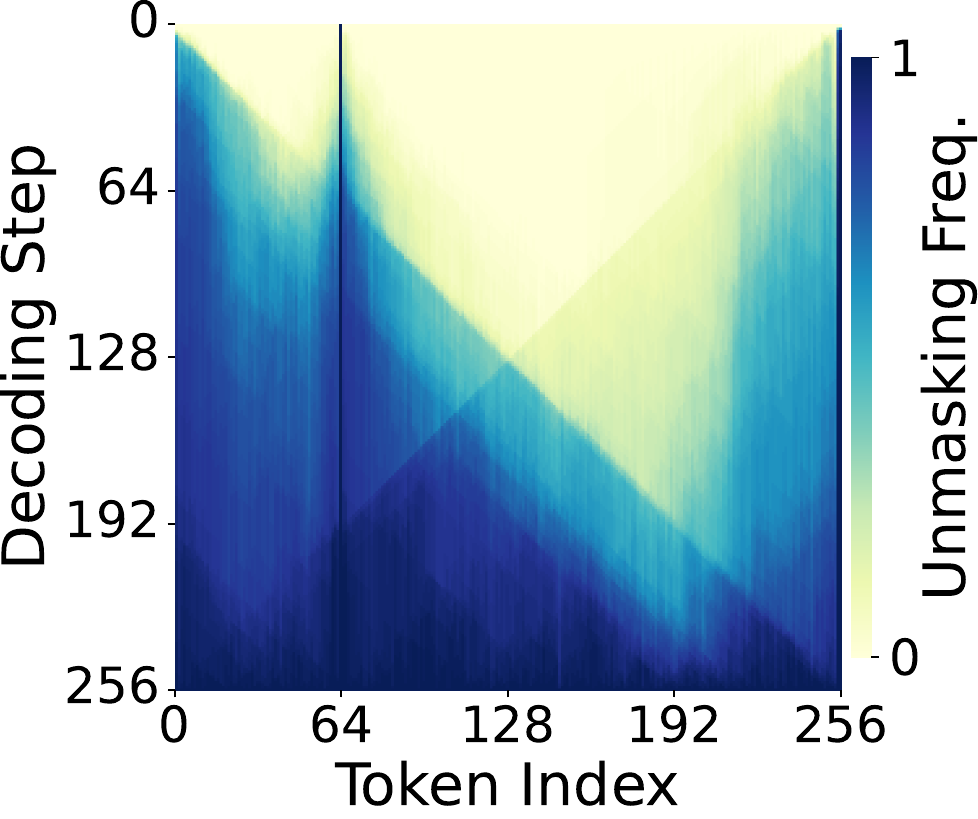}
        \label{fig:interface_ex_space25}
    }
    \hfill 
    \subfigure[``\texttt{<space>}'' inserted at 75\% of the sequence length.]{
        \includegraphics[width=0.44\linewidth]{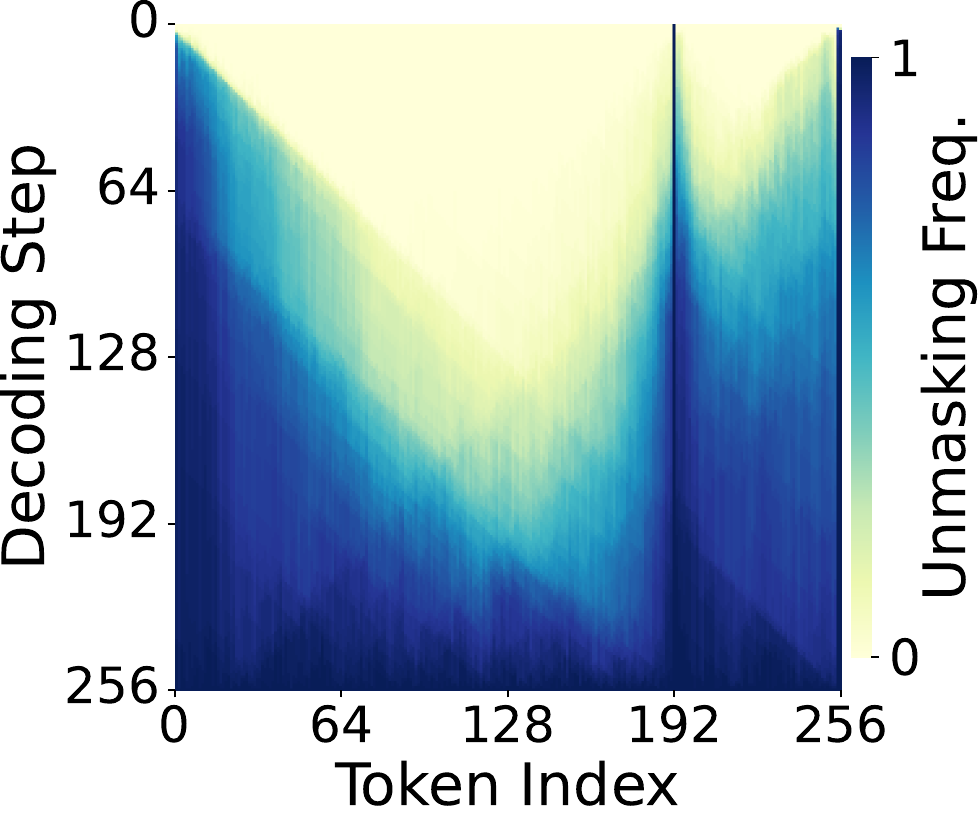}
        \label{fig:interface_ex_space75}
    }


    \subfigure[``the'' inserted at 25\% of the sequence length.]{
        \includegraphics[width=0.44\linewidth]{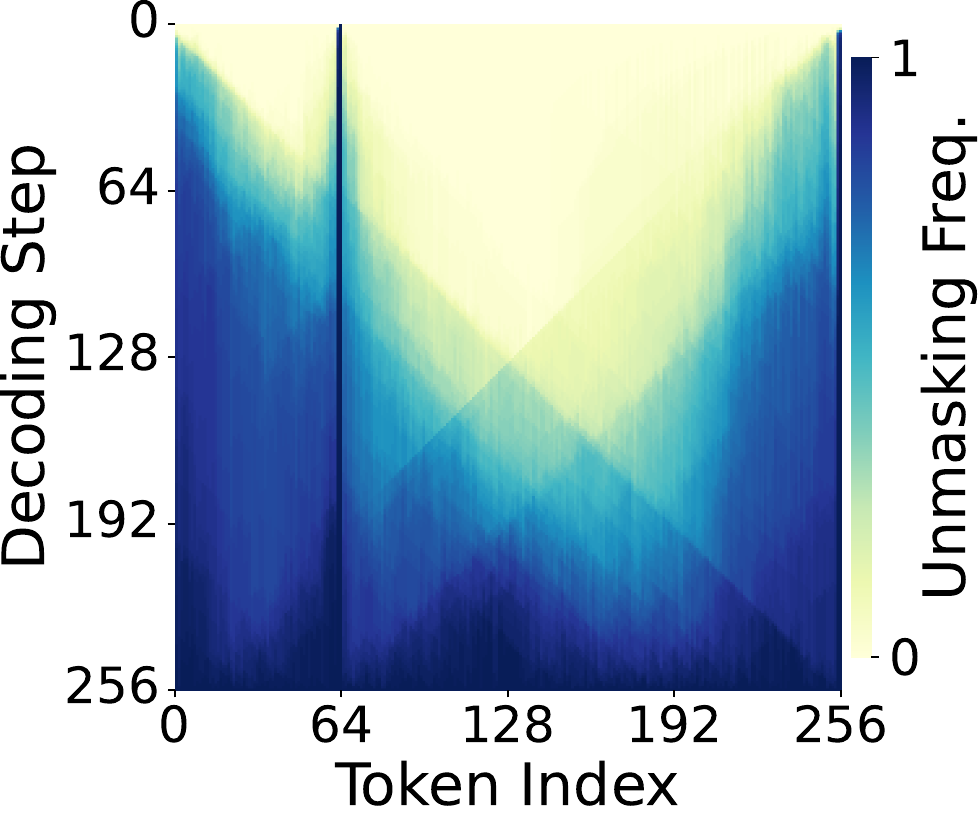}
        \label{fig:interface_ex_the25}
    }
    \hfill 
    \subfigure[``the'' inserted at 75\% of the sequence length.]{
        \includegraphics[width=0.44\linewidth]{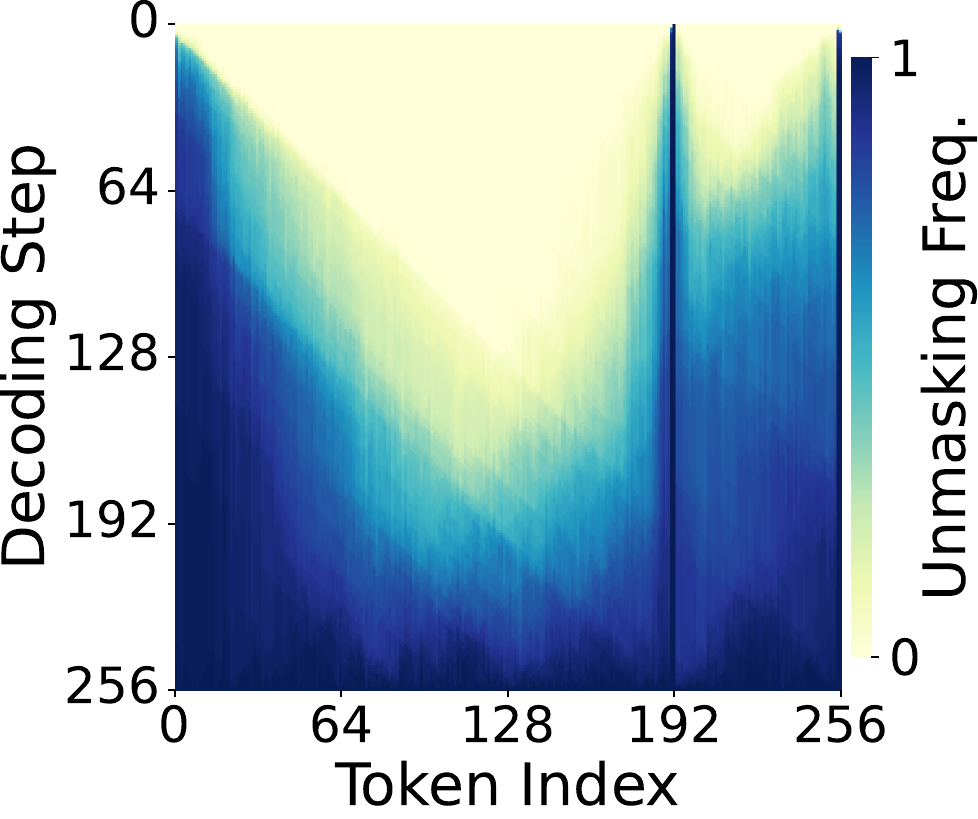}
        \label{fig:interface_ex_the75}
    }

    \subfigure[``.'' inserted at 25\% of the sequence length.]{
        \includegraphics[width=0.44\linewidth]{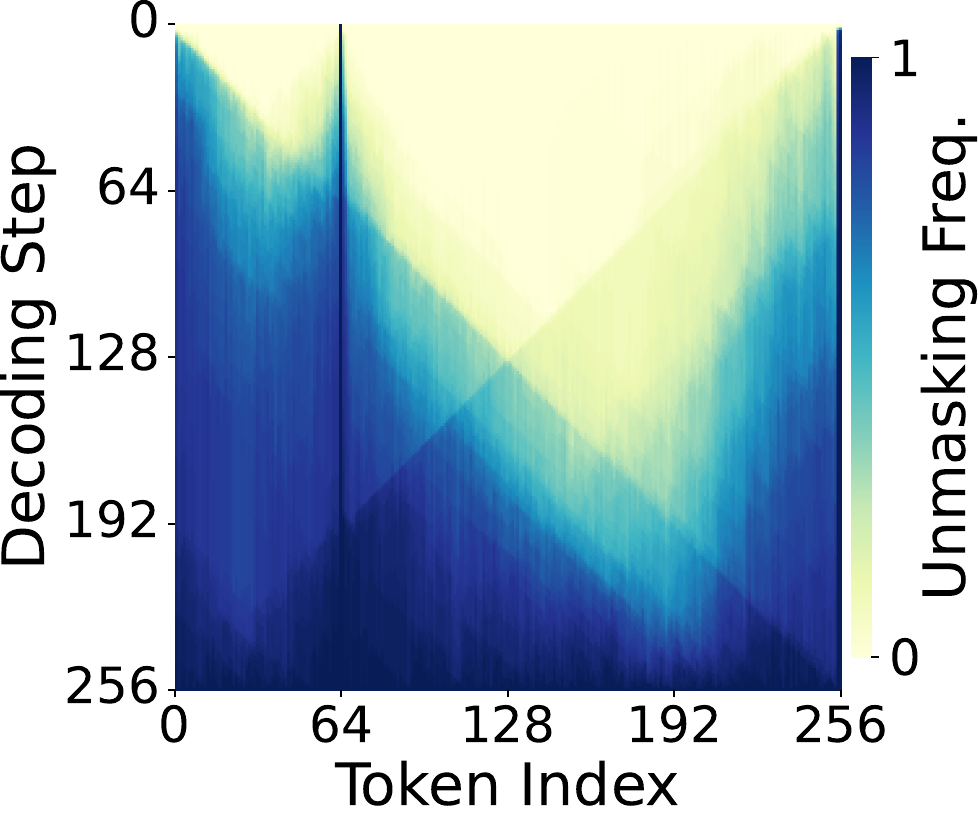}
        \label{fig:interface_ex_dot25}
    }
    \hfill 
    \subfigure[``.'' inserted at 75\% of the sequence length.]{
        \includegraphics[width=0.44\linewidth]{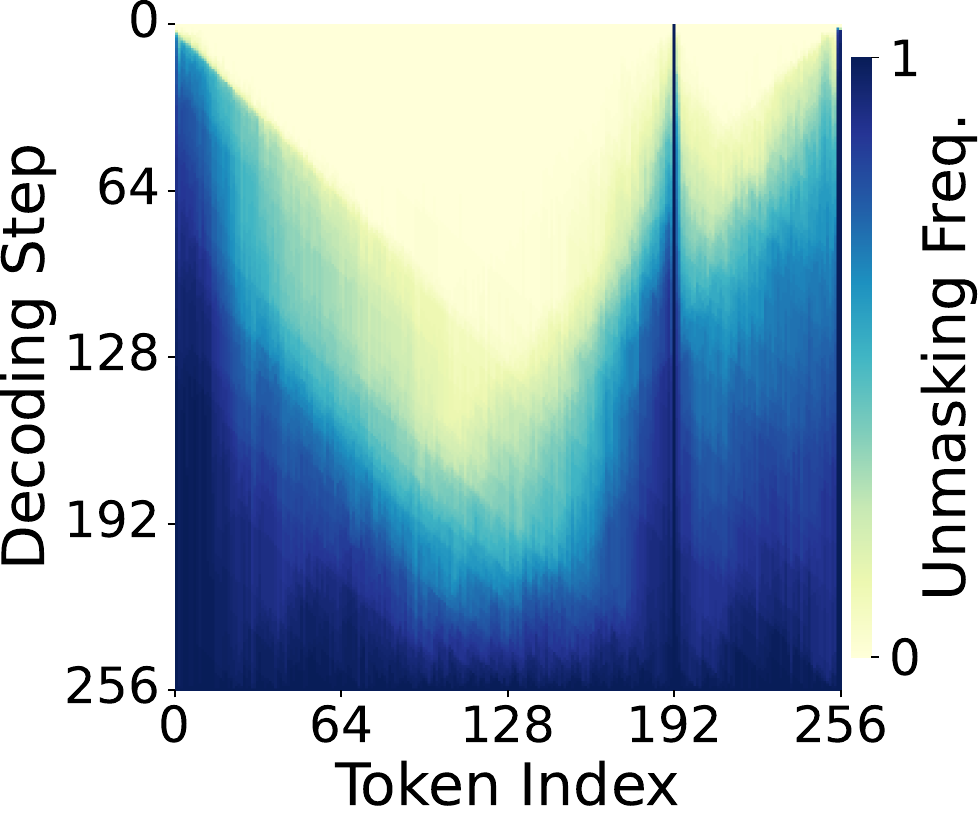}
        \label{fig:interface_ex_dot75}
    }
    \vspace{-0.5em}
    \caption{Unmasking dynamics under anchor insertion. A trivial high-frequency token (e.g., space, ``the'', or ``.'') is manually unmasked before decoding at an intermediate position (25\% or 75\% of the sequence length). The resulting unmasking patterns reveal a strong local positional bias, with decoding trajectories propagating outward from the inserted anchor.}
    \label{fig:interface_ex}
    \vspace{-1em}
\end{figure}

%% file: appendix_figs/entropy_and_margin_GSM8K.tex
\begin{figure}[!t]
    \centering
    \subfigure[Confidence-based Sampling]{
        \includegraphics[width=0.465\linewidth]{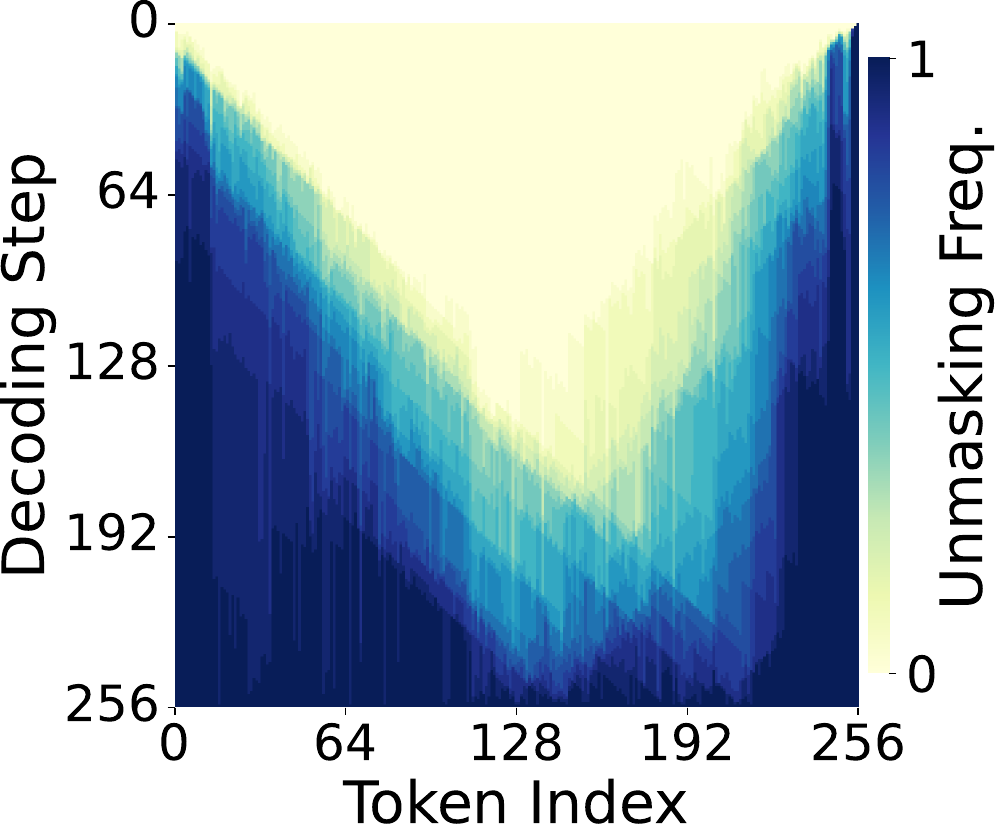}
        \label{fig:appendix_gsm8k_confidence}
    }
    \hfill 
    \subfigure[Entropy-based Sampling]{
        \includegraphics[width=0.465\linewidth]{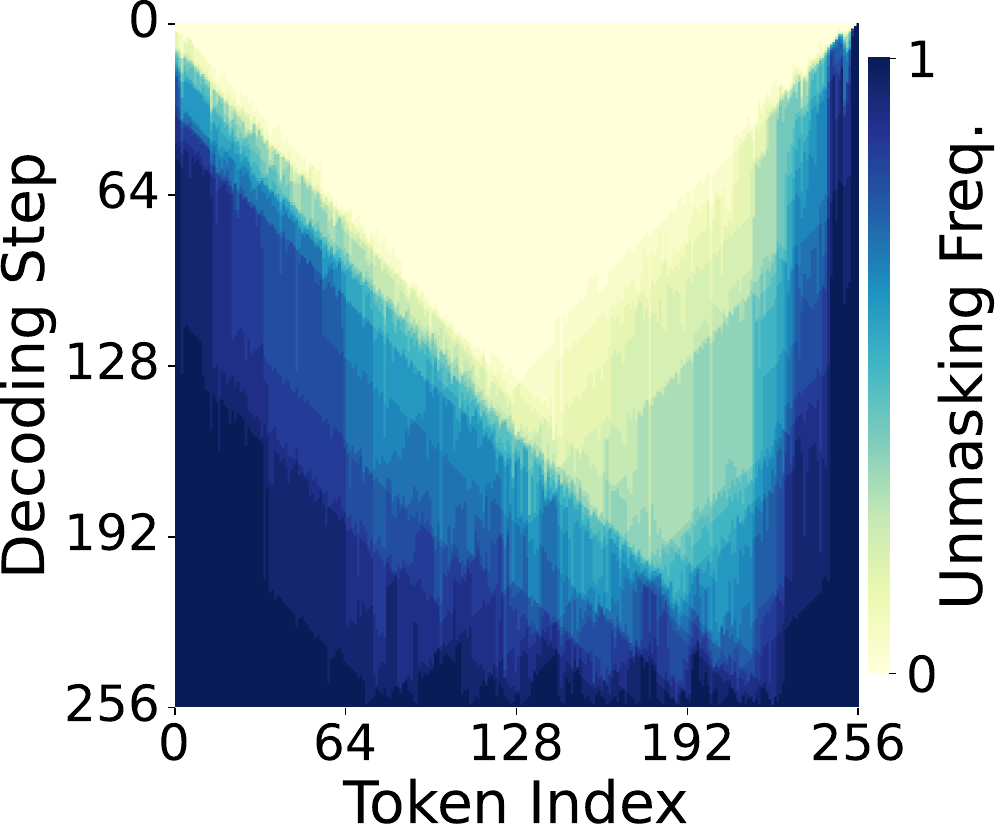}
        \label{fig:appendix_gsm8k_entropy}
    }


    \subfigure[Margin-based Sampling]{

        \includegraphics[width=0.465\linewidth]{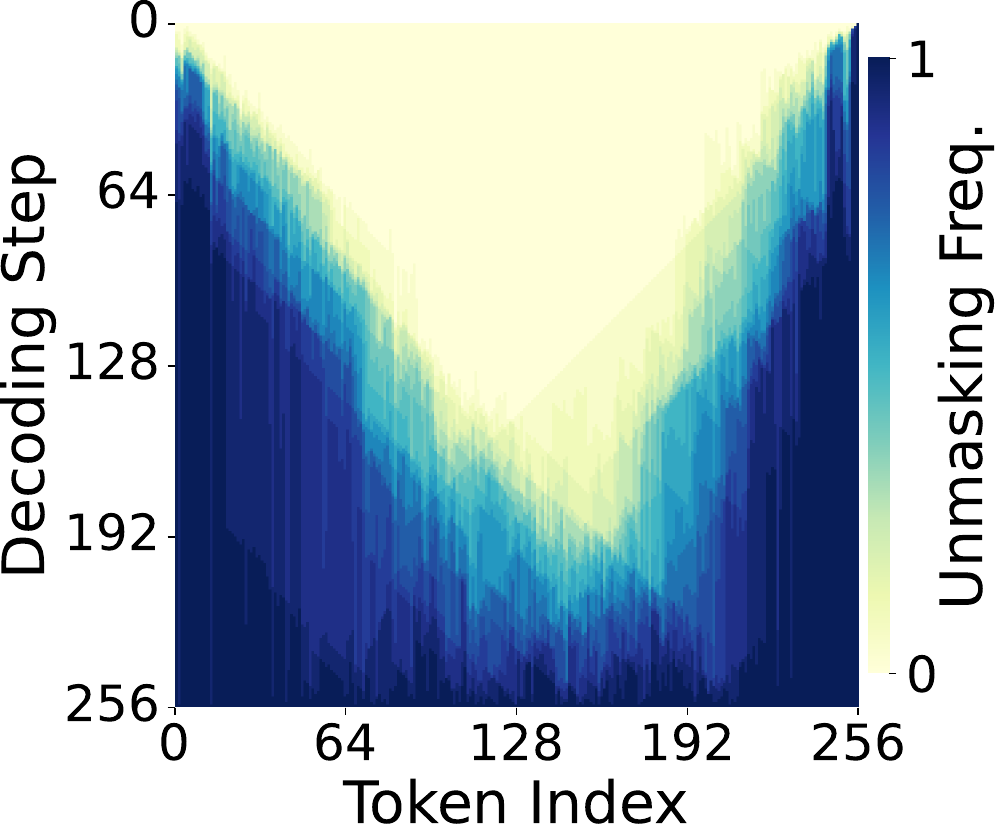}
        \label{fig:appendix_gsm8k_margin}
    }
    \hfill 
    \subfigure[Performance Comparison]{
        \includegraphics[width=0.465\linewidth]{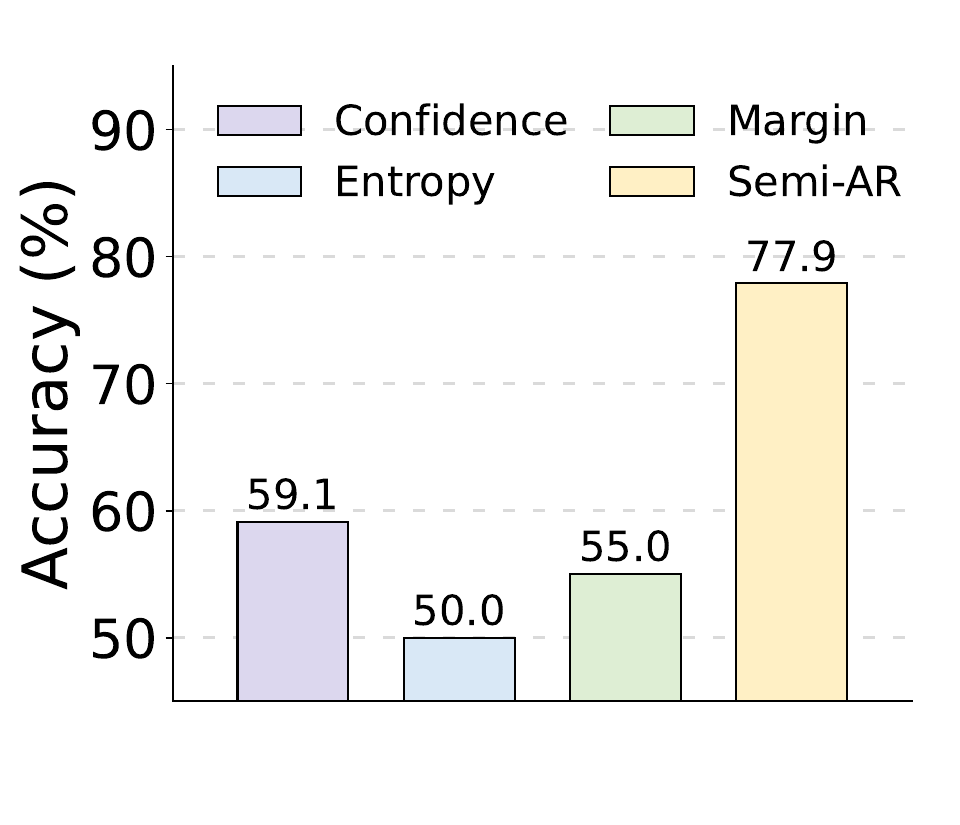}
        \label{fig:appendix_gsm8k_performance}
    }
    \vspace{-0.5em}
    \caption{Consistent boundary-first unmasking patterns across various uncertainty metrics. (a-c) Visualization of unmasking probability for Confidence, Entropy, and Margin samplers on GSM8K. (d) Corresponding performance comparison.}
    \vspace{-1em}
    \label{fig:gsm8k_method_comparison}
\end{figure}

%% file: appendix_figs/entropy_and_margin_MBPP.tex
\begin{figure}[!t]
    \centering
    \subfigure[Confidence-based Sampling]{
        \includegraphics[width=0.465\linewidth]{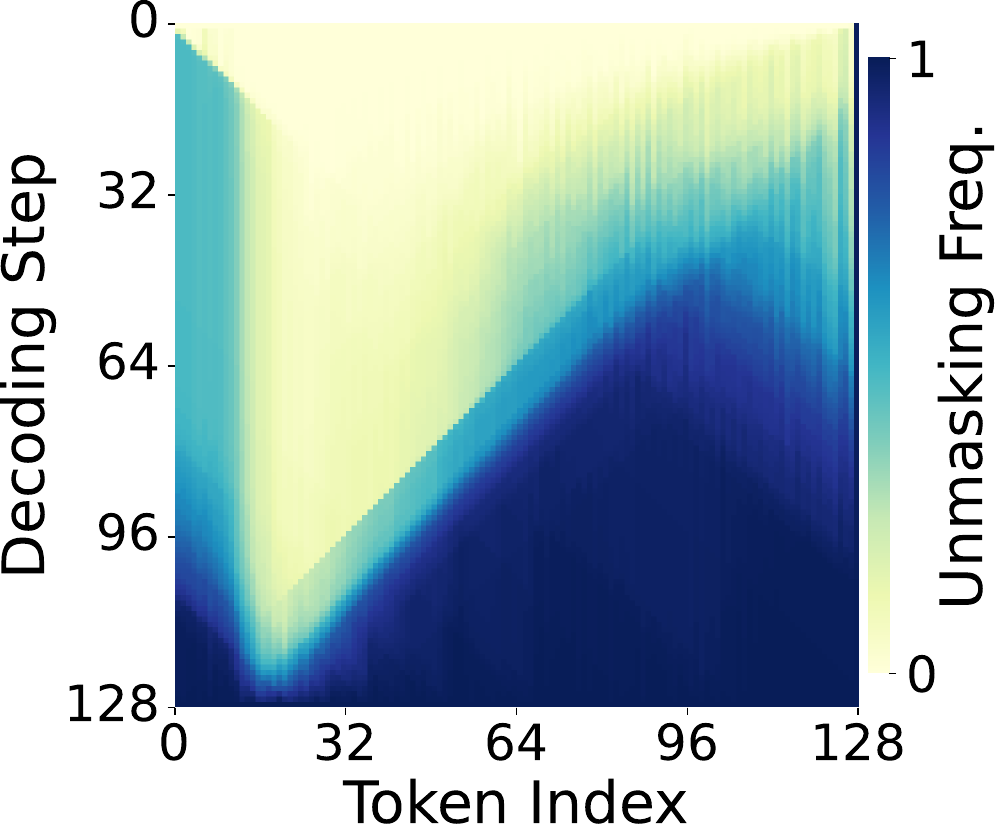}
        \label{fig:appendix_mbpp_confidence}
    }
    \hfill
    \subfigure[Entropy-based Sampling]{
        \includegraphics[width=0.465\linewidth]{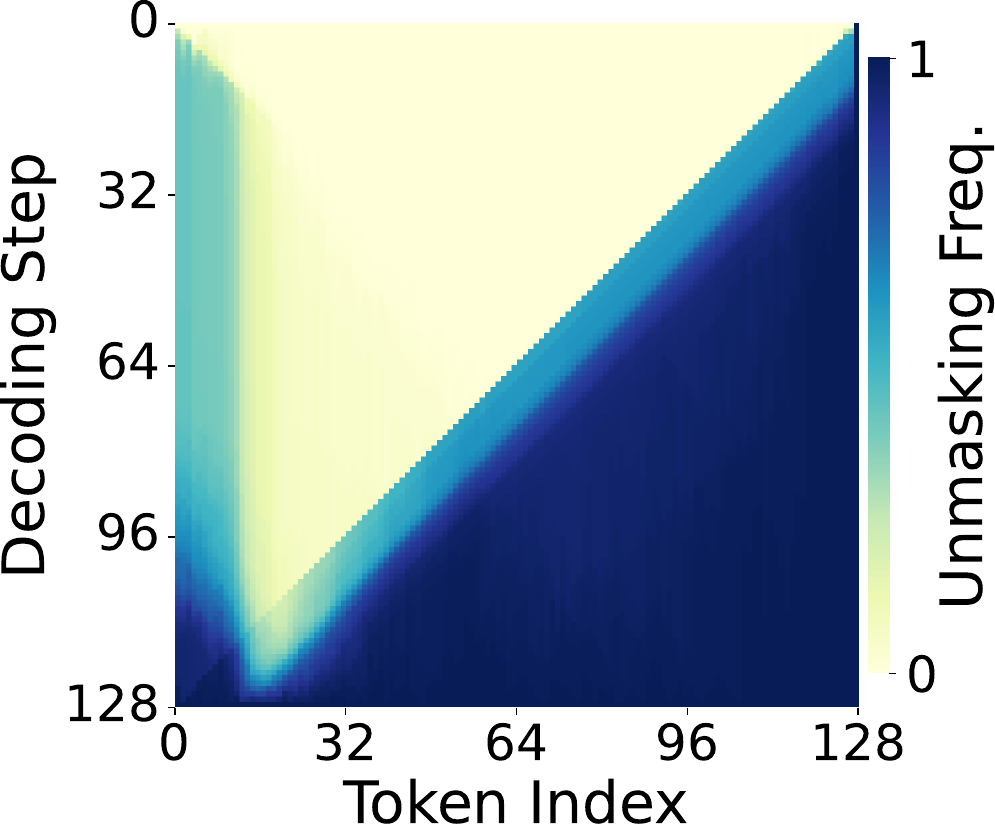}
        \label{fig:appendix_mbpp_entropy}
    }


    \subfigure[Margin-based Sampling]{
        \includegraphics[width=0.465\linewidth]{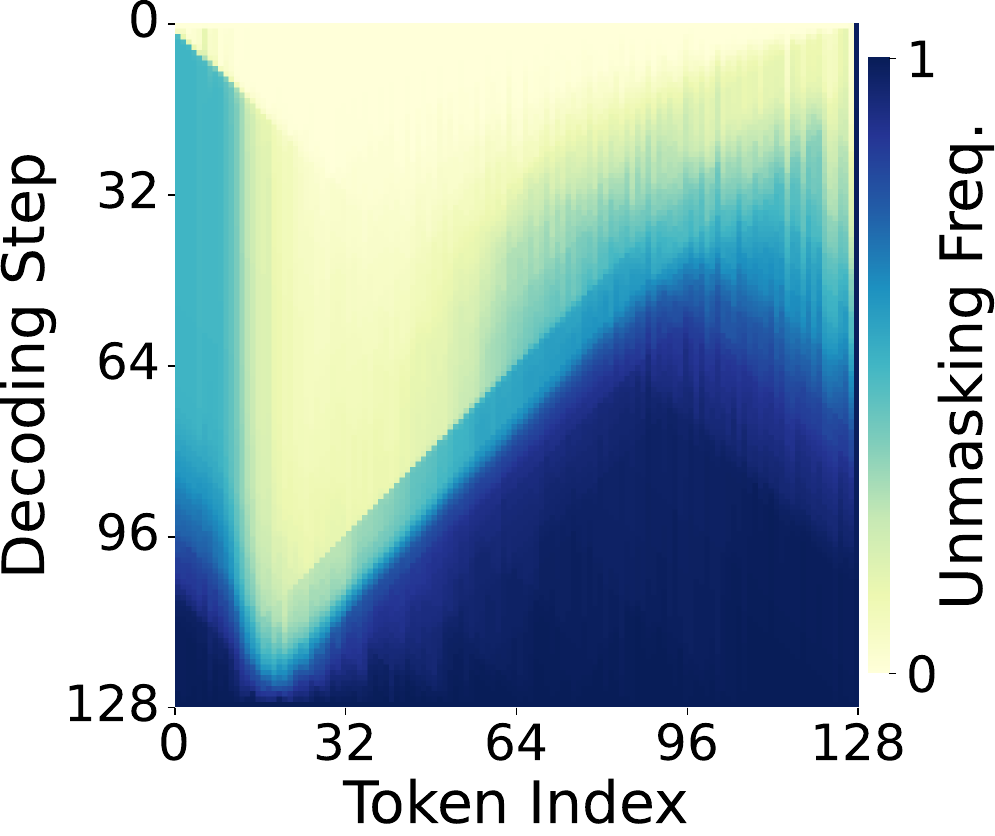}
        \label{fig:appendix_mbpp_margin}
    }
    \hfill
    \subfigure[Performance Comparison]{
        \includegraphics[width=0.465\linewidth]{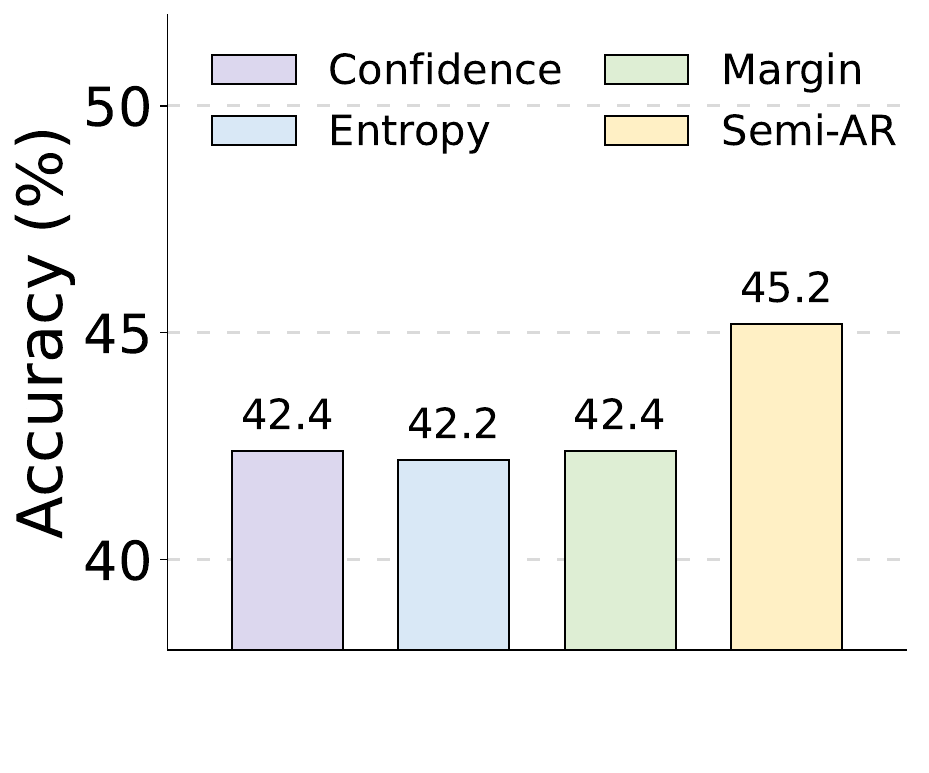}
        \label{fig:appendix_mbpp_performance}
    }
    \vspace{-0.5em}
    \caption{Consistent boundary-first unmasking patterns across various uncertainty metrics. (a-c) Visualization of unmasking probability for Confidence, Entropy, and Margin samplers on MBPP. (d) Corresponding performance comparison.}
    \vspace{-1em}
    \label{fig:mbpp_method_comparison}
\end{figure}

%% file: appendix_figs/entropy_and_margin_humaneval.tex
\begin{figure}[t!]
    \centering
    \subfigure[Confidence-based Sampling]{
        \includegraphics[width=0.465\linewidth]{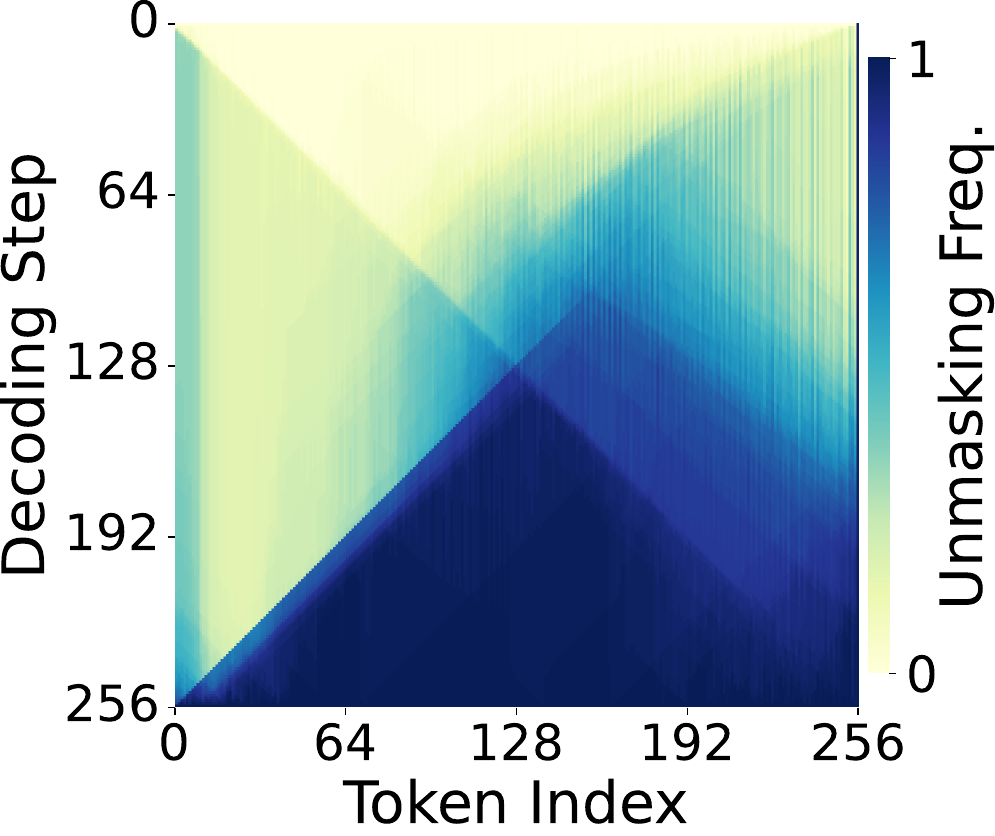}
        \label{fig:appendix_humaneval_confidence}
    }
    \hfill
    \subfigure[Entropy-based Sampling]{
        \includegraphics[width=0.465\linewidth]{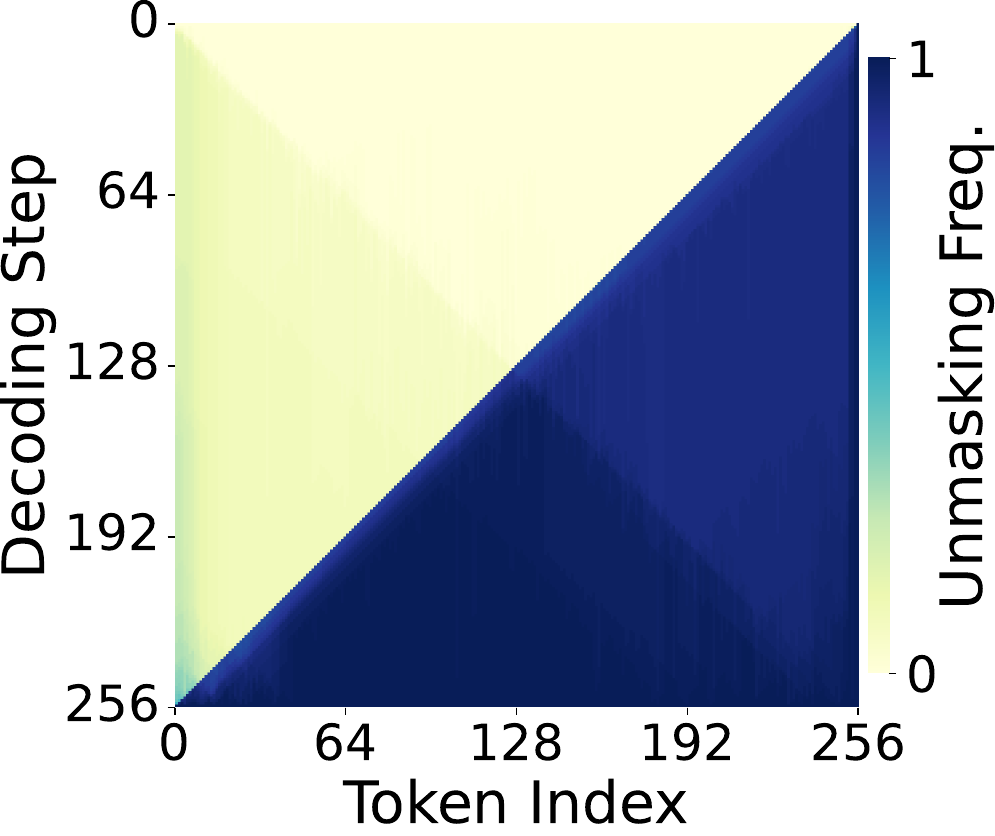}
        \label{fig:appendix_humaneval_entropy}
    }


    \subfigure[Margin-based Sampling]{
        \includegraphics[width=0.465\linewidth]{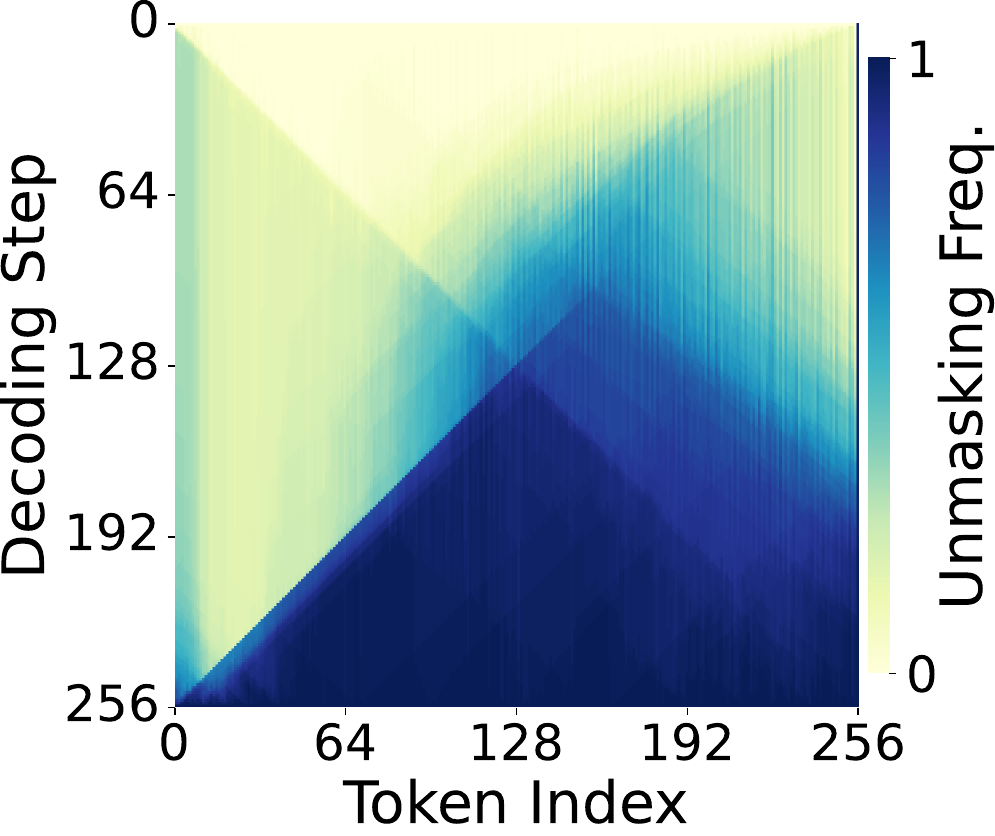}
        \label{fig:appendix_humaneval_margin}
    }
    \hfill
    \subfigure[Performance Comparison]{
        \includegraphics[width=0.465\linewidth]{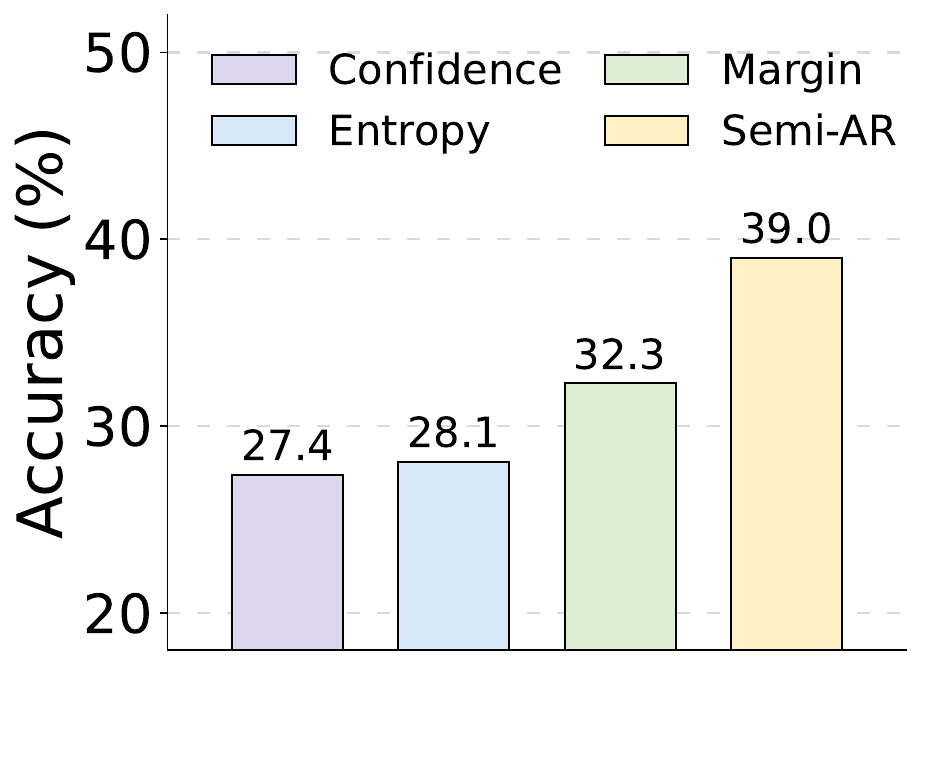}
        \label{fig:appendix_humaneval_performance}
    }
    \vspace{-0.5em}
    \caption{Consistent boundary-first unmasking patterns across various uncertainty metrics. (a-c) Visualization of unmasking probability for Confidence, Entropy, and Margin samplers on HumanEval. (d) Corresponding performance comparison.}
    \vspace{-1em}
    \label{fig:humaneval_method_comparison}
\end{figure}

%% file: appendix_figs/entropy_and_margin_GPQA.tex
\begin{figure}[t!]
    \centering
    \subfigure[Confidence-based Sampling]{
        \includegraphics[width=0.465\linewidth]{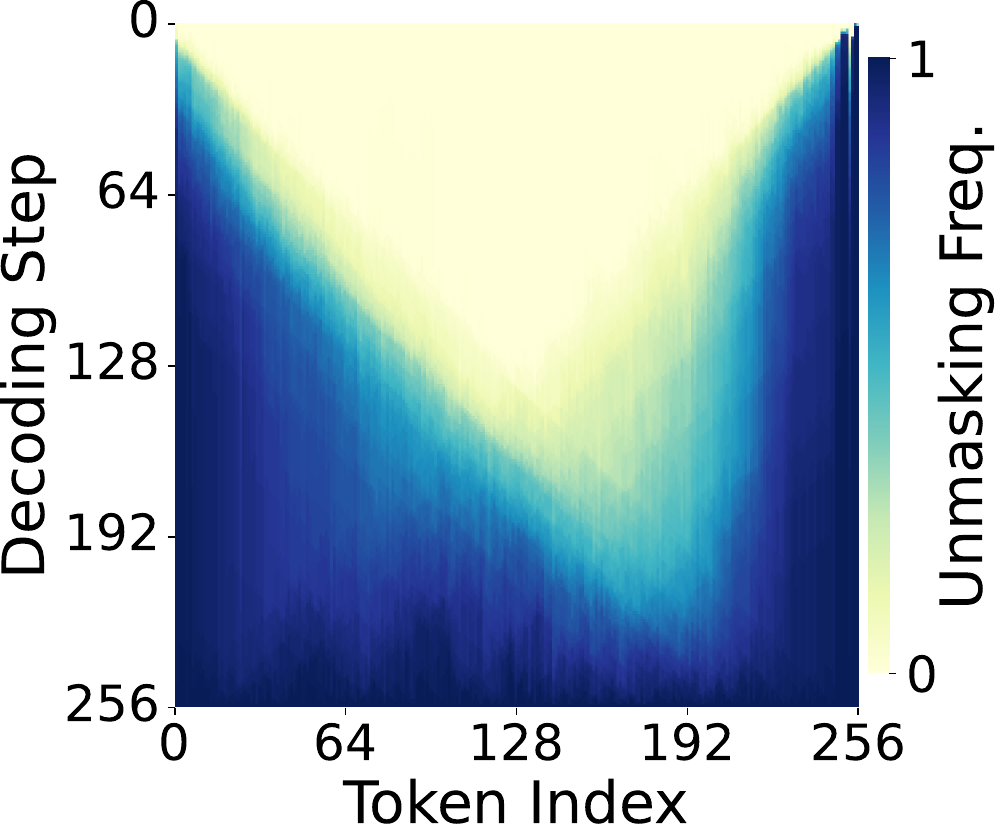}
        \label{fig:appendix_gpqa_confidence}
    }
    \hfill
    \subfigure[Entropy-based Sampling]{
        \includegraphics[width=0.465\linewidth]{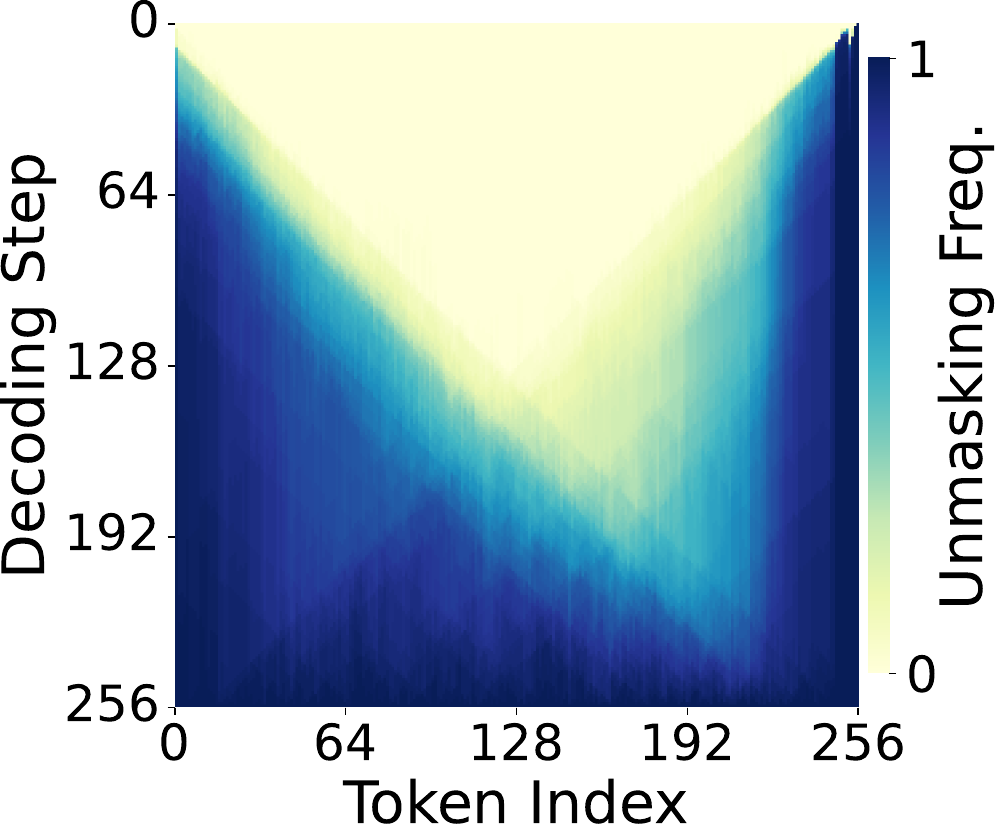}
        \label{fig:appendix_gpqa_entropy}
    }


    \subfigure[Margin-based Sampling]{
        \includegraphics[width=0.465\linewidth]{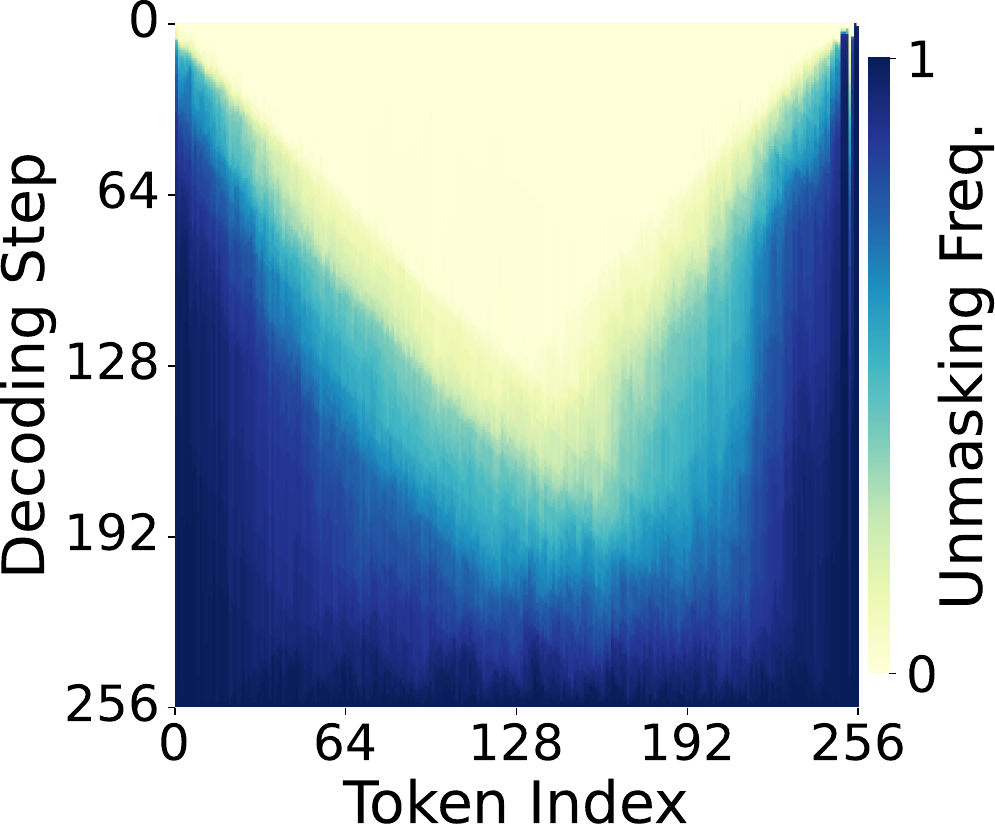}
        \label{fig:appendix_gpqa_margin}
    }
    \hfill
    \subfigure[Performance Comparison]{
        \includegraphics[width=0.465\linewidth]{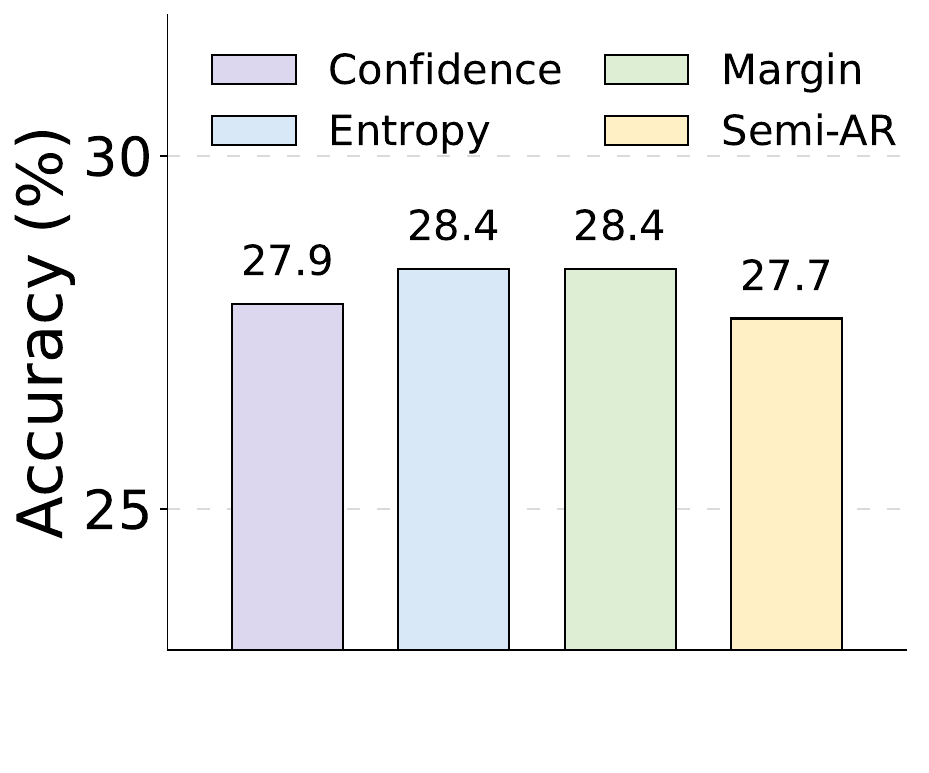}
        \label{fig:appendix_gpqa_performance}
    }
    \vspace{-0.5em}
    \caption{Consistent boundary-first unmasking patterns across various uncertainty metrics. (a-c) Visualization of unmasking probability for Confidence, Entropy, and Margin samplers on GPQA. (d) Corresponding performance comparison.}
    \vspace{-1em}
    \label{fig:gpqa_method_comparison}
\end{figure}

%% file: appendix_figs/entropy_and_margin_countdown.tex
\begin{figure}[!t]
    \centering
    \subfigure[Confidence-based Sampling]{
        \includegraphics[width=0.465\linewidth]{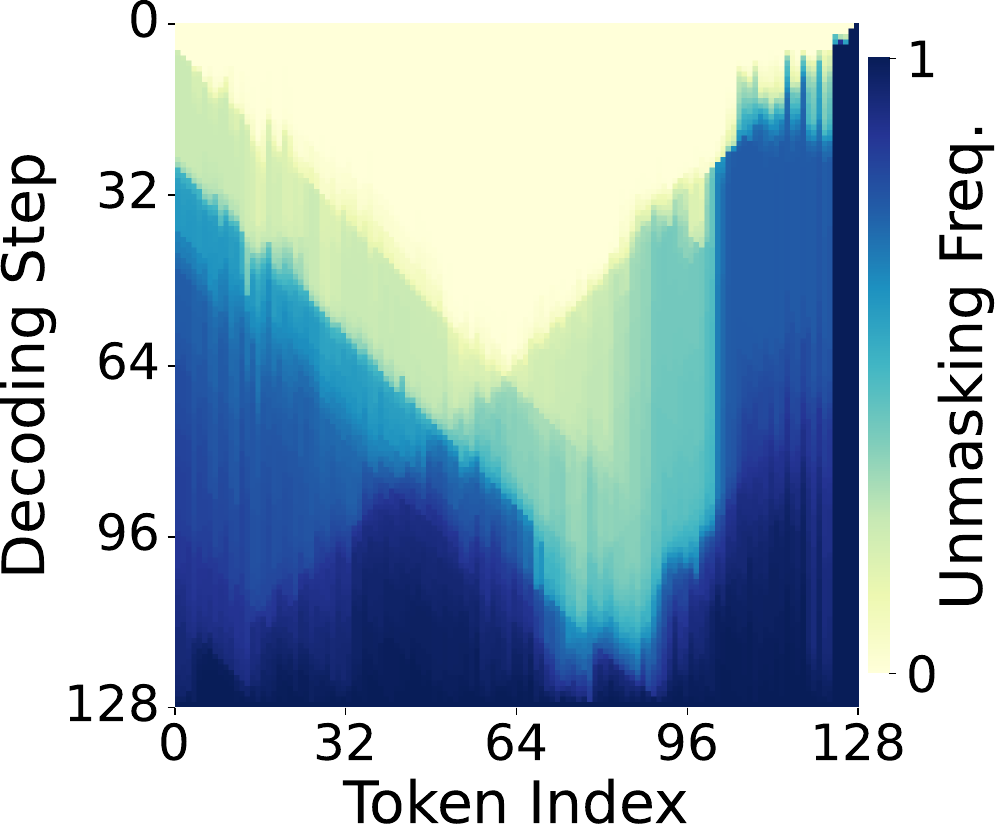}
        \label{fig:appendix_countdown_confidence}
    }
    \hfill
    \subfigure[Entropy-based Sampling]{
        \includegraphics[width=0.465\linewidth]{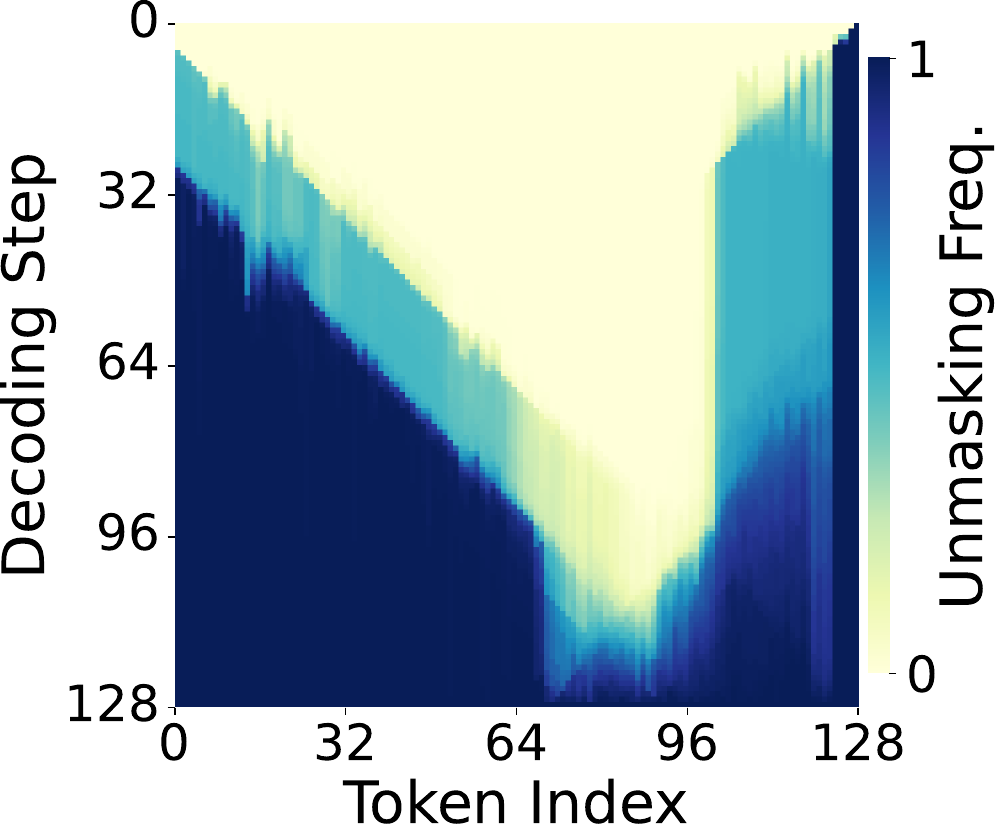}
        \label{fig:appendix_countdown_entropy}
    }


    \subfigure[Margin-based Sampling]{
        \includegraphics[width=0.465\linewidth]{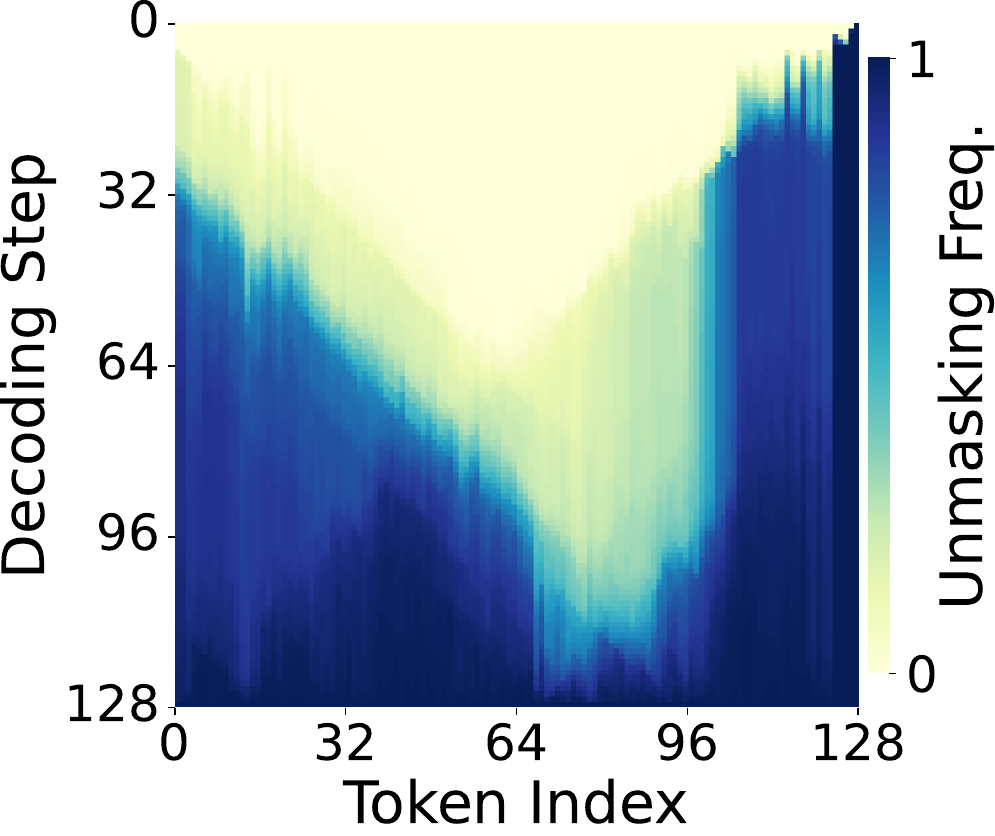}
        \label{fig:appendix_countdown_margin}
    }
    \hfill
    \subfigure[Performance Comparison]{
        \includegraphics[width=0.465\linewidth]{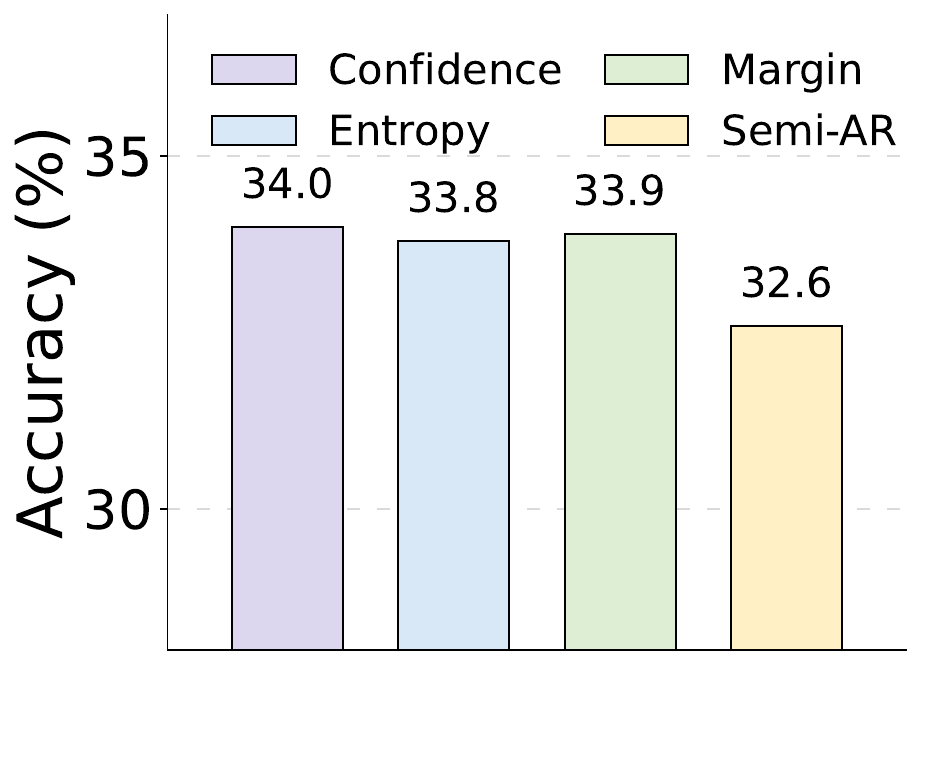}
        \label{fig:appendix_countdown_performance}
    }
    \vspace{-0.5em}
    \caption{Consistent boundary-first unmasking patterns across various uncertainty metrics. (a-c) Visualization of unmasking probability for Confidence, Entropy, and Margin samplers on Countdown. (d) Corresponding performance comparison.}
    \vspace{-1em}
    \label{fig:countdown_method_comparison}
\end{figure}

%% file: appendix_figs/entropy_and_margin_sudoku.tex
\begin{figure}[!t]
    \centering
    \subfigure[Confidence-based Sampling]{
        \includegraphics[width=0.465\linewidth]{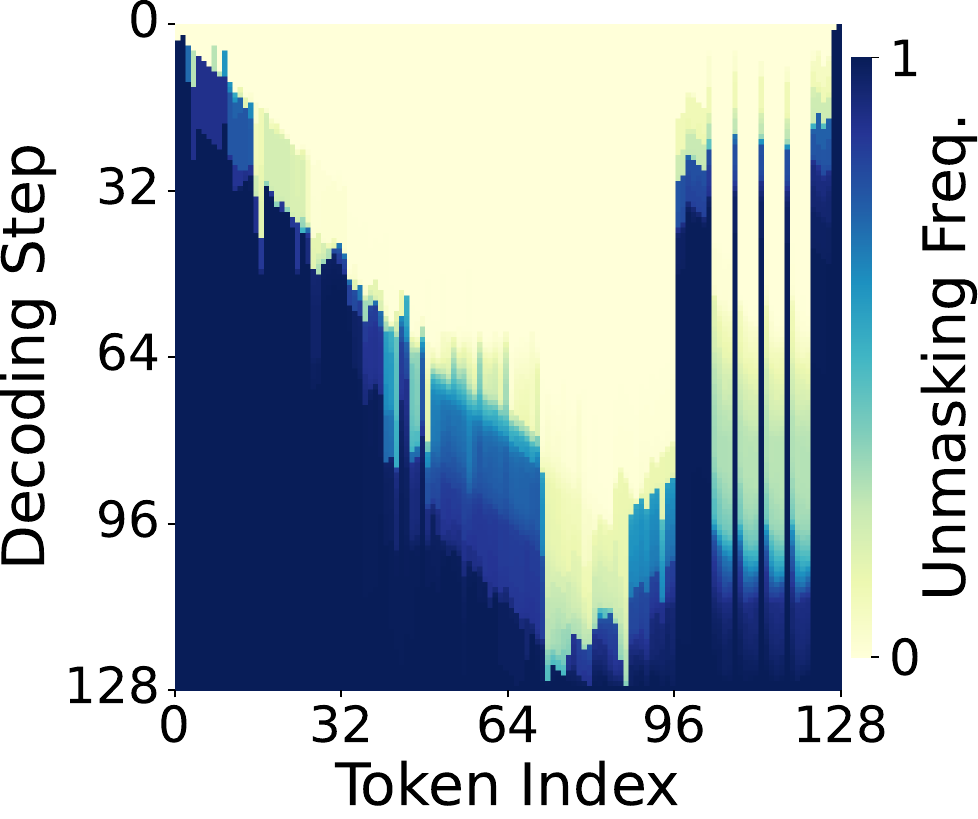}
        \label{fig:appendix_sudoku_confidence}
    }
    \hfill
    \subfigure[Entropy-based Sampling]{
        \includegraphics[width=0.465\linewidth]{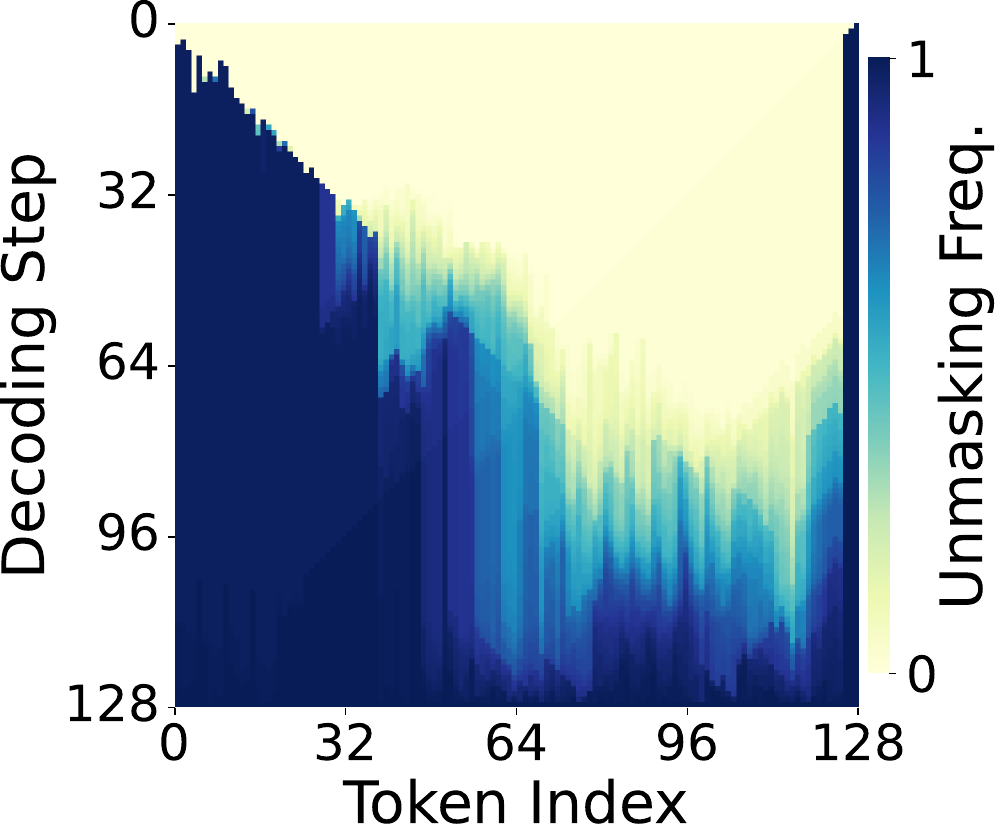}
        \label{fig:appendix_sudoku_entropy}
    }


    \subfigure[Margin-based Sampling]{
        \includegraphics[width=0.465\linewidth]{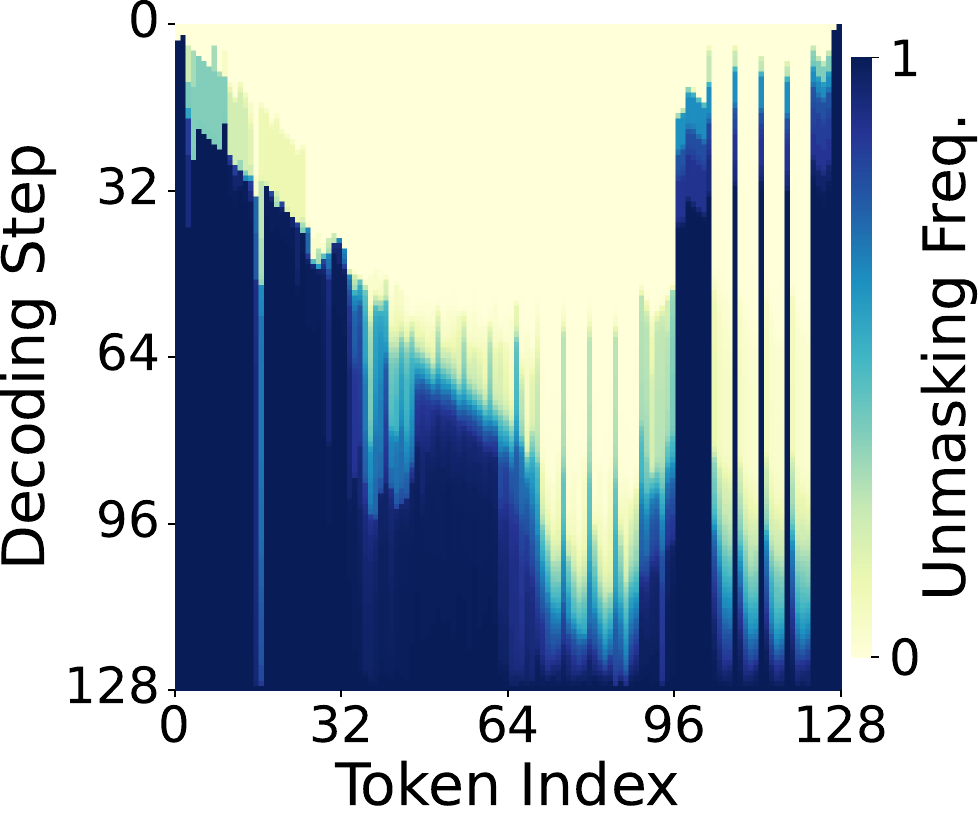}
        \label{fig:appendix_sudoku_margin}
    }
    \hfill
    \subfigure[Performance Comparison]{
        \includegraphics[width=0.465\linewidth]{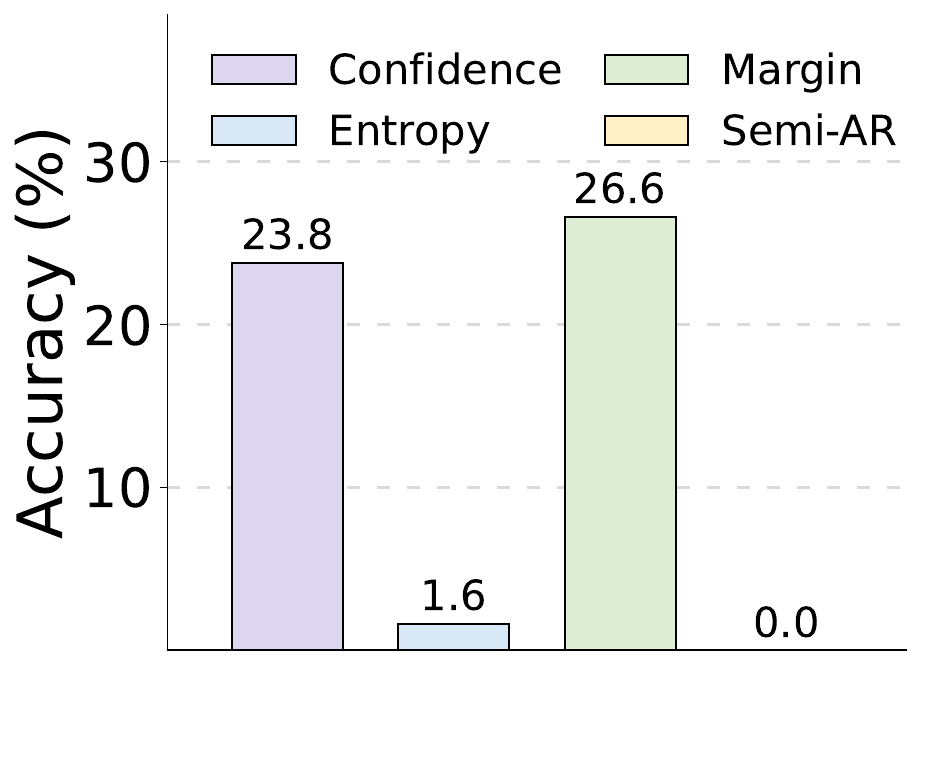}
        \label{fig:appendix_sudoku_performance}
    }
    \vspace{-0.5em}
    \caption{Consistent boundary-first unmasking patterns across various uncertainty metrics. (a-c) Visualization of unmasking probability for Confidence, Entropy, and Margin samplers on Sudoku. (d) Corresponding performance comparison.}
    \vspace{-1em}
    \label{fig:sudoku_method_comparison}
\end{figure}

%% file: tables/data_statistics.tex
\begin{table}[t]
\centering
\small
\begin{tabular}{lccc}
\toprule
Dataset & Task Type & \#Test  & Metric \\
\midrule
HumanEval& Code Generation     & 164  &  Pass@1 \\
MBPP     & Code Generation     & 427  &  Pass@1 \\
GSM8K     & Math Reasoning    & 1,319 &  Acc. \\
MATH500  & Math Reasoning      & 500  &  Acc. \\
GPQA     & Science QA       & 448 & Acc. \\
Countdown & Planning         & 1,000   & Acc. \\
Sudoku   & Planning          & 501  &  Acc. \\
\bottomrule
\end{tabular}
\caption{Statistics of evaluation datasets.}
\label{tab:dataset_stats}
\end{table}

%% file: figs/hyperparameters.tex
\begin{figure}[!t]
    \centering
    \subfigure[Effect of $\lambda$.]{
        \includegraphics[width=0.46\linewidth]{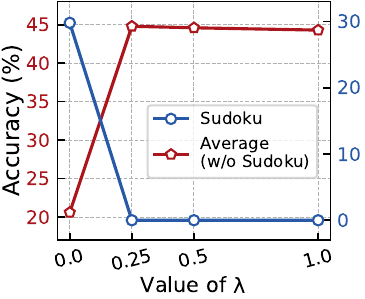}
        \label{fig:sub_lambda}
    }
    \hfill
    \subfigure[Effect of $\alpha$.]{
        \includegraphics[width=0.47\linewidth]{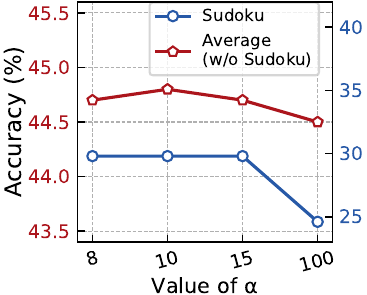}
        \label{fig:sub_alpha}
    }
    \caption{Analysis of decoding behavior under different $\lambda$ and $\alpha$ settings. We find that $\lambda$ substantially affects the induced decoding trajectory, whereas performance remains largely stable across a broad range of $\alpha$, suggesting that \method{} is not sensitive to the clipping threshold. To examine task-specific effects, we report average results across non-Sudoku datasets (blue) and separately for Sudoku (red).
}
    \label{fig:ablation_hyperparams}
\end{figure}


%% file: figs/entropy_performance.tex
\begin{figure}[!t]
    \centering
    \subfigure[GSM8K.]{
        \includegraphics[width=0.469\linewidth]{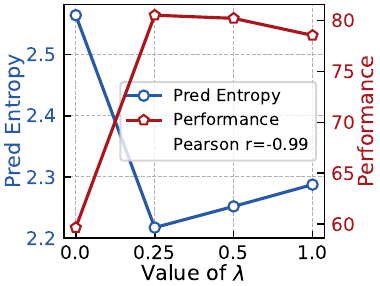}
        \label{fig:entropy_per_gsm}
    }
    \hfill
    \subfigure[HumanEval.]{
        \includegraphics[width=0.469\linewidth]{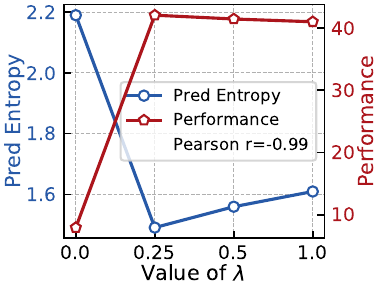}
        \label{fig:entropy_per_humaneval}
    }
    \caption{Correlation analysis between Mean Predictive Entropy and task performance across varying $\lambda$ values. The blue curve denotes mean predictive entropy (left axis), while the red curve indicates performance (right axis) on (a) GSM8K and (b) HumanEval. The strong negative correlation ($r \approx -0.99$) empirically validates that lower mean predictive entropy serves as a reliable indicator of higher reasoning quality.
}
    \label{fig:entropy_per}
\end{figure}


%% file: tables/appendix_different_corpus.tex
\begin{table*}[t]
\centering
\small 
\begin{tabular}{l cccccccc}
\toprule
\textbf{Calibration Corpus} & \textbf{HumanEval} & \textbf{MBPP} & \textbf{GSM8K} & \textbf{MATH500} & \textbf{GPQA} & \textbf{Countdown} & \textbf{Sudoku} & \textbf{Avg.}\\
\midrule
\rowcolor{gray!15}
\multicolumn{9}{c}{\textbf{\textit{LLaDA-Instruct-8B}}} \\
\rowcolor{blue!5}
Book \& Code (\citeyear{Zhu_2015_ICCV,openr1_math_220k}) & 42.1 & 47.3 & 79.2 & 35.0 & 27.9 & 36.3 & 24.4 & 41.7 \\
\rowcolor{red!5}
SlimPajama (\citeyear{cerebras2023slimpajama}) & 40.9 & 47.1 & 79.1 & 35.2 & 28.4 & 36.3 & 30.8 & 42.5 \\
\rowcolor{green!5}
C4 (16 GB) (\citeyear{raffel2020exploring}) & 42.1 & 47.8 & 79.2 & 34.8 & 29.2 & 36.3 & 29.8 & 42.7\\
\rowcolor{green!5}
C4 (50 GB) (\citeyear{raffel2020exploring}) & 42.2 & 47.8 & 79.2 & 36.0 & 29.0 & 36.3 & 30.0 & 42.9 \\
\rowcolor{green!5}
C4 (200 GB) (\citeyear{raffel2020exploring}) & 42.9 & 47.8 & 79.2 & 36.2 & 29.0 & 36.3 & 30.0 & 43.1 \\
\rowcolor{green!5}
C4 (400 GB) (\citeyear{raffel2020exploring}) & 42.9 & 47.8 & 79.2 & 36.2 & 29.0 & 36.3 & 30.0 & 43.1 \\

\bottomrule
\end{tabular}
\caption{Evaluation of \method{} across various calibration corpora. `Book \& Code' refers to a combination of BookCorpusOpen and OpenR1-Math-220K datasets. C4 ($x$ GB) denotes a sampled subset of the C4 dataset, where $x$ represents the size of the subset in gigabytes.}
\label{tab:diff_corpus}
\end{table*}

%% file: tables/main_res_dream.tex
\begin{table*}[t!]
\centering
\small 
\begin{tabular}{l cccccccc}
\toprule
\textbf{Methods \& LLMs} & \textbf{HumanEval} & \textbf{MBPP} & \textbf{GSM8K} & \textbf{MATH500} & \textbf{GPQA} & \textbf{Countdown} & \textbf{Sudoku} & \textbf{Avg.}\\
\midrule
\rowcolor{gray!15}
\multicolumn{9}{c}{\textbf{\textit{Autoregressive LLMs}}} \\
LLaMA-3.1-8B-Instruct & \underline{53.1} & \underline{56.7} & \textbf{83.9} & \underline{23.8} & \underline{31.0}  & \textbf{27.0} & 0.0 & \underline{39.4}\\
Mistral-7B-Instruct & 43.9 & 37.0 & 49.4 & 7.2 & 28.1  & \underline{22.7} & 0.0 & 26.9\\
Qwen-2.5-7B-Instruct & \textbf{78.1} & \textbf{62.8} & \underline{71.9} & \textbf{64.2} & \textbf{32.8}  & 7.7 & 0.0 & \textbf{45.3}\\
\midrule
\rowcolor{gray!15}
\multicolumn{9}{c}{\textbf{\textit{LLaDA-8B-Instruct}}} \\
Uniform (\citeyear{NEURIPS2021_958c5305}) & 15.2 & 24.6 & 48.8 & 15.0 & 29.0  & 14.4 & 2.2 & 21.3\\
Confidence (\citeyear{chang2022maskgit}) & 27.4 & 42.4 & 59.1 & 20.8 & 27.9  & 34.0 & 23.8 & 33.6\\
Entropy (\citeyear{dream2025}) & 28.1 & 42.2 & 60.9 & 11.2 & 28.4  & 33.8 & 1.6 & 29.4\\
Margin (\citeyear{kim2025train}) & 32.3 & 42.4 & 58.3 & 19.8 & 28.4  & 33.9 & \underline{26.6} & 34.5\\
EB-Sampler (\citeyear{ben2025accelerated}) & 26.8 & 43.3 & 61.2 & 11.6 & \textbf{29.5}  & \underline{34.1} & 24.2 & 33.0\\
Semi-AR$^\dagger$ (\citeyear{nie2025large}) & \underline{39.0} & \underline{45.2} & 77.9 & 27.6 & 27.7  & 32.6 & 0.0 & \underline{35.7}\\
Fast-dLLM$^\dagger$ (\citeyear{wu2025fast}) & 35.4 & 44.7 & \underline{78.2} & \underline{28.4} & 28.6  & 23.6 & 0.0 & 34.1 \\
\rowcolor{blue!15}
\textbf{\method{}} & \textbf{42.1} & \textbf{47.8} & \textbf{79.2} & \textbf{34.8} & \underline{29.2} & \textbf{36.3} & \textbf{29.8} & \textbf{42.7} \\
\midrule
\rowcolor{gray!15}
\multicolumn{9}{c}{\textbf{\textit{LLaDA-1.5-8B}}} \\
Uniform (\citeyear{NEURIPS2021_958c5305}) & 17.7 & 23.0 & 52.7 & 20.0 & 28.1  & 15.8 & 3.4 & 23.0\\
Confidence (\citeyear{chang2022maskgit}) & 28.1 & 43.3 & 60.7 & 22.8 & \underline{28.7}  & 33.8 & 24.8 & 34.6\\
Entropy (\citeyear{dream2025}) & 32.9 & 44.0 & 60.3 & 11.2 & 26.6  & \underline{34.7} & 0.2 & 30.0\\
Margin (\citeyear{kim2025train}) & 25.0 & 43.3 & 57.5 & 23.2 & 28.4  & 31.8 & \textbf{33.6} & 34.7\\
EB-Sampler (\citeyear{ben2025accelerated}) & 32.9 & 43.6 & 61.1 & 13.4 & 26.6  & 34.6 & 0.2 & 30.3\\
Semi-AR$^\dagger$ (\citeyear{nie2025large}) & \underline{39.6} & \underline{46.8} & 80.7 & \underline{34.2} & 26.1  & 32.4 & 0.0 & \underline{37.1}\\
Fast-dLLM$^\dagger$ (\citeyear{wu2025fast}) & 37.2 & 46.1 & \underline{80.8} & 31.2 & 27.9  & 23.6 & 0.0 & 36.7\\
\rowcolor{blue!15}
\textbf{\method{}} & \textbf{46.3} & \textbf{49.9} & \textbf{82.2} & \textbf{37.4} & \textbf{28.8} & \textbf{35.0} & \underline{33.4} & \textbf{44.7}\\
\rowcolor{gray!15}
\multicolumn{9}{c}{\textbf{\textit{Dream-v0-Instruct-7B}}} \\
Uniform (\citeyear{NEURIPS2021_958c5305}) & 17.7 & 31.9 & 31.5 & 17.0 & 32.8  & 4.1 & \textbf{0.2} & 19.3\\
Confidence (\citeyear{chang2022maskgit}) & 27.4 & 41.5 & 45.4 & 20.8 & \underline{35.3}  & \textbf{19.8} & 0.0 & 27.2\\
Entropy (\citeyear{dream2025}) & 26.2 & 42.4 & 36.8 & 17.0 & 33.5  & \underline{19.0} & 0.0 & 25.0\\
Margin (\citeyear{kim2025train}) & \underline{28.1} & 41.7 & \underline{48.3} & \underline{22.0} & \textbf{35.7}  & \underline{19.0} & 0.0 & \underline{27.8}\\
EB-Sampler (\citeyear{ben2025accelerated}) & 26.8 & \underline{43.6} & 37.5 & 17.4  & 33.3 & 18.6 & 0.0 & 25.3\\
Fast-dLLM (\citeyear{wu2025fast}) & 12.8 & 23.9 & 46.1 & 19.2 & 34.4  & 11.6 & 0.0 & 21.1 \\
\rowcolor{blue!15}
\textbf{\method{}} & \textbf{57.9} & \textbf{56.4} & \textbf{76.4} & \textbf{37.8} & 33.9 & 18.4 & 0.0 & \textbf{40.1}\\
\bottomrule
\end{tabular}
\caption{
Experimental results on seven different benchmarks. The best score is highlighted in \textbf{bold}, and the second-best is \underline{underlined}. Following prior practices~\cite{nie2025large}, we use 4-shot settings for GSM8K and MATH500, 5-shot for GPQA and Sudoku, 0-shot for HumanEval and MBPP, and 3-shot for Countdown. Methods marked with $^\dagger$ denote samplers using the Semi-AR strategy. All settings follow prior works~\cite{nie2025large,zhao2025d1,wang2025mro} for fair comparison.}
\label{tab:main_results_revised_with_dream}
\end{table*}

%% file: tables/case_gsm8k.tex
\begin{figure*}[t!]
    \centering
    \includegraphics[width=0.97\linewidth]{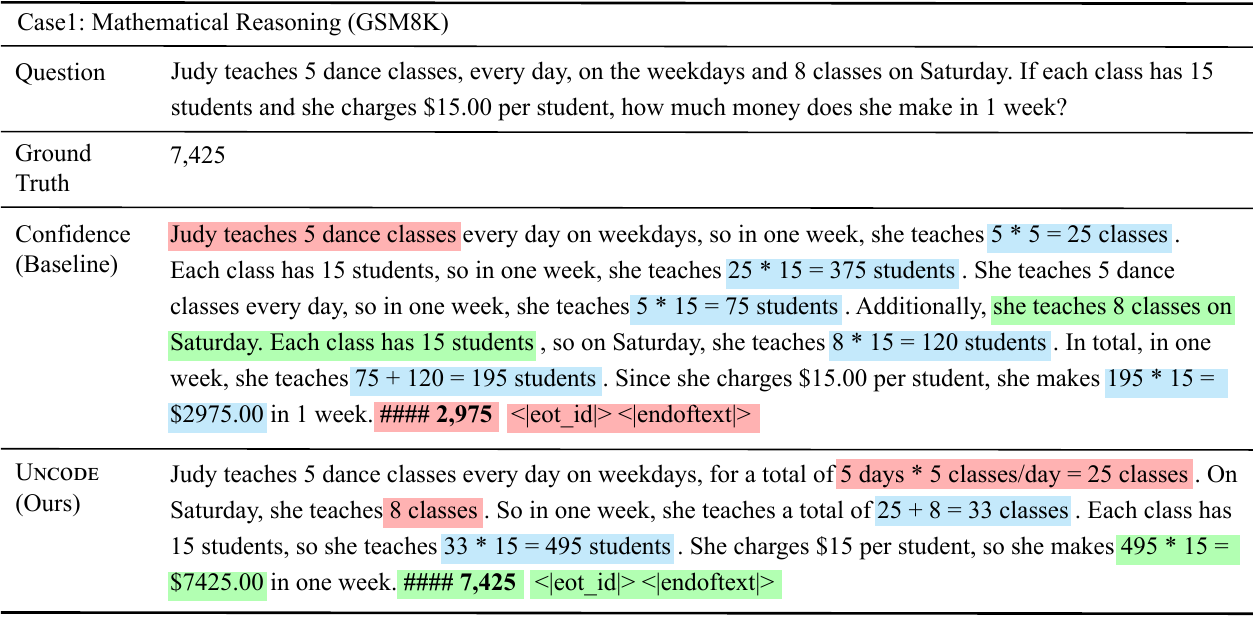}
    \caption{A case study on the GSM8K dataset illustrating how the generation order impacts mathematical reasoning. The baseline's premature commitment to an incorrect answer highlights a critical failure mode, whereas \method{}'s coherent process leads to the correct solution. The generation timing is color-coded as follows: \colorbox{red!30}{Early Stage}, \colorbox{cyan!20}{Middle Stage}, and \colorbox{green!30}{Late Stage}.}
    \label{tab:case_study_gsm8k}
\end{figure*}

%% file: tables/case_gsm8k_2.tex
\begin{figure*}[t!]
    \centering
    \includegraphics[width=0.97\linewidth]{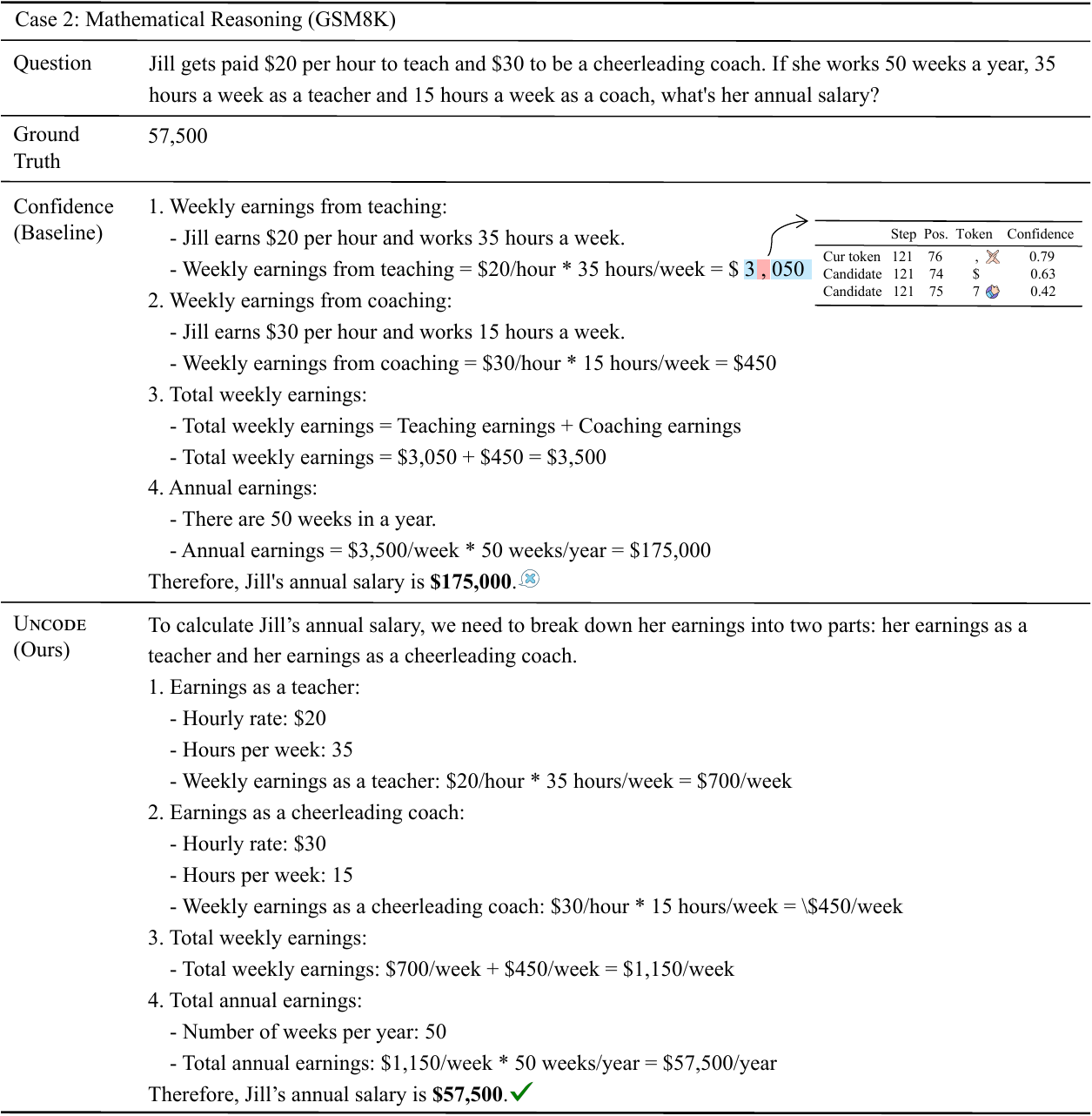}
    \caption{A case study on the GSM8K dataset illustrating how the trivial token bias impacts mathematical reasoning. The baseline decodes trivial tokens at critical steps (e.g., the comma is decoded at step 121, at position 76, instead of the 7 at position 75), locking the reasoning space (e.g., the comma, used as a thousand separator, causes failure as 0,700 is invalid), and leading to incorrect predictions. In contrast, \method{} suppresses the selection of trivial tokens, resulting in the correct solution. The generation timing is color-coded as follows: \colorbox{red!30}{Early Stage} and \colorbox{cyan!20}{Middle Stage}.}
    \label{tab:case_study_gsm8k_2}
\end{figure*}

%% file: tables/case_code.tex
\begin{figure*}[t!]
    \centering
    \includegraphics[width=0.97\linewidth]{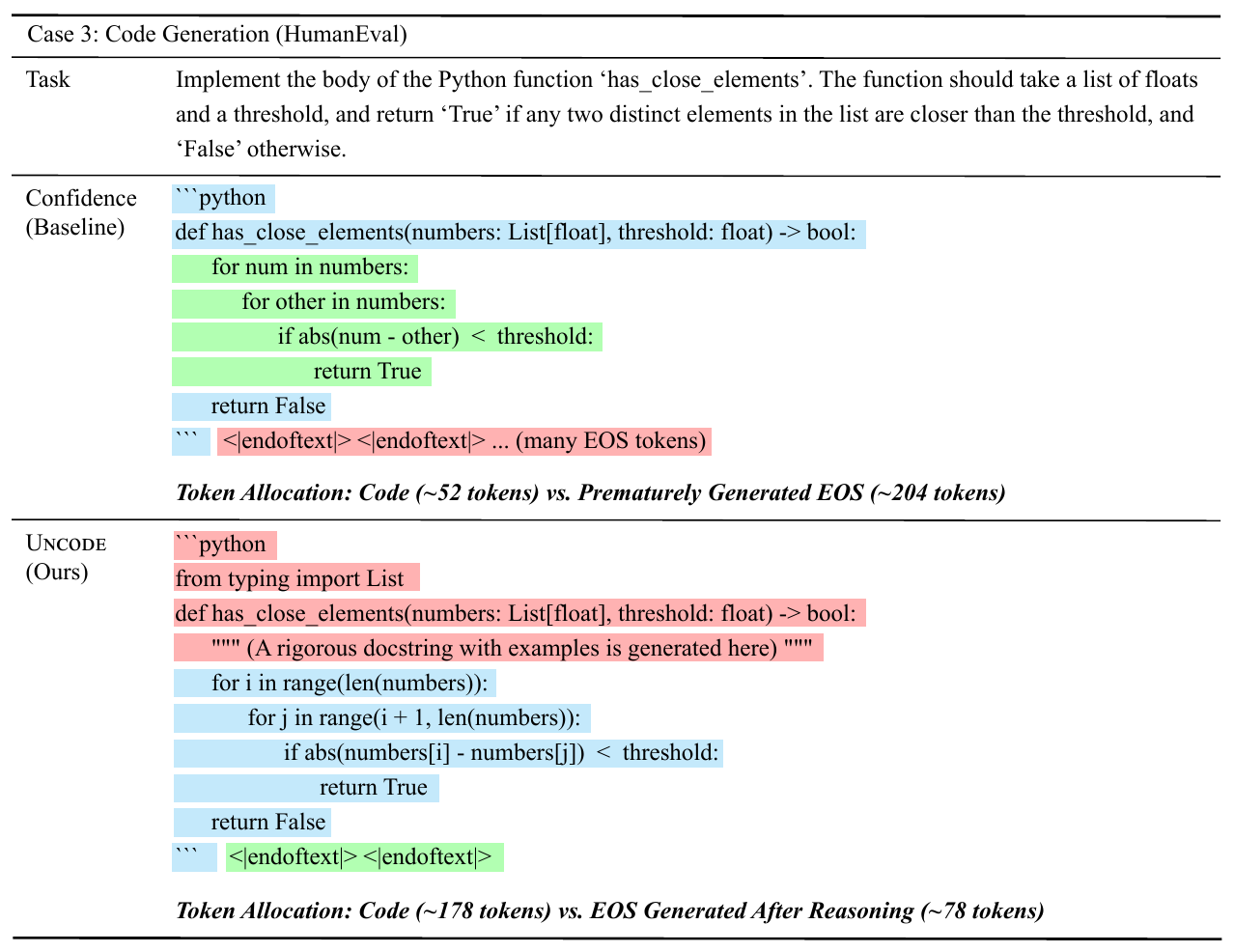}
    \caption{Case study on the HumanEval dataset for code generation. The baseline method exhibits catastrophic failure due to the premature generation of trivial tokens, whereas \method{}, through explicit trajectory control, produces robust and complete code. The generation timing is color-coded as follows: \colorbox{red!30}{Early Stage}, \colorbox{cyan!20}{Middle Stage}, and \colorbox{green!30}{Late Stage}.}
    \label{tab:case_study_humaneval}
\end{figure*}

%% file: bib_new.bib
@misc{cerebras2023slimpajama,
    author = {Soboleva, Daria and Al-Khateeb, Faisal and Myers, Robert and Steeves, Jacob R and Hestness, Joel and Dey, Nolan},
    title = {{SlimPajama: A 627B token cleaned and deduplicated version of RedPajama}},
    url = {https://huggingface.co/datasets/cerebras/SlimPajama-627B},
    year = {2023}
}

@dataset{openr1_math_220k,
    author = {Open R1 Project},
    note = {Accessed: 2025-11-24},
    publisher = {Hugging Face},
    title = {OpenR1-Math-220k: A Large Annotated Dataset for Math Reasoning},
    url = {https://huggingface.co/datasets/open-r1/OpenR1-Math-220k},
    year = {2025}
}

@inproceedings{Zhu_2015_ICCV,
    author = {Yukun Zhu and
Ryan Kiros and
Richard S. Zemel and
Ruslan Salakhutdinov and
Raquel Urtasun and
Antonio Torralba and
Sanja Fidler},
    bibsource = {dblp computer science bibliography, https://dblp.org},
    booktitle = {2015 {IEEE} International Conference on Computer Vision, {ICCV} 2015,
Santiago, Chile, December 7-13, 2015},
    doi = {10.1109/ICCV.2015.11},
    pages = {19--27},
    publisher = {{IEEE} Computer Society},
    title = {Aligning Books and Movies: Towards Story-Like Visual Explanations
by Watching Movies and Reading Books},
    url = {https://doi.org/10.1109/ICCV.2015.11},
    year = {2015}
}

@article{ben2025accelerated,
    author = {Ben-Hamu, Heli and Gat, Itai and Severo, Daniel and Nolte, Niklas and Karrer, Brian},
    journal = {ArXiv preprint},
    title = {Accelerated Sampling from Masked Diffusion Models via Entropy Bounded Unmasking},
    url = {https://arxiv.org/abs/2505.24857},
    volume = {abs/2505.24857},
    year = {2025}
}

@article{nie2025large,
    author = {Shen Nie and
Fengqi Zhu and
Zebin You and
Xiaolu Zhang and
Jingyang Ou and
Jun Hu and
Jun Zhou and
Yankai Lin and
Ji{-}Rong Wen and
Chongxuan Li},
    title = {Large Language Diffusion Models},
    url = {https://doi.org/10.48550/arXiv.2502.09992},
    year = {2025}
}

@inproceedings{zhang2024pretraining,
    author = {Weichao Zhang and
Ruqing Zhang and
Jiafeng Guo and
Maarten de Rijke and
Yixing Fan and
Xueqi Cheng},
    booktitle = {Proceedings of the 2024 Conference on Empirical Methods in Natural
Language Processing, {EMNLP} 2024, Miami, FL, USA, November 12-16,
2024},
    pages = {5263--5274},
    title = {Pretraining Data Detection for Large Language Models: {A} Divergence-based
Calibration Method},
    url = {https://doi.org/10.18653/v1/2024.emnlp-main.300},
    year = {2024}
}

@article{wu2025fast,
    author = {Wu, Chengyue and Zhang, Hao and Xue, Shuchen and Liu, Zhijian and Diao, Shizhe and Zhu, Ligeng and Luo, Ping and Han, Song and Xie, Enze},
    journal = {ArXiv preprint},
    title = {Fast-dllm: Training-free acceleration of diffusion llm by enabling kv cache and parallel decoding},
    url = {https://arxiv.org/abs/2505.22618},
    volume = {abs/2505.22618},
    year = {2025}
}

@article{peng2025path,
    author = {Zhangzhi Peng and
Zachary Bezemek and
Sawan Patel and
Jarrid Rector{-}Brooks and
Sherwood Yao and
Alexander Tong and
Pranam Chatterjee},
    journal = {CoRR},
    title = {Path Planning for Masked Diffusion Model Sampling},
    url = {https://doi.org/10.48550/arXiv.2502.03540},
    year = {2025}
}

@article{kim2025train,
    author = {Jaeyeon Kim and
Kulin Shah and
Vasilis Kontonis and
Sham M. Kakade and
Sitan Chen},
    journal = {CoRR},
    title = {Train for the Worst, Plan for the Best: Understanding Token Ordering
in Masked Diffusions},
    url = {https://doi.org/10.48550/arXiv.2502.06768},
    year = {2025}
}

@inproceedings{nie2024scaling,
    author = {Shen Nie and
Fengqi Zhu and
Chao Du and
Tianyu Pang and
Qian Liu and
Guangtao Zeng and
Min Lin and
Chongxuan Li},
    booktitle = {The Thirteenth International Conference on Learning Representations,
{ICLR} 2025, Singapore, April 24-28, 2025},
    title = {Scaling up Masked Diffusion Models on Text},
    url = {https://openreview.net/forum?id=WNvvwK0tut},
    year = {2025}
}

@article{dream2025,
  title={Dream 7b: Diffusion large language models},
  author={Ye, Jiacheng and Xie, Zhihui and Zheng, Lin and Gao, Jiahui and Wu, Zirui and Jiang, Xin and Li, Zhenguo and Kong, Lingpeng},
  url = {https://arxiv.org/abs/2508.15487},
  journal={arXiv preprint arXiv:2508.15487},
  year={2025}
}

@inproceedings{chang2022maskgit,
    author = {Huiwen Chang and
Han Zhang and
Lu Jiang and
Ce Liu and
William T. Freeman},
    bibsource = {dblp computer science bibliography, https://dblp.org},
    booktitle = {{IEEE/CVF} Conference on Computer Vision and Pattern Recognition,
{CVPR} 2022, New Orleans, LA, USA, June 18-24, 2022},
    doi = {10.1109/CVPR52688.2022.01103},
    pages = {11305--11315},
    publisher = {{IEEE}},
    title = {MaskGIT: Masked Generative Image Transformer},
    url = {https://doi.org/10.1109/CVPR52688.2022.01103},
    year = {2022}
}

@article{zhu2025llada,
    author = {Fengqi Zhu and
Rongzhen Wang and
Shen Nie and
Xiaolu Zhang and
Chunwei Wu and
Jun Hu and
Jun Zhou and
Jianfei Chen and
Yankai Lin and
Ji{-}Rong Wen and
Chongxuan Li},
    journal = {CoRR},
    title = {LLaDA 1.5: Variance-Reduced Preference Optimization for Large Language
Diffusion Models},
    url = {https://doi.org/10.48550/arXiv.2505.19223},
    year = {2025}
}

@inproceedings{arriola2025block,
    author = {Marianne Arriola and
Aaron Gokaslan and
Justin T. Chiu and
Zhihan Yang and
Zhixuan Qi and
Jiaqi Han and
Subham Sekhar Sahoo and
Volodymyr Kuleshov},
    booktitle = {The Thirteenth International Conference on Learning Representations,
{ICLR} 2025, Singapore, April 24-28, 2025},
    title = {Block Diffusion: Interpolating Between Autoregressive and Diffusion
Language Models},
    url = {https://openreview.net/forum?id=tyEyYT267x},
    year = {2025}
}

@article{cobbe2021training,
    author = {Karl Cobbe and
Vineet Kosaraju and
Mohammad Bavarian and
Mark Chen and
Heewoo Jun and
Lukasz Kaiser and
Matthias Plappert and
Jerry Tworek and
Jacob Hilton and
Reiichiro Nakano and
Christopher Hesse and
John Schulman},
    journal = {ArXiv preprint},
    title = {Training Verifiers to Solve Math Word Problems},
    url = {https://arxiv.org/abs/2110.14168},
    volume = {abs/2110.14168},
    year = {2021}
}

@inproceedings{lightman2023let,
    author = {Hunter Lightman and
Vineet Kosaraju and
Yuri Burda and
Harrison Edwards and
Bowen Baker and
Teddy Lee and
Jan Leike and
John Schulman and
Ilya Sutskever and
Karl Cobbe},
    bibsource = {dblp computer science bibliography, https://dblp.org},
    booktitle = {The Twelfth International Conference on Learning Representations,
{ICLR} 2024, Vienna, Austria, May 7-11, 2024},
    publisher = {OpenReview.net},
    title = {Let's Verify Step by Step},
    url = {https://openreview.net/forum?id=v8L0pN6EOi},
    year = {2024}
}

@article{chen2021evaluating,
    author = {Mark Chen and
Jerry Tworek and
Heewoo Jun and
Qiming Yuan and
Henrique Pond{\'{e}} de Oliveira Pinto and
Jared Kaplan and
Harri Edwards},
    journal = {ArXiv preprint},
    title = {Evaluating Large Language Models Trained on Code},
    url = {https://arxiv.org/abs/2107.03374},
    volume = {abs/2107.03374},
    year = {2021}
}

@article{austin2021program,
    author = {Jacob Austin and
Augustus Odena and
Maxwell I. Nye and
Maarten Bosma and
Henryk Michalewski and
David Dohan and
Ellen Jiang and
Carrie J. Cai and
Michael Terry and
Quoc V. Le and
Charles Sutton},
    journal = {ArXiv preprint},
    title = {Program Synthesis with Large Language Models},
    url = {https://arxiv.org/abs/2108.07732},
    volume = {abs/2108.07732},
    year = {2021}
}

@article{nolte2024transformers,
    author = {Niklas Nolte and
Ouail Kitouni and
Adina Williams and
Mike Rabbat and
Mark Ibrahim},
    journal = {CoRR},
    title = {Transformers Can Navigate Mazes With Multi-Step Prediction},
    url = {https://doi.org/10.48550/arXiv.2412.05117},
    year = {2024}
}

@inproceedings{ye2024beyond,
    author = {Jiacheng Ye and
Jiahui Gao and
Shansan Gong and
Lin Zheng and
Xin Jiang and
Zhenguo Li and
Lingpeng Kong},
    booktitle = {The Thirteenth International Conference on Learning Representations,
{ICLR} 2025, Singapore, April 24-28, 2025},
    title = {Beyond Autoregression: Discrete Diffusion for Complex Reasoning and
Planning},
    url = {https://openreview.net/forum?id=NRYgUzSPZz},
    year = {2025}
}

@article{rein2024gpqa,
    author = {David Rein and
Betty Li Hou and
Asa Cooper Stickland and
Jackson Petty and
Richard Yuanzhe Pang and
Julien Dirani and
Julian Michael and
Samuel R. Bowman},
    journal = {CoRR},
    title = {{GPQA:} {A} Graduate-Level Google-Proof Q{\&}A Benchmark},
    url = {https://doi.org/10.48550/arXiv.2311.12022},
    year = {2023}
}

@article{zhao2025d1,
    author = {Siyan Zhao and
Devaansh Gupta and
Qinqing Zheng and
Aditya Grover},
    journal = {CoRR},
    title = {d1: Scaling Reasoning in Diffusion Large Language Models via Reinforcement
Learning},
    url = {https://doi.org/10.48550/arXiv.2504.12216},
    year = {2025}
}

@article{dubey2024llama,
    author = {Abhimanyu Dubey and
Abhinav Jauhri and
Abhinav Pandey and
Abhishek Kadian and
Ahmad Al{-}Dahle and
Aiesha Letman and
Akhil Mathur},
    journal = {CoRR},
    title = {The Llama 3 Herd of Models},
    url = {https://doi.org/10.48550/arXiv.2407.21783},
    year = {2024}
}

@article{jiang2023mistral7b,
    author = {Albert Q. Jiang and
Alexandre Sablayrolles and
Arthur Mensch and
Chris Bamford and
Devendra Singh Chaplot and
Diego de Las Casas and
Florian Bressand and
Gianna Lengyel and
Guillaume Lample and
Lucile Saulnier and
L{\'{e}}lio Renard Lavaud and
Marie{-}Anne Lachaux and
Pierre Stock and
Teven Le Scao and
Thibaut Lavril and
Thomas Wang and
Timoth{\'{e}}e Lacroix and
William El Sayed},
    journal = {CoRR},
    title = {Mistral 7B},
    url = {https://doi.org/10.48550/arXiv.2310.06825},
    year = {2023}
}

@article{team2024qwen2,
    author = {An Yang and
Baosong Yang and
Binyuan Hui and
Bo Zheng and
Bowen Yu and
Chang Zhou and
Chengpeng Li and
Chengyuan Li},
    journal = {ArXiv preprint},
    title = {Qwen2 Technical Report},
    url = {https://arxiv.org/abs/2407.10671},
    volume = {abs/2407.10671},
    year = {2024}
}

@inproceedings{NEURIPS2021_958c5305,
    author = {Jacob Austin and
Daniel D. Johnson and
Jonathan Ho and
Daniel Tarlow and
Rianne van den Berg},
    bibsource = {dblp computer science bibliography, https://dblp.org},
    booktitle = {Advances in Neural Information Processing Systems 34: Annual Conference
on Neural Information Processing Systems 2021, NeurIPS 2021, December
6-14, 2021, virtual},
    editor = {Marc'Aurelio Ranzato and
Alina Beygelzimer and
Yann N. Dauphin and
Percy Liang and
Jennifer Wortman Vaughan},
    pages = {17981--17993},
    title = {Structured Denoising Diffusion Models in Discrete State-Spaces},
    url = {https://proceedings.neurips.cc/paper/2021/hash/958c530554f78bcd8e97125b70e6973d-Abstract.html},
    year = {2021}
}

@article{chen2023accelerating,
    author = {Charlie Chen and
Sebastian Borgeaud and
Geoffrey Irving and
Jean{-}Baptiste Lespiau and
Laurent Sifre and
John Jumper},
    journal = {CoRR},
    title = {Accelerating Large Language Model Decoding with Speculative Sampling},
    url = {https://doi.org/10.48550/arXiv.2302.01318},
    year = {2023}
}

@article{israel2025accelerating,
    author = {Israel, Daniel and Broeck, Guy Van den and Grover, Aditya},
    journal = {CoRR},
    title = {Accelerating Diffusion LLMs via Adaptive Parallel Decoding},
    url = {https://doi.org/10.48550/arXiv.2506.00413},
    year = {2025}
}

@inproceedings{lou2023discrete,
    author = {Aaron Lou and
Chenlin Meng and
Stefano Ermon},
    bibsource = {dblp computer science bibliography, https://dblp.org},
    booktitle = {Forty-first International Conference on Machine Learning, {ICML} 2024,
Vienna, Austria, July 21-27, 2024},
    publisher = {OpenReview.net},
    title = {Discrete Diffusion Modeling by Estimating the Ratios of the Data Distribution},
    url = {https://openreview.net/forum?id=CNicRIVIPA},
    year = {2024}
}

@article{raffel2020exploring,
    author = {Colin Raffel and
Noam Shazeer and
Adam Roberts and
Katherine Lee and
Sharan Narang and
Michael Matena and
Yanqi Zhou and
Wei Li and
Peter J. Liu},
    bibsource = {dblp computer science bibliography, https://dblp.org},
    journal = {J. Mach. Learn. Res.},
    pages = {140:1--140:67},
    title = {Exploring the Limits of Transfer Learning with a Unified Text-to-Text
Transformer},
    url = {http://jmlr.org/papers/v21/20-074.html},
    volume = {21},
    year = {2020}
}

@inproceedings{wang2024chain,
    author = {Xuezhi Wang and
Denny Zhou},
    bibsource = {dblp computer science bibliography, https://dblp.org},
    booktitle = {Advances in Neural Information Processing Systems 38: Annual Conference
on Neural Information Processing Systems 2024, NeurIPS 2024, Vancouver,
BC, Canada, December 10 - 15, 2024},
    editor = {Amir Globersons and
Lester Mackey and
Danielle Belgrave and
Angela Fan and
Ulrich Paquet and
Jakub M. Tomczak and
Cheng Zhang},
    title = {Chain-of-Thought Reasoning Without Prompting},
    url = {http://papers.nips.cc/paper\_files/paper/2024/hash/7a8e7fd295aa04eac4b470ae27f8785c-Abstract-Conference.html},
    year = {2024}
}

@article{li2025survey,
    author = {Li, Tianyi and Chen, Mingda and Guo, Bowei and Shen, Zhiqiang},
    journal = {ArXiv preprint},
    title = {A survey on diffusion language models},
    url = {https://arxiv.org/abs/2508.10875},
    volume = {abs/2508.10875},
    year = {2025}
}

@article{khanna2025mercury,
  title={Mercury: Ultra-fast language models based on diffusion},
  author={Khanna, Samar and Kharbanda, Siddhant and Li, Shufan and Varma, Harshit and Wang, Eric and Birnbaum, Sawyer and Luo, Ziyang and Miraoui, Yanis and Palrecha, Akash and Ermon, Stefano and others},
  journal={arXiv preprint arXiv:2506.17298},
  volume={1},
  year={2025}
}

@article{wang2025mro,
    author = {Wang, Chenglong and Gan, Yang and Zhou, Hang and Hu, Chi and Mu, Yongyu and Song, Kai and Yang, Murun and Li, Bei and Zhang, Chunliang and Liu, Tongran and others},
    journal = {ArXiv preprint},
    title = {MRO: Enhancing Reasoning in Diffusion Language Models via Multi-Reward Optimization},
    url = {https://arxiv.org/abs/2510.21473},
    volume = {abs/2510.21473},
    year = {2025}
}

@inproceedings{li2023diffusion,
    author = {Yifan Li and
Kun Zhou and
Wayne Xin Zhao and
Ji{-}Rong Wen},
    bibsource = {dblp computer science bibliography, https://dblp.org},
    booktitle = {Proceedings of the Thirty-Second International Joint Conference on
Artificial Intelligence, {IJCAI} 2023, 19th-25th August 2023, Macao,
SAR, China},
    doi = {10.24963/IJCAI.2023/750},
    pages = {6692--6701},
    publisher = {ijcai.org},
    title = {Diffusion Models for Non-autoregressive Text Generation: {A} Survey},
    url = {https://doi.org/10.24963/ijcai.2023/750},
    year = {2023}
}

@article{wu2025emergence,
    author = {Wu, Xinyi and Wang, Yifei and Jegelka, Stefanie and Jadbabaie, Ali},
    journal = {ArXiv preprint},
    title = {On the emergence of position bias in transformers},
    url = {https://arxiv.org/abs/2502.01951},
    volume = {abs/2502.01951},
    year = {2025}
}

@article{jiang2025d,
    author = {Jiang, Yuchu and Cai, Yue and Luo, Xiangzhong and Fu, Jiale and Wang, Jiarui and Liu, Chonghan and Yang, Xu},
    journal = {ArXiv preprint},
    title = {d2 Cache: Accelerating Diffusion-Based LLMs via Dual Adaptive Caching},
    url = {https://arxiv.org/abs/2509.23094},
    volume = {abs/2509.23094},
    year = {2025}
}

@article{song2025seed,
    author = {Song, Yuxuan and Zhang, Zheng and Luo, Cheng and Gao, Pengyang and Xia, Fan and Luo, Hao and Li, Zheng and Yang, Yuehang and Yu, Hongli and Qu, Xingwei and others},
    journal = {ArXiv preprint},
    title = {Seed diffusion: A large-scale diffusion language model with high-speed inference},
    url = {https://arxiv.org/abs/2508.02193},
    volume = {abs/2508.02193},
    year = {2025}
}

@article{garg2025masked,
    author = {Garg, Prateek and Kohli, Bhavya and Sarawagi, Sunita},
    journal = {ArXiv preprint},
    title = {Masked Diffusion Models are Secretly Learned-Order Autoregressive Models},
    url = {https://arxiv.org/abs/2511.19152},
    volume = {abs/2511.19152},
    year = {2025}
}

@book{cover1999elements,
    author = {Cover, Thomas M},
    publisher = {John Wiley \& Sons},
    title = {Elements of information theory},
    url = {https://www.academia.edu/download/58191902/Elements_of_Information_Theory_Elements.pdf},
    year = {1999}
}

@article{lee2025lookahead,
    author = {Lee, Sanghyun and Kim, Seungryong and Park, Jongho and Park, Dongmin},
    journal = {ArXiv preprint},
    title = {Lookahead Unmasking Elicits Accurate Decoding in Diffusion Language Models},
    url = {https://arxiv.org/abs/2511.05563},
    volume = {abs/2511.05563},
    year = {2025}
}

@article{li2025beyond,
    author = {Li, Jinsong and Dong, Xiaoyi and Zang, Yuhang and Cao, Yuhang and Wang, Jiaqi and Lin, Dahua},
    journal = {ArXiv preprint},
    title = {Beyond fixed: Training-free variable-length denoising for diffusion large language models},
    url = {https://arxiv.org/abs/2508.00819},
    volume = {abs/2508.00819},
    year = {2025}
}

@article{yu2025dimple,
    author = {Yu, Runpeng and Ma, Xinyin and Wang, Xinchao},
    journal = {ArXiv preprint},
    title = {Dimple: Discrete diffusion multimodal large language model with parallel decoding},
    url = {https://arxiv.org/abs/2505.16990},
    volume = {abs/2505.16990},
    year = {2025}
}

@article{wei2025accelerating,
    author = {Wei, Qingyan and Zhang, Yaojie and Liu, Zhiyuan and Liu, Dongrui and Zhang, Linfeng},
    journal = {ArXiv preprint},
    title = {Accelerating Diffusion Large Language Models with SlowFast: The Three Golden Principles},
    url = {https://arxiv.org/abs/2506.10848},
    volume = {abs/2506.10848},
    year = {2025}
}

@article{hong2025wide,
    author = {Hong, Feng and Yu, Geng and Ye, Yushi and Huang, Haicheng and Zheng, Huangjie and Zhang, Ya and Wang, Yanfeng and Yao, Jiangchao},
    journal = {ArXiv preprint},
    title = {Wide-in, narrow-out: Revokable decoding for efficient and effective dllms},
    url = {https://arxiv.org/abs/2507.18578},
    volume = {abs/2507.18578},
    year = {2025}
}

@article{horvitz2025no,
    author = {Horvitz, Zachary and Singhal, Raghav and Zou, Hao and Domingo-Enrich, Carles and Yu, Zhou and Ranganath, Rajesh and McKeown, Kathleen},
    journal = {ArXiv preprint},
    title = {No Compute Left Behind: Rethinking Reasoning and Sampling with Masked Diffusion Models},
    url = {https://arxiv.org/abs/2510.19990},
    volume = {abs/2510.19990},
    year = {2025}
}

@misc{eval-harness,
    author = {Gao, Leo and Tow, Jonathan and Abbasi, Baber and Biderman, Stella and Black, Sid and DiPofi, Anthony and Foster, Charles and Golding, Laurence and Hsu, Jeffrey and Le Noac'h, Alain and Li, Haonan and McDonell, Kyle and Muennighoff, Niklas and Ociepa, Chris and Phang, Jason and Reynolds, Laria and Schoelkopf, Hailey and Skowron, Aviya and Sutawika, Lintang and Tang, Eric and Thite, Anish and Wang, Ben and Wang, Kevin and Zou, Andy},
    doi = {10.5281/zenodo.12608602},
    publisher = {Zenodo},
    title = {The Language Model Evaluation Harness},
    url = {https://zenodo.org/records/12608602},
    version = {v0.4.3},
    year = {2024}
}

@article{martinez2024mitigating,
    author = {Martinez, Richard Diehl and Goriely, Z{\'e}bulon and Caines, Andrew and Buttery, Paula and Beinborn, Lisa},
    journal = {ArXiv preprint},
    title = {Mitigating frequency bias and anisotropy in language model pre-training with syntactic smoothing},
    url = {https://arxiv.org/abs/2410.11462},
    volume = {abs/2410.11462},
    year = {2024}
}
